%% file: neurips22.tex
\pdfoutput=1

\documentclass{article}

\PassOptionsToPackage{numbers}{natbib}

\usepackage[final]{neurips_2022}

\usepackage[utf8]{inputenc} 
\usepackage[T1]{fontenc}    
\usepackage{hyperref}       
\usepackage{url}            
\usepackage{booktabs}       
\usepackage{amsfonts}       
\usepackage{nicefrac}       
\usepackage{microtype}      
\usepackage{xcolor}         
\input{preamble}
\usepackage{tablefootnote}

\title{GREED: A Neural Framework for Learning Graph Distance Functions}

\author{%
  Rishabh Ranjan, Siddharth Grover \\
  Department of Computer Science \& Engineering, IIT Delhi, India\\
  \texttt{\{rishabh.ranjan.cs118, siddharth.grover.cs118\}@cse.iitd.ac.in} \\
  \And
  Sourav Medya \\
  Department of Computer Science \\
  University of Illinois, Chicago, USA\\
   \texttt{medya@uic.edu} \\
   \And
  Venkat Chakravarthy, Yogish Sabharwal \\
  IBM Research\\
  Delhi, India \\
  \texttt{\{vechakra, ysabharwal\}}@in.ibm.com \\
   \AND
  Sayan Ranu \\
  Department of Computer Science \& Engineering and Yardi School of AI (Jointly), IIT Delhi, India \\
  \texttt{sayanranu@cse.iitd.ac.in}  
}

\begin{document}

\maketitle

\input{0_abstract}
\input{1_introduction}

\input{2_background}
\input{3_prob}
\input{4_approach}
\input{5_experiment}

\input{7_conclusion}

\bibliographystyle{plain}
\bibliography{neurips22}

\newpage
\appendix

\input{8_appendix}


\end{document}

%% file: preamble.tex
\usepackage{soul}
\usepackage{url}
\usepackage{float}
\usepackage[utf8]{inputenc}
\usepackage[export]{adjustbox}
\usepackage{graphicx}
\usepackage{amsmath}
\usepackage{amssymb}
\usepackage{booktabs}
\usepackage{algorithm}
\usepackage[noend]{algpseudocode}
\usepackage{comment}
\usepackage{subcaption}
\usepackage{booktabs} 
\usepackage{graphicx,array}
\usepackage{color,soul,xspace}

\usepackage{comment}
\usepackage{epsfig}
\usepackage{mathrsfs,mdwlist,enumitem}

\newcommand{\imdb}{\texttt{IMDB}\xspace}
\newcommand{\aids}{\texttt{AIDS'}\xspace}
\newcommand{\linux}{\texttt{Linux}\xspace}
\newcommand{\seer}{\texttt{CiteSeer}\xspace}
\newcommand{\cora}{\texttt{Cora\_ML}\xspace}
\newcommand{\protein}{\texttt{Protein}\xspace}
\newcommand{\pubmed}{\texttt{PubMed}\xspace}
\newcommand{\dblp}{\texttt{Dblp}\xspace}
\newcommand{\amz}{\texttt{Amazon}\xspace}

\newcommand{\aidsg}{\texttt{AIDS}\xspace}

\newcommand{\branch}{\textsc{Branch}\xspace}

\newcommand{\f}{\textsc{Mip-F2}\xspace}
\newcommand{\sedg}{\sed(\CG_1,\CG_2)\xspace}

\newcommand{\name}{\textsc{Greed}\xspace}
\newcommand{\gotsim}{\textsc{GotSIM}\xspace}

\newcommand{\gnns}{\textsc{Gnn}s\xspace}
\newcommand{\sed}{\textsc{Sed}\xspace}
\newcommand{\ged}{\textsc{Ged}\xspace}

\newcommand{\gnn}{\textsc{Gnn}\xspace}
\newcommand{\mlp}{\textsc{Mlp}\xspace}
\newcommand{\neuromatch}{\textsc{NeuroMatch}\xspace}
\newcommand{\nsc}{\textsc{Nsc}\xspace}
\newcommand{\simgnn}{\textsc{SimGnn}\xspace}
\newcommand{\genn}{\textsc{Genn-A$^*$}\xspace}
\newcommand{\hmnr}{\textsc{H$^2$MN}\xspace}
\newcommand{\hmnn}{\textsc{H$^2$MN-ne}\xspace}
\newcommand{\gin}{\textsc{Gin}\xspace}
\newcommand{\gine}{\textsc{Gine}\xspace}
\newcommand{\gsage}{\textsc{GraphSage}\xspace}

\newcommand{\shadow}{\textsc{shaDow}\xspace}
\newcommand{\rwr}{\textsc{Rwr}\xspace}
\newcommand{\RW}{\textsc{Rw}\xspace}

\newcommand{\rev}[1]{\textcolor{black}{#1}}

\newcommand{\CG}{\mathcal{G}\xspace}
\newcommand{\CT}{\mathcal{T}\xspace}
\newcommand{\CQ}{\mathcal{Q}\xspace}

\newcommand{\CN}{\mathcal{N}\xspace}

\newcommand{\CP}{\mathcal{P}\xspace}
\newcommand{\CV}{\mathcal{V}\xspace}
\newcommand{\CE}{\mathcal{E}\xspace}

\newcommand{\CZ}{\mathbf{Z}\xspace}
\newcommand{\CS}{\mathcal{S}\xspace}
\newcommand{\CF}{\mathcal{F}\xspace}

\newcommand{\CL}{\mathcal{L}\xspace}
\newcommand{\cx}{\mathbf{x}\xspace}
\newcommand{\cy}{\mathbf{y}\xspace}
\newcommand{\ch}{\mathbf{h}\xspace}
\newcommand{\ce}{\mathbf{e}\xspace}

\newcommand{\cz}{\mathbf{z}\xspace}

\newcommand{\lhs}{\textsc{Lhs}\xspace}
\newcommand{\rhs}{\textsc{Rhs}\xspace}
\newcommand{\lb}{\textsc{Lb}\xspace}
\newcommand{\ub}{\textsc{Ub}\xspace}
\newcommand{\relu}{\text{ReLU}}

\setlist{nolistsep,leftmargin=*}

\newtheorem{thm}{\textbf{Theorem}}

\newtheorem{definition}{\textbf{Definition}}
\newtheorem{lem}{\textbf{Lemma}}

\newtheorem{cor}{\textbf{Corollary}}

\newtheorem{obs}{\textbf{Observation}}

\newtheorem{example}{\textbf{Example}}
\newtheorem{prob}{\textbf{Problem}}

\newcommand{\revision}[1]{\textcolor{black}{#1}}


%% file: 0_abstract.tex
\begin{abstract}
Among various distance functions for graphs, graph and subgraph edit distances (\ged and \sed respectively) are two of the most popular and expressive measures. Unfortunately, exact computations for both are NP-hard. To overcome this computational bottleneck, neural approaches to learn and predict edit distance in polynomial time have received much interest. While considerable progress has been made, there exist limitations that need to be addressed. First, the efficacy of an approximate distance function lies not only in its approximation accuracy, but also in the preservation of its properties. To elaborate, although \ged is a metric, its neural approximations do not provide such a guarantee. This prohibits their usage in higher order tasks that rely on metric distance functions, such as clustering or indexing. Second, several existing frameworks for \ged do not extend to \sed due to \sed being asymmetric. In this work, we design a novel siamese graph neural network called \name, which through a carefully crafted inductive bias, learns \ged and \sed in a property-preserving manner. Through extensive experiments across $10$ real graph datasets containing up to $7$ million edges, we establish that \name is not only more accurate than the state of the art, but also up to $3$ orders of magnitude faster. Even more significantly, due to preserving the triangle inequality, the generated embeddings are indexable and consequently, even in a CPU-only environment, \name is up to $50$ times faster than GPU-powered baselines for graph / subgraph retrieval.
\end{abstract}


%% file: 1_introduction.tex
\section{Introduction and Related Work}
\label{sec:intro}
\begin{figure}[t]
    \centering
    \includegraphics[width=4.2in, valign=top]{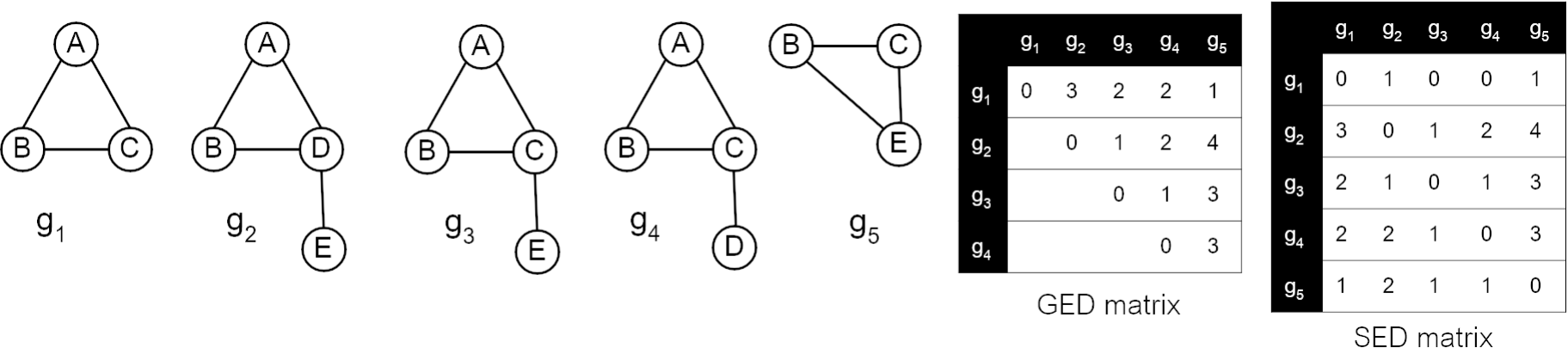}\hfill
        \includegraphics[width=1.3in, valign=top]{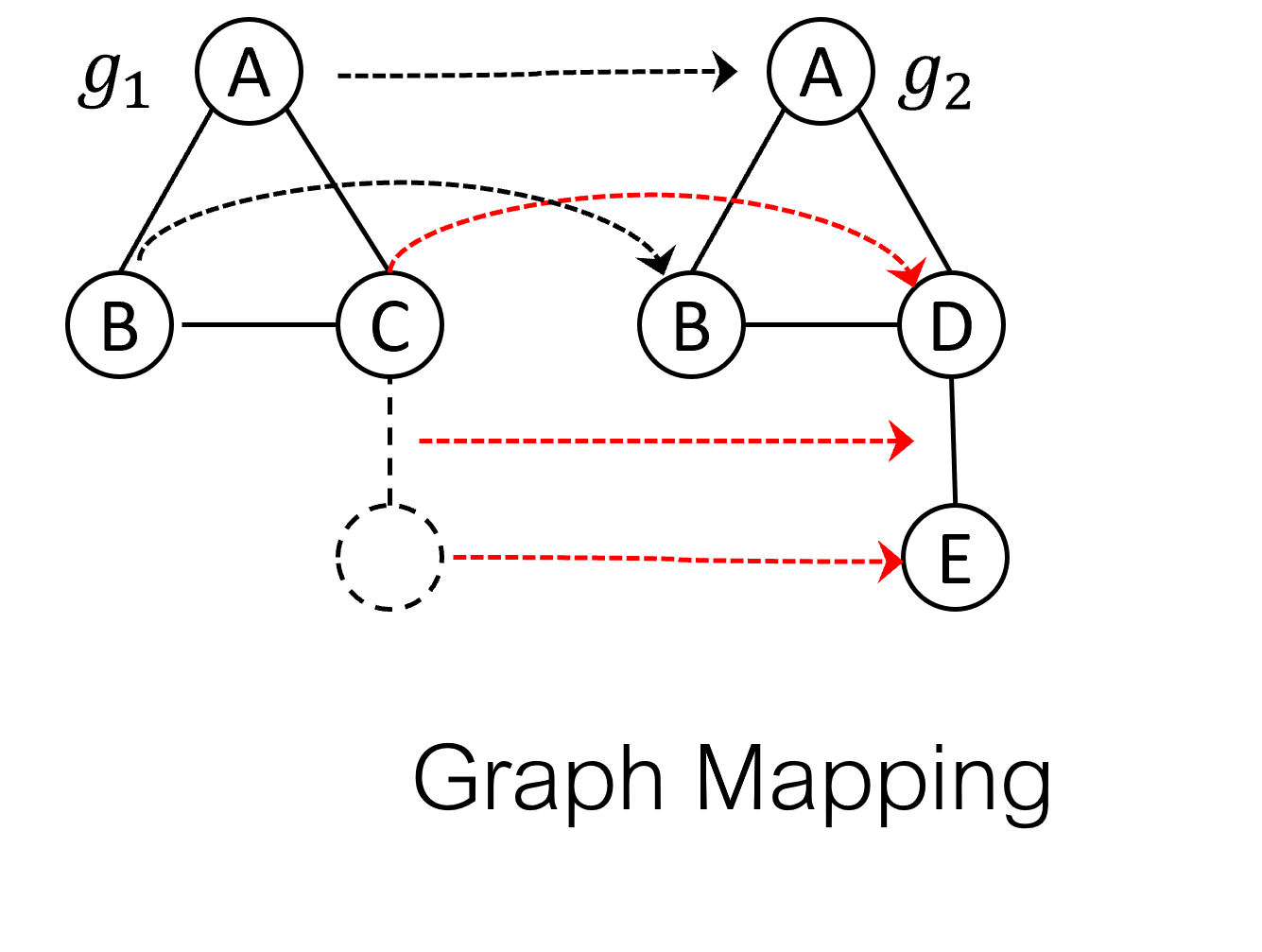}
    \caption{\textbf{A sample set of graphs ($g_1$-$g_5$), their corresponding \ged and \sed matrices and an example of a graph mapping from $g_1$ to $g_2$. The dashed nodes and edges in the mapping represent dummy nodes and edges. The red arrows denote either insertion or change of label.}}
    \label{fig:gedsed}
    \vspace{-0.20in}
\end{figure}

A distance function on any dataset, including graphs, is a fundamental operator. Among several distance measures on graphs, \textit{edit distance} is one of the most powerful and popular mechanisms~\citep{ged1,ged2,stars,ctree}. Edit distance can be posed in two forms: \textit{graph edit distance} (\ged) and \textit{subgraph edit distance} (\sed). Given two graphs $\CG_1$ and $\CG_2$, $\ged(\CG_1,\CG_2)$ returns the minimum cost of \textit{edits} needed to convert $\CG_1$ to $\CG_2$, i.e., for $\CG_1$ to become \textit{isomorphic} to $\CG_2$. An edit can be the addition or deletion of edges and nodes, or the replacement of edge or node labels, with an associated cost. In $\sed(\CG_1,\CG_2)$, the goal is to identify the minimum cost of edits so that $\CG_1$ is a subgraph (\textit{subgraph isomorphic}) of $\CG_2$. For examples, see Fig.~\ref{fig:gedsed}. 

\ged is typically restricted to graph databases containing small graphs to facilitate distance computation with queries of similar sizes. As an example, given a repository of molecules, and a query molecule, we may want to identify the closest molecule in the repository that is similar to the query~\cite{nbindex,ctree}. \sed, on the other hand, is useful when the database has large graphs and the query is a comparatively smaller graph. As examples, subgraph queries are used on knowledge graphs for analogy reasoning~\citep{kg}. In PPI and chemical compounds, \sed is of central importance to identify functional motifs and binding pockets~\citep{reaction, ctree, vldbppi, graphsig_jcim,pgraphsig,graphsig}.
Unfortunately, both \ged and \sed are NP-hard to compute~\citep{stars,ctree}. To mitigate this computational bottleneck, several heuristics~\citep{ged3,ged4} and index structures~\citep{ctree,stars,ged1,ged2} have been proposed. Recently, graph neural networks have been shown to be effective in learning and predicting \ged~\citep{simgnn,cvpr,icmlged,graphsim,h2mn,hgmn,graphotsim,tagsim}. The basic goal in all these algorithms is to learn a neural model from a training set of graph pairs and their distances, such that, at inference time, given an unseen graph pair, we are able to predict its distance accurately.
\textcolor{black}{Other works seek to incorporate non-neural graph matching solvers~\cite{bbgm}, or generic integer linear programming solvers~\cite{ilploss,comboptnet}, to learn the graph matching task from natural data such as images in an end-to-end trainable manner. Here, the NP-hardness of the problem is relegated to the non-neural component, and the neural part is only concerned with representation learning, unlike in our setting where we want the neural network to handle both.}
In the domain of subgraphs, \neuromatch~\cite{neuromatch} \rev{and \textsc{IsoNet}~\cite{isonet} generate embeddings to detect subgraph isomorphism.}  
\nsc~\cite{Wang2020Inductive} generates subgraph level embeddings that can count the number of subgraph instances of a query graph on a target graph.  
  While the progress made is impressive, there is scope to do more.
  \looseness=-1
\begin{itemize}
\item \textbf{Preservation of theoretical properties:} \ged is a \textit{metric} distance function. While \sed is not metric due to being asymmetric, it satisfies the \textit{triangle inequality}, \textit{non-negativity}, and \textit{subgraph-identity} ($\sed=0$ for subgraphs). Metrics (and triangle inequality) exhibit significant computational advantages over non-metrics. Specifically, operations such as clustering~\cite{cluster1}, nearest neighbor search~\cite{index1,index2,index4,mtree}, outlier detection~\cite{outlier} and diameter computation~\cite{diameter} admit efficient algorithms precisely when the objects being studied are embedded in a metric space. 
  Existing neural approaches do not preserve these properties, which limits their usability for these higher order tasks.
  
\item {\bf Indexable embeddings:} Given  graph pair $\CG_1$ and $\CG_2$, neural approaches first embed them into a feature space. Next, they compute a distance on these feature vectors, which is an approximation of the distance in the original graph space. The literature on indexing range and $k$-NN queries over feature vectors is rich~\cite{ti1,ti2,mtree}. Index structures typically allow sub-linear computation costs with respect to the database size. Unfortunately, none of the existing neural approaches generate indexable feature vectors since they perform \textit{pair-dependent} computations. Specifically, the neural computations on $\CG_1$ depend on both $\CG_1$ and $\CG_2$ (and same for $\CG_2$). Consequently, the computations can only be done at query-time and thereby negating the possibility of indexing pre-computed feature space embeddings.

\item  \textbf{Modeling \sed:} Prior to this work, there have been no neural approaches for \sed. Further, existing neural methods to learning \ged cannot easily be adapted to learn \sed. While \ged is symmetric, \sed is not. Several neural architectures for \ged have the assumption of symmetry at its core and hence modeling \sed is non-trivial~\cite{simgnn,graphsim,icmlged}. 

\item \textbf{Exponential Search Space: } Computing $\sedg$ conceptually requires us to compare the query graph $\CG_1$ with the exponentially many subgraphs of the target graph $\CG_2$. Therefore, it is imperative that the model has an efficient and effective mechanism to prune the search space without compromising on the prediction accuracy. 
     
     
     
\end{itemize}
In this work, we address the above limitations through the following contributions.
\begin{itemize}
    \item \textbf{Novel neural architecture:} We address the above mentioned  challenges through a novel architecture called \name: \underline{GR}aph \underline{E}mbeddings for \underline{E}dit \underline{D}istances. \name utilizes a \textit{siamese graph isomorphism network}~\citep{gin} to embed graphs in a \textit{pair-independent} fashion. A simple, but theoretically well-characterized, function on this embedding space predicts the \sed and \ged. 
  The carefully crafted prediction function serves as an \textit{inductive bias} for the model, which, in addition to enabling high generalization accuracy, preserves the metric property of \ged and the triangle inequality of \sed in the embedding space.
    \item\textbf{Indexable embeddings:} Owing to pair-independent embeddings and preservation of the triangle inequality over the embedding space for both \sed and \ged, \name can exploit the rich literature on index structures~\citep{ti1,ti2,mtree} to boost efficiency.
    \item\textbf{Accurate, Fast and Scalable:} Extensive experiments on real graph datasets containing up to a million nodes establish that \name is more accurate in both \ged and \sed when compared to the state of the art algorithms and is more than $3$ orders of magnitude faster in range and $k$-NN queries. Furthermore, owing to indexable embeddings, even in a CPU-only environment, \name is up to $50$ times faster than the closest baseline run on a GPU. 
\end{itemize}

%% file: 2_background.tex

%% file: 3_prob.tex
\section{Preliminaries and Problem Formulation}
\label{sec:formulation}
We denote a labeled undirected graph as $\CG = (\CV, \CE, \CL)$ where $\CV$ is the node set, $\CE$ is the edge set and $\CL:\CV\cup\CE\rightarrow \Sigma$ is the labeling function over nodes and edges. $\Sigma$ is the universe of all labels and  contains a special empty label $\epsilon$. $\CL(v)$ and $\CL(e)$ denote the labels of node $v$ and edge $e$ respectively. $\CG_1\subseteq\CG_2$ denotes that $\CG_1$ is a \textit{subgraph} of $\CG_2$. 

The problem of learning \ged and \sed~\cite{ctree} is defined as follows.

\begin{prob}[Learning \ged/\sed]
\label{prob:sed}
Given a training set of tuples of the form $\langle \CG_1,\CG_2,\ged(\CG_1,\CG_2)\rangle$ (or $\CG_1,\CG_2,\sed(\CG_1,\CG_2)\rangle$), learn a neural model to predict $\ged(\CQ_1,\CQ_2)$ (or $\sed(\CQ_1,\CQ_2)$) on unseen graphs $\CQ_1$ and $\CQ_2$.
\end{prob}

For details on the exact definition of \ged and \sed, we refer to the appendix (App.~\ref{app:gedsed}). 
\begin{figure*}
    \centering
    \subfloat[Siamese architecture of \name]{
    \label{fig:siamese}
    \includegraphics[width=2in]{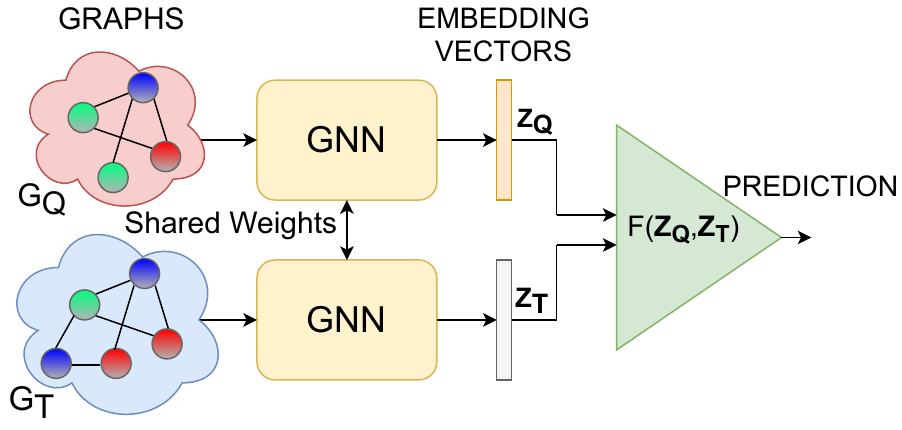}}
    \subfloat[The \gnn component in \name.]{
    \label{fig:gnn}
    \includegraphics[width=3.7in]{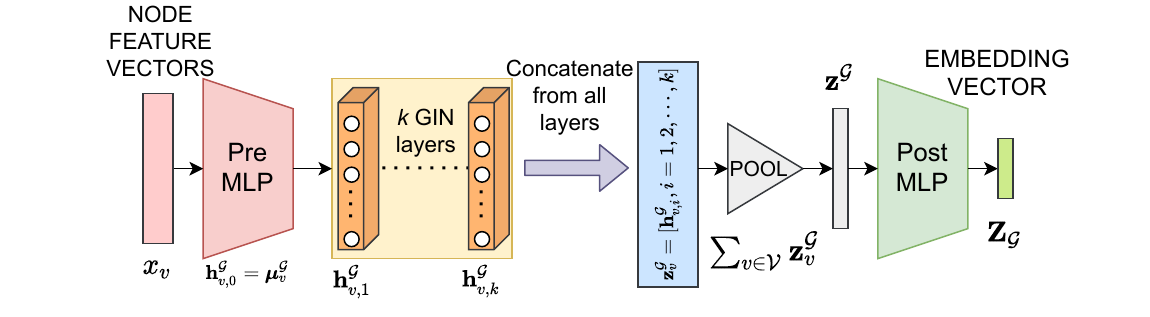}}
    \caption{\textbf{The architecture of \name.}}
    \label{fig:name}
\end{figure*}
\subsection{Properties of \ged and \sed}
\label{sec:prop_sed}
 \begin{thm}
\label{thm:sed_as_ged}
Let $\widehat{d}:\Sigma \times \Sigma \rightarrow \mathbb{R}^+_0$ be a distance function over $\Sigma$, where (i) $\widehat{d}(\ell_1, \ell_2) = 0$ if $\ell_1 = \epsilon$, and 
(ii) $\widehat{d}(\ell_1, \ell_2) = d(\ell_1, \ell_2)$ otherwise; the following holds:
$\sed(\CG_1, \CG_2) = \widehat{\ged}(\CG_1, \CG_2)$, where $\widehat{\ged}$ denotes \ged with $\widehat{d}$ as the label set distance function. In simple words, the \sed between two graphs is equivalent to \ged with a label set distance function where we ignore insertion costs.
\end{thm}
\textsc{Proof.} See App.~\ref{app:sedti}.
\begin{obs} 
\label{obs:metric}
\ged satisfies the triangle inequality if the distance function $d$ over label set $\Sigma$ satisfies the triangle inequality~\citep{ctree}. As defined in the paragraph following Def.~\ref{def:gedmap}, $d$ satisfies the triangle inequality~\citep{ctree}. Furthermore, it is trivial to see that \ged is symmetric, non-negative and satisfies identity as long as $d$ satisfies the analogues. Hence, \ged with distance function $d$ is metric.
\end{obs} 
\begin{thm}
\label{thm:sed_actual_triangle}
\sed is not metric due to violating the properties of symmetry and identity. However, it satisfies the triangle inequality, i.e., $\sed(\CG_1, \CG_3) \le \sed(\CG_1, \CG_2) + \sed(\CG_2,\CG_3)$.
\end{thm}
\textsc{Proof:}  See App.~\ref{app:sedti1} for details.

\begin{obs}
Computing \ged and \sed is NP-hard~\cite{stars}.
\end{obs}

Hereon, we use \ged as the illustrative distance function being modeled. The architecture trivially extends to \sed. The specific places that need separate treatment will be discussed explicitly.

%% file: 4_approach.tex
\vspace{-0.05in}
\section{\name: The Proposed Architecture}
\vspace{-0.10in}
Fig.~\ref{fig:name} presents the architecture of \name. The input to our learning framework is a pair of graphs $\CG_{\CQ}$ (query), $\CG_{\CT}$ (target) along with the supervision data $\ged(\CG_Q,\CG_T)$ (or $\sed(\CG_Q,\CG_T)$). Our objective is to train a model that can predict \ged on unseen query and target graphs. The design of our model must be cognizant of the fact that computing \ged is NP-hard and high quality training data is scarce. Thus, we use a \textit{Siamese} architecture~\cite{bromley1993signature}, where there are \textit{two} networks with \textit{shared} parameters applied to two inputs independently to compute representations. 
\vspace{-0.05in}
\subsection{Siamese Graph Neural Network} 
\vspace{-0.05in}
As depicted in Fig.~\ref{fig:siamese}, we use a siamese graph neural network (\gnn) with shared parameters to embed both $\CG_{\CQ}$ and $\CG_{\CT}$. While one could use two different \gnn models for the query and the target, this design increases the model parameters and consequently, the training time. Furthermore, an architecture with higher number of parameters also requires larger amount of training data, which is difficult due to \ged being NP-hard. 

Fig.~\ref{fig:gnn} focuses on the \gnn component of \name. We next discuss each of its individual components.

\noindent
\textbf{Pre-\mlp:} \rev{$\cx_v$ in Fig.~\ref{fig:gnn} is a one-hot encoding of the categorical node labels. The dimension of the one-hot vector increases linearly with the number of labels in a graph database and hence can be very large. The primary-job of the pre-mlp is to reduce it to a desirable dimension size through the operation $\boldsymbol{\mu}^{\CG}_{v} = \mlp \left ( \cx_v \right)$. Indeed, a similar effect may also be obtained by directly feeding the one-hot encoding to the first layer of \gin. However, since a \gin constructs embeddings by incorporating both structure and label information, one may desire a different dimensionality in the \gin layers. Hence, the pre-mlp is motivated more from a conceptual separation of its task to that of \gin rather than a purely performance point of view (See App.~\ref{app:ablation}).} We do not explicitly model edge labels in our experiments. \name can easily be extended to edge labels by using \gine~\citep{hu2019strategies} instead of \gin.
\looseness=-1

\noindent
\textbf{Graph Isomorphism Network (\gin): }\gin \citep{gin} consumes the information from the Pre-\mlp to learn hidden representations that encode both the graph structure as well as the node feature information. \gin is as powerful as the \textit{Weisfeiler-Lehman (WL) graph isomorphism test} \citep{leman1968reduction} in distinguishing graph structures. Since our goal is to accurately characterize graph topology and learn similarity, \gin emerges as the natural choice. \gin develops its expressive power by using an \textit{injective} aggregation function. Specifically, in the initial layer, each node $v$ in graph $\CG$ is characterized by the representation learned by the \mlp, i.e., $\ch^{\CG}_{v,0}=\boldsymbol{\mu}_v^{\CG}$. Subsequently, in each hidden layer $i$, we learn an embedding through the following transformation.
\vspace{-0.05in}
\begin{equation}
\label{eq:gin}
    \ch^{\CG}_{v,i}=\mlp\left((1+\epsilon^i)\cdot\ch^{\CG}_{v,i-1}+\sum_{u\in \CN_{\CG}(v)}\ch^{\CG}_{u,i-1}\right)
\end{equation}
\vspace{-0.10in}

Here, $\epsilon^i$ is a layer-specific learnable parameter, $\CN_{\CG}(v)$ is one-hop neighbourhood of the node $v$, and $\ch^{\CG}_{v,0}=\boldsymbol{\mu}^{\CG}_{v}$. The $k$-th layer embedding is $\ch^{\CG}_{v,k}$, where $k$ is final hidden layer. 

\noindent
\textbf{Concatenation, Pool and Post-\mlp: }Intuitively, $\ch^{\CG}_{v,i}$ captures a feature-space representation of the $i$-hop neighborhood of $v$. Typically, \gnns operate on node or edge level predictive tasks, such as node classification or link prediction, and hence, the node representations are passed through an \mlp for the final prediction task. In our problem, we need to capture a graph level representation. Furthermore, the representation should be rich enough to also capture the various subgraphs within the input graph so that \sed can be predicted accurately. To fulfil these requirements, we first \textit{concatenate} the representation of a node across \textit{all} hidden layers, i.e., the final node embedding is $\cz_v^{\CG}=\textsc{Concat} \left(\ch^{\CG}_{v,i}, \forall i \in \{1,2,\cdots, k\} \right)$. This allows us to capture a multi-granular view of the subgraphs centered on $v$ at different radii in the range $[1,k]$. Next, to construct the graph-level representation, we perform a sum-pool, which adds the node representations to give a single vector. This information is then fed to the Post-\mlp to enable post-processing. Mathematically:
\vspace{-0.05in}
\begin{equation}
    \CZ_{\CG}= \mlp(\cz^{\CG})=\mlp\left(\sum_{v \in \CV} \cz_v^{\CG}\right)
\end{equation}
\noindent
\textbf{\ged and \sed Prediction:} The final task is to predict the \ged (and \sed) as a function of query graph embedding $\CZ_{\CG_{\CQ}}$ and target graph embedding $\CZ_{\CG_{\CT}}$. The natural choice would be to feed these embeddings into another \mlp to learn $\ged(\CZ_{\CG_{\CQ}},\CZ_{\CG_{\CT}})$. This \mlp can then be trained jointly with the graph embedding model in an \textit{end-to-end} fashion. However, an \mlp prediction does not have any theoretical guarantees with respect to the preservation of metric properties of \ged and the triangle inequality of \sed. We, therefore, focus on learning prediction functions $\CF_g\left(\CZ_{\CG_{\CQ}}, \CZ_{\CG_{\CT}}\right)$ and $\CF_s\left(\CZ_{\CG_{\CQ}}, \CZ_{\CG_{\CT}}\right)$ for \ged and \sed respectively, such that they are accurate and respects the desirable properties from the original graph space. As we will empirically substantiate in \S~\ref{sec:ablation}, the inductive bias injected through the prediction functions also lead to more effective learning over low volumes of training data than an MLP.

\vspace{-0.05in}
\subsubsection{\ged}
\vspace{-0.05in}
We require the following four properties to ensure that the prediction is also a metric.
\begin{alignat}{4}
\label{eq:ged_nonneg}
\CF_g\left(\CZ_{\CG_{\CQ}}, \CZ_{\CG_{\CT}}\right) &\ge 0\\
\label{eq:ged_identity}
\CF_g\left(\CZ_{\CG_{\CQ}}, \CZ_{\CG_{\CT}}\right) &= 0 \iff \forall i: \CZ_{\CG_{\CQ}}[i] = \CZ_{\CG_{\CT}}[i]\\
\label{eq:symmetry}
\CF_g\left(\CZ_{\CG_{\CQ}}, \CZ_{\CG_{\CT}}\right) &=\CF_g\left(\CZ_{\CG_{\CT}}, \CZ_{\CG_{\CQ}}\right)\\
\label{eq:ged_ti}
\CF_g\left(\CZ_{\CG_{\CQ}}, \CZ_{\CG_{\CT}}\right) &\le \CF_g\left(\CZ_{\CG_{\CQ}}, \CZ_{\CG'}\right) + \CF_g\left(\CZ_{\CG'},\CZ_{\CG_{\CT}}\right)
\end{alignat}
To achieve this, we establish an important connection of metrics on vector spaces to norms. Every norm $\|.\|$ gives a metric $(\cx,\cy) \mapsto \|\cx-\cy\|$. Moreover for a metric, there exists a norm $\|.\|$ such that the metric can be expressed as $(\cx,\cy) \mapsto \|\cx-\cy\|$, \textit{iff} the metric is \textit{translation invariant} and \textit{homogeneous}. Thus, we add these properties to the desiderata for $\CF_g$:
\begin{alignat}{2}
\label{eq:ged_trans_inv}
\CF_g\left(\CZ_{\CG_{\CQ}}+\mathbf{k}, \CZ_{\CG_{\CT}}+\mathbf{k}\right) &= \CF_g\left(\CZ_{\CG_{\CQ}}, \CZ_{\CG_{\CT}}\right), \forall \mathbf{k} \in \mathbb{R}^d\\
\label{eq:ged_homogeneous}
\CF_g\left(r\CZ_{\CG_{\CQ}}, r\CZ_{\CG_{\CT}}\right) &= |r|\CF_g\left(\CZ_{\CG_{\CQ}}, \CZ_{\CG_{\CT}}\right), \forall r \in \mathbb{R}
\end{alignat}
Armed with these observations, we define the class of functions that may be used for $\CF_g$.
\begin{obs}
\label{obs:fg_charac}
$\CF_g$ may be defined as any function $(x,y) \mapsto \|x-y\|$ for some norm $\|.\|$ over the vector space $\mathbb{R}^d$ such that $\CF_g$ satisfies Eqs.~\ref{eq:ged_trans_inv} - \ref{eq:ged_homogeneous}.
\end{obs}
The $L_p$ norm satisfies Obs.~\ref{obs:fg_charac}. Hence, we define $\CF_g$ as:
\begin{equation}
\label{eq:f_g}
    \CF_g\left(\CZ_{\CG_{\CQ}}, \CZ_{\CG_{\CT}}\right) = \left\lVert  \CZ_{\CG_{\CQ}}-\CZ_{\CG_{\CT}}\right\rVert_p
\end{equation}
In our implementation we use the $L_2$ norm. Finally, the parameters of the entire model are learned by minimizing the mean squared error 
 (here $\mathbb{T}$ is the training set).
\vspace{-0.05in}
\begin{equation}
\label{eq:original_loss}
    \mathscr{L}=\frac{1}{\lvert\mathbb{T} \rvert}\sum_{\forall \langle \CG_{\CQ}, \CG_{\CT} \rangle\in\mathbb{T}} \left(\CF_g\left(\CZ_{\CG_{\CQ}}, \CZ_{\CG_{\CT}}\right)-\ged\left(\CG_{\CQ},\CG_{\CT}\right)\right)^2
\end{equation}
\textbf{Intuition:} Regardless of the graph representations generated by our model, $\CF_g$ ensures that the predicted distance is a metric. 
One the other hand, by training the model to produce embeddings $\CZ_{\CG_{\CQ}}$ and $\CZ_{\CG_{\CT}}$ such that $\CF_g\left(\CZ_{\CG_{\CQ}}, \CZ_{\CG_{\CT}}\right)\approx\ged\left(\CZ_{\CG_{\CQ}}, \CZ_{\CG_{\CT}}\right)$, we enforce a rich structure on the embedding space such that $\CF_g$ is also accurate. Thus,  $\CF_g$ injects an inductive bias satisfying the dual needs of accuracy and preservation of original space properties.

\vspace{-0.05in}
\subsubsection{\sed} 
\vspace{-0.05in}
\sed satisfies non-negativity and triangle inequality. Following a similar reasoning as above, we define 
$\CF_s$ as follows:
\vspace{-0.10in}

{\small
\begin{equation}
\label{eq:f}
    \CF_s\left(\CZ_{\CG_{\CQ}}, \CZ_{\CG_{\CT}}\right) = \lVert ReLU(\CZ_{\CG_{\CQ}}-\CZ_{\CG_{\CT}})\rVert_2 =\left\lVert \max\left\{0, \CZ_{\CG_{\CQ}}-\CZ_{\CG_{\CT}}\right\}\right\rVert_2
\end{equation}}
\vspace{-0.10in}

Intuitively, for those co-ordinates where the value of $\CZ_{\CG_{\CQ}}$ is greater than $\CZ_{\CG_{\CT}}$, a distance penalty is accounted by $\CF_s$ in terms of how much those values differ; otherwise $\CF_s$ considers $0$. This follows the intuition that the \sed accounts for those features of $\CG_{\CQ}$ that are not in $\CG_{\CT}$. Moreover, consistent with \sed, the additional features in $\CG_{\CT}$ that are not in $\CG_{\CQ}$, do not incur any cost. 
\begin{lem}
\label{thm:inductive_bias_sed}
The following properties hold on predicted \sed.  
\vspace{0.05in}
\begin{enumerate}
\item $\CF_s\left(\CZ_{\CG_{\CQ}}, \CZ_{\CG_{\CT}}\right) \ge 0$ 
\item $\CF_s\left(\CZ_{\CG_{\CQ}}, \CZ_{\CG_{\CT}}\right) = 0 \iff \CZ_{\CG_{\CQ}} \le \CZ_{\CG_{\CT}}$
\item $\CF_s\left(\CZ_{\CG_{\CQ}}, \CZ_{\CG_{\CT}}\right)\le \CF_s(\CZ_{\CG_{\CQ}}, \CZ_{\CG_{\CT'}}) + \CF_s(\CZ_{\CG_{\CT'}}, \CZ_{\CG_{\CT}})$ 
\end{enumerate}
\end{lem}
\textsc{Proof.} Properties (1) and (2) follow from the definition of $\CF$ itself. Property (3) follows from the fact that we take the $L_2$ norm. Formally, we state it as follows.
\begin{lem}
\label{lem:ti_for_f}
$\CF_s\left(\CZ_{\CG_{\CQ}}, \CZ_{\CG_{\CT}}\right)\le \CF_s(\CZ_{\CG_{\CQ}}, \CZ_{\CG_{\CT'}}) + \CF_s(\CZ_{\CG_{\CT'}}, \CZ_{\CG_{\CT}})$, where $\CF_s(\cx,\cy)=\|\relu(\cx-\cy)\|$ for any monotonic norm $\|.\|$.
\end{lem}
\textsc{Proof.} See App.~\ref{app:sed_embed_ti}.

\vspace{-0.05in}
\subsection{Characterization of \name}
\vspace{-0.10in}
\textbf{Importance of pair-independence and siamese architecture:}
A pair-independent siamese architecture enables theoretical guarantees for \name by constraining the model to learn a single mapping from graph space to embedding space for both query and target. Despite these restrictions, \name outperforms prior works which freely use cross-graph information and don't provide theoretical guarantees. This further confirms that the siamese architecture is a useful prior. \\
\textbf{Complexity Analysis:} The complexity of \ged and \sed inference in \name is \textit{linear} in the number of nodes and edges in the query and target graphs (See App.~\ref{app:complexity} for derivation). This computation cost is drastically lower than the factorial computation cost of optimal \ged and \sed. With respect to neural methods for graph similarity~\citep{simgnn,graphsim,icmlged,cvpr, h2mn}, all have at least quadratic computation cost, i.e., $O(|\CV|^2)$.\\
\textbf{Indexing Embeddings:} Since the generated embeddings for both \ged and \sed satisfy triangle inequality, they are indexable leading to fast querying times. We develop an index structure to exploit this property. Due to space limitations, the details are included in App.~\ref{app:index}. In addition, we also design a \textit{neighborhood decomposition} scheme, which enables fast pruning of the exponential search space~\S~\ref{sec:scaling}. In \S~\ref{sec:efficiency}, we empirically analyze the impact of index structures on querying time.

%% file: 5_experiment.tex
\section{Empirical Evaluation}
\label{sec:experiments}
In this section, we establish the following:
\begin{itemize}
    \item \textbf{Efficacy:} \name is more accurate than the state of the art approaches for both \ged and \sed. 
    \item \textbf{Efficiency:} \name is orders of magnitude faster than existing approaches and scales well to graphs with millions of nodes.
    \item{\bf Scalability:} Pair-independence and indexability further enhances the scalability of \name and enables it to be run on CPU-only platforms.
\end{itemize}
\noindent
Our code base and datasets are available at 
\url{https://github.com/idea-iitd/greed}.
\subsection{Experimental Setup}
\label{sec:exp_setup}
We use a machine with an Intel Xeon Gold 6142 processor and GeForce GTX 1080 Ti GPU for all our experiments. \\
\textbf{Datasets:} Table~\ref{tab:datasets} lists the datasets used for benchmarking. Further details on the dataset semantics are provided in the App.~\ref{app:datasets}. We include a mixture of both \textit{graph databases} (\#graphs >1), as well as \textit{single large} graphs (\#graphs = 1). \linux and \imdb are unlabeled. We note that this is the first study to evaluate neural graph distance approaches on million-scale graphs.\\
\noindent
\textbf{Baselines:} To evaluate performance in \textbf{\ged}, we compare with \simgnn
~\citep{simgnn}, \genn~\citep{cvpr}, \textsc{H$^2$MN}\citep{h2mn} and \rev{\gotsim~\cite{graphotsim}}. These are the most recent neural frameworks and have shown better efficacy than other neural approaches such as \simgnn~\cite{simgnn}, \textsc{GraphSim}~\cite{graphsim}, and GMN~\cite{icmlged}. 

 For \textbf{\sed}, no neural approaches exist. However, \textsc{H$^2$MN} and \simgnn can be trained by replacing \ged with \sed along with minor modifications in training. \nsc~\cite{nsc} is a method for counting subgraphs using graph embeddings. Since this is a related operation, we use \nsc as a baseline by changing the loss function to minimize the RMSE between true and predicted \sed. We also use  \neuromatch~\citep{neuromatch} as a baseline, which was originally designed to detect subgraph isomorphism. While \neuromatch cannot predict \sed, it generates a \textit{violation score}, which can be interpreted as the likelihood of the query being subgraph isomorphic to the target. The violation score can be used as a proxy for \sed and used in ranking of $k$-NN ($k$-NearestNeighbour) queries. Thus, \neuromatch comparisons are limited to $k$-NN queries on \sed. 
 GMN~\citep{icmlged}, \textsc{GraphSim}~\citep{graphsim}, \rev{\gotsim~\cite{graphotsim}} and \genn are not included since they cannot be easily adapted for \sed. See App.~\ref{app:baseline} for details. 
 
In the \textbf{non-neural} category, we use \textit{mixed integer programming} based method \f~\citep{LEROUGE2017254} with a time bound of $0.1$ seconds per pair for both \ged and \sed. \f provides the optimal solution given infinite time. We also compare with \branch ~\citep{blumenthal2019new}, which achieves an excellent trade-off between accuracy and time~\citep{blumenthal2020comparing}. \branch uses \textit{linear sum assignment problem with error-correction} (\textsc{Lsape}) to process the search space. We use \textsc{Gedlib}'s ~\citep{blumenthal2019gedlib} implementation of these methods.\\
\noindent
\textbf{Training (and Test) Data Generation: } For \ged, we use $\langle query, target\rangle$ graph pairs from \imdb, \aids, and \linux. Our setup is identical to \simgnn~\citep{simgnn} and \hmnr~\citep{h2mn}. 

For \sed, the target graphs are taken from datasets listed in Table~\ref{tab:datasets}. For the query graph, in \aidsg, we use known \textit{functional groups}~\cite{graphsig_jcim}. In the rest of the graph datasets, queries are sampled by performing a random BFS traversal (depth up to 5). Table~\ref{tab:datasets} shows the average query sizes ($|\CV_{\CQ}|$, $|\CE_{\CQ}|$).
 We use \textit{mixed integer programming} method \textsc{F2}~\citep{LEROUGE2017254} implemented in \textsc{Gedlib}~\citep{blumenthal2019gedlib} with a large time limit to generate ground-truth data.\\
\noindent
\textbf{Train-Validation-Test:} We use $100K$ query-target pairs for training and $10K$ pairs each for validation and test. All models are trained till validation loss is minimized or there is less than $0.05\%$ change in validation loss over a number of extended epochs. 
For \name, we set the number of layers in \gin to $8$. The hidden layer dimension is set to $64$. For all baselines, we use the default parameters suggested by the authors.

\begin{table}[t]
    \centering
\begin{subtable}{.28\textwidth}
\scalebox{0.65}{
\begin{tabular}{rrrr}
    \toprule
        \textbf{Methods} & \aids & \linux & \imdb \\ 
        \midrule
        \textbf{\name} & \textbf{0.796}& 0.415& \textbf{6.734} \\
        \textbf{\hmnr} & 0.994& 0.734& 86.077 \\
        \textbf{\genn} & 0.907& \textbf{0.267}& NA \\
        \rev{\textbf{\gotsim}} & \rev{0.996}& \rev{0.574}& \rev{37.831} \\
        \textbf{\simgnn} & 1.037& 0.666& 66.250 \\
        \textbf{Branch} & 3.322& 2.474& 6.875 \\
        \textbf{MIP-F2} & 2.929& 1.245& 82.124 \\
        \bottomrule
\end{tabular}}
\caption{Prediction of \ged}
\label{tab:rmse_ged}
\end{subtable}\hfill
\begin{subtable}{.63\textwidth}
\scalebox{0.68}{
\begin{tabular}{cccccccc}
    \toprule
\textbf{Methods} &\dblp &\amz &\pubmed &\seer &\cora &\protein &\aidsg \\
\midrule
\textbf{\name} &\textbf{0.964} &\textbf{0.495} &\textbf{0.728} &\textbf{0.519} &\textbf{0.635} &\textbf{0.524} &\textbf{0.512} \\
\textbf{\hmnr} &1.470 &1.294 &1.213 &1.502 &1.446 &0.941 &0.749 \\
\textbf{\nsc} & NA & 2.141& 1.095 &1.66 & 1.661&0.662& 0.562 \\
\textbf{\simgnn} &1.482 &2.810 &1.322 &1.781 &1.289 &1.223 &0.696 \\
\textbf{Branch} &2.917 &4.513 &2.613 &3.161 &3.102 &2.391 &1.379 \\
\textbf{MIP-F2} &3.427 &5.595 &3.399 &4.474 &3.871 &2.249 &1.537 \\
\bottomrule
\end{tabular}}
\caption{Prediction of \sed.}
\label{tab:rmse_sed} 
\end{subtable}
    \caption{\textbf{(a) Datasets. (b-c) RMSE scores (lower is better) in (a) \ged and (b) \sed. \genn does not scale on graphs beyond $10$ nodes and hence results in \imdb are not reported. \nsc does not scale in \dblp due to memory consumption.}}
    \label{tab:one}
\end{table}

\subsection{Prediction Accuracy of \sed and \ged}
\label{sec:rmse}
Tables~\ref{tab:rmse_ged} and~\ref{tab:rmse_sed} present the accuracy of all techniques on \ged and \sed in terms of \textit{Root Mean Square Error (RMSE}). 
\name outperforms all other techniques in 9 out of 10 settings across \ged and \sed. While \hmnr and \nsc are the second best performers in \sed, \genn performs well in \ged. \genn, however, is extremely slow and does not scale on graphs of size beyond $10$. \hmnr, thus, provides the second best balance between efficacy and efficiency after  \name.
  The gap in accuracy is the highest in \imdb for \ged, where \name is more than $10$ times better than the neural baselines. \imdb graphs are significantly denser and larger than \aids or \linux. Thus, computing the optimal \ged is harder. While all techniques have  higher errors in \imdb, the deterioration is more severe in the baselines indicating that \name scales better with graph sizes.\\
  \begin{figure}[t]
\centering
   \captionsetup[subfigure]{labelfont={normalsize,bf},textfont={normalsize}}
    \subfloat[\name, \imdb]{\includegraphics[width=1.07in]{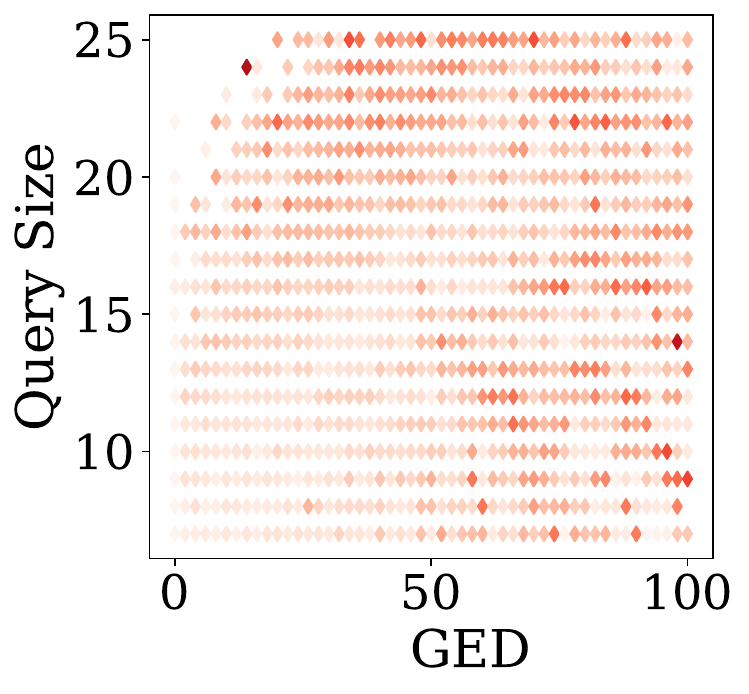}}
    \subfloat[\hmnr, \imdb]{\includegraphics[width=1.01in]{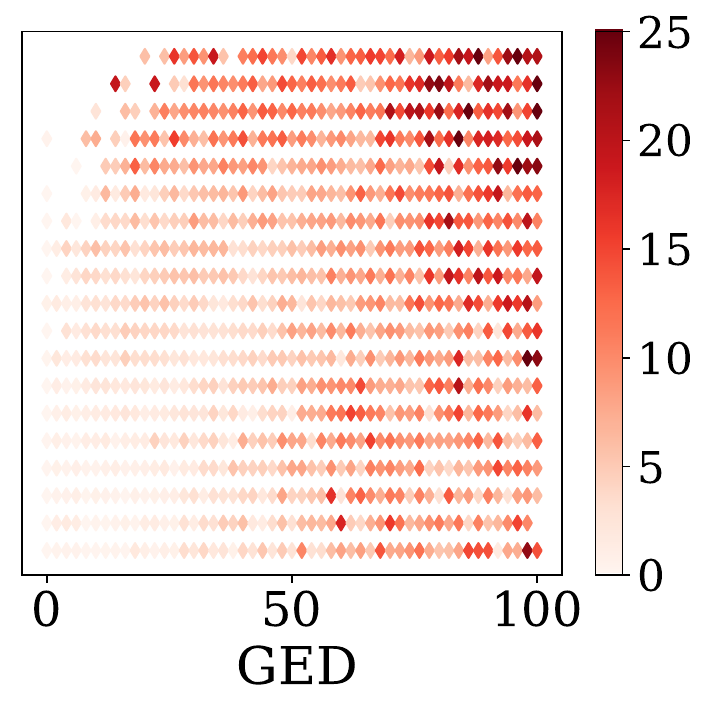}}
\subfloat[\name, \dblp]{
\includegraphics[width=1.05in]{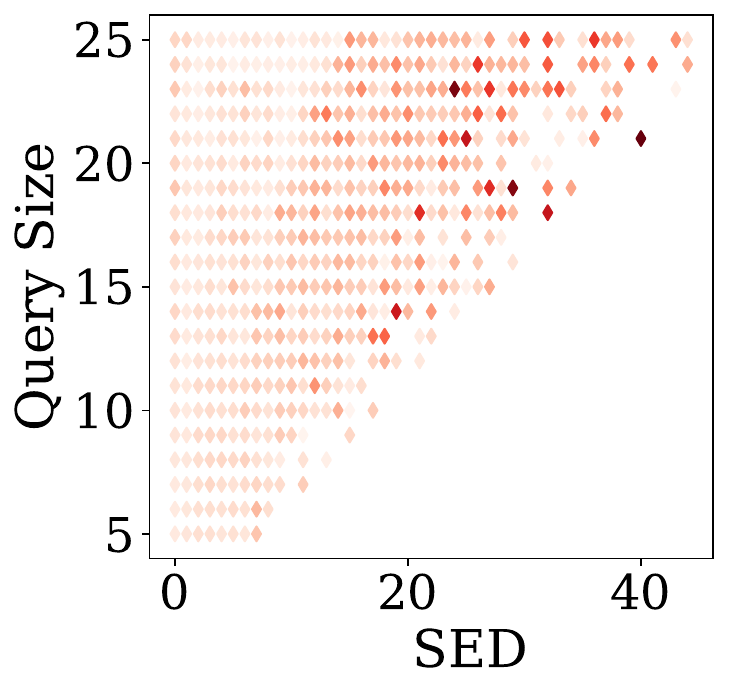}}
\subfloat[\hmnr, \rev{\dblp}]{
\includegraphics[width=0.995in]{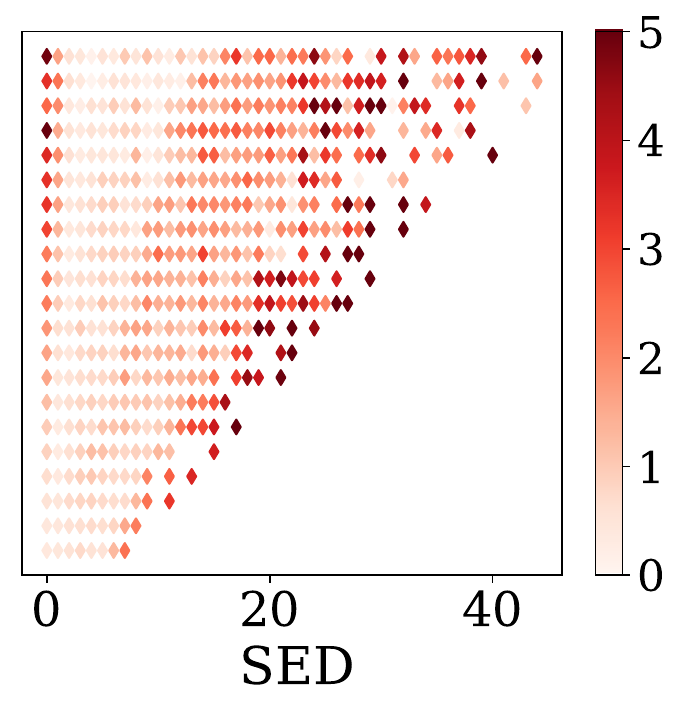}}
{
\caption{\textbf{Heat Map of RMSE in (a-b) \ged and (c-d) \sed against query size in \imdb and \dblp.}}
\label{fig:heatmap}}
\end{figure}
 \noindent
\textbf{Impact of Query Size: } We next investigate how the accuracy varies against the query size. Intuitively, the task gets harder with query size since the combinatorial space of possible maps increases exponentially. For this analysis, we compare \name with \hmnr in \imdb and \dblp for \ged and \sed respectively. \genn fails to scale on both datasets.

\begin{figure}[t]
    \centering
    \captionsetup[subfigure]{labelfont={normalsize,bf},textfont={normalsize}}
    \subfloat[\pubmed]{\label{fig:sed_range_pubmed}\includegraphics[height=0.16\textwidth]{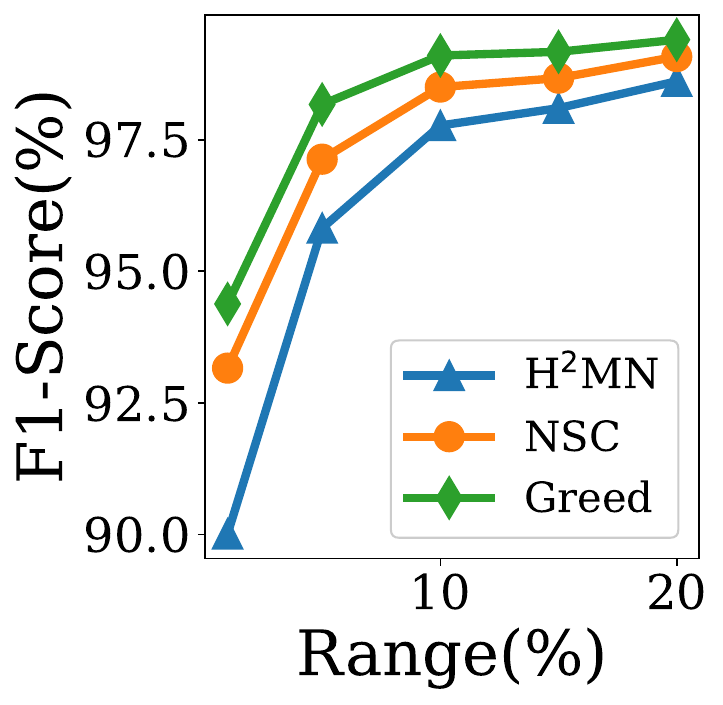}}
    \subfloat[\seer]{\label{fig:sed_range_cite}\includegraphics[height=0.16\textwidth]{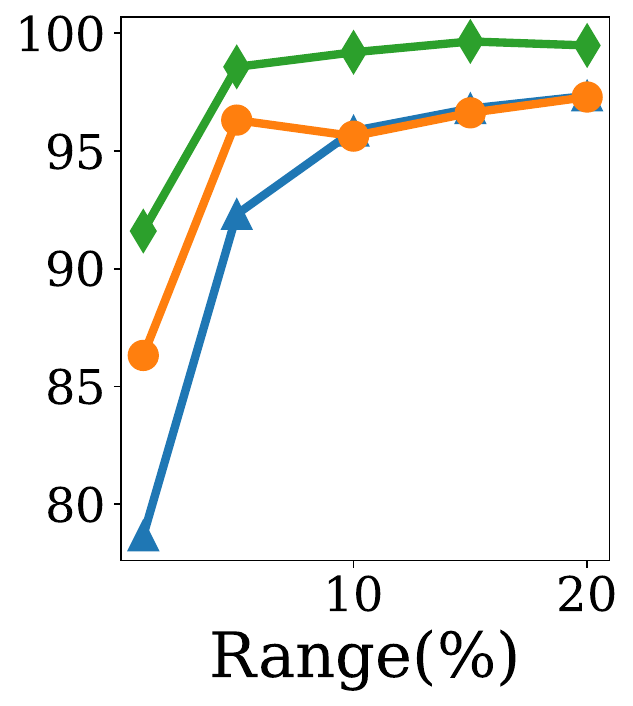}}
    \subfloat[\cora]{\label{fig:sed_range_cora}\includegraphics[height=0.16\textwidth]{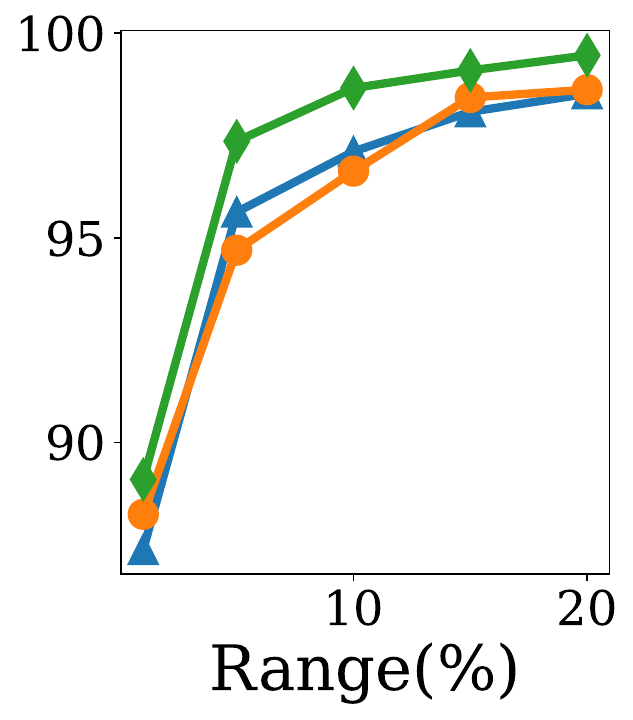}}
    \subfloat[\protein]{\label{fig:sed_range_protein}\includegraphics[height=0.16\textwidth]{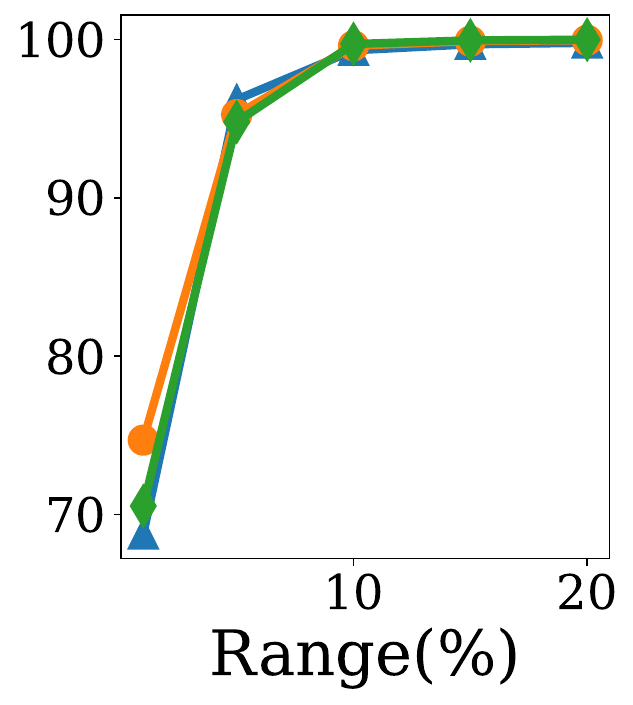}}
    \subfloat[\aidsg]{\label{fig:sed_range_aids}\includegraphics[height=0.16\textwidth]{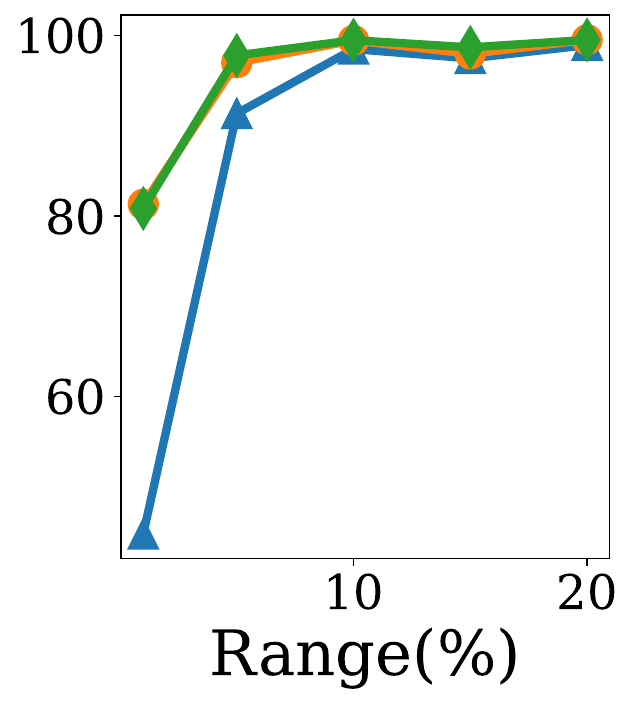}}\hfill\\
        \subfloat[\aids]{\label{fig:ged_range_aids}\includegraphics[height=0.13\textwidth]{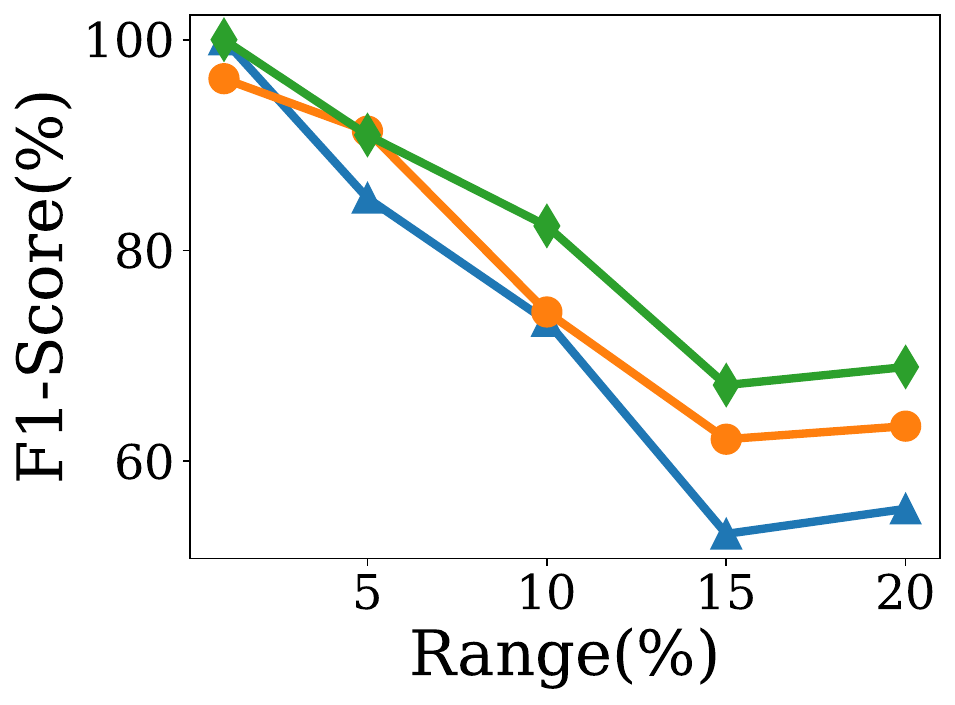}}
       \subfloat[\linux]{\label{fig:ged_range_linux}\includegraphics[height=0.13\textwidth]{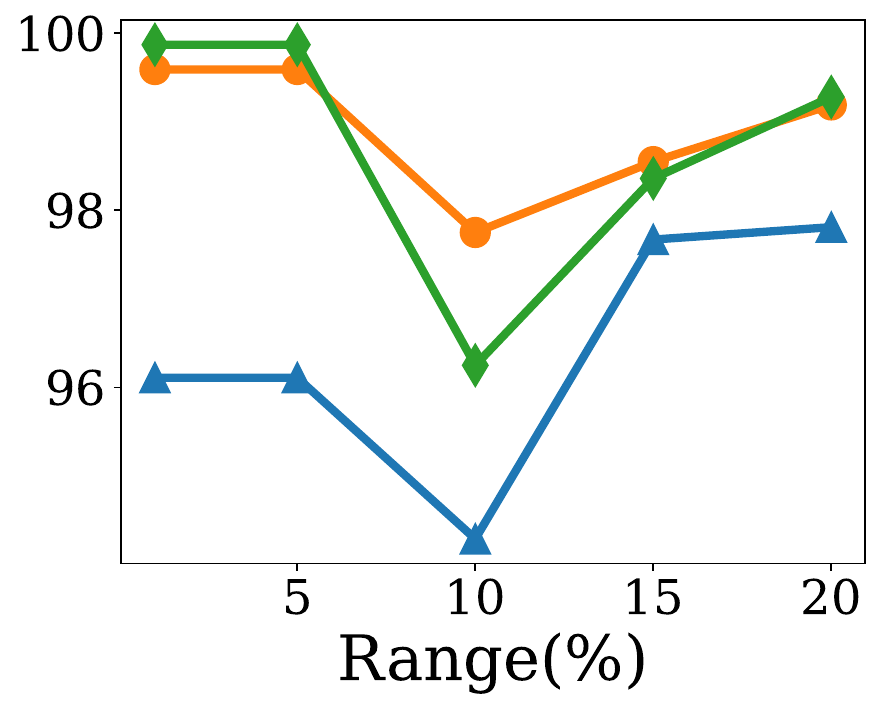}} 
    \subfloat[\imdb]{\label{fig:ged_range_imdb}\includegraphics[height=0.13\textwidth]{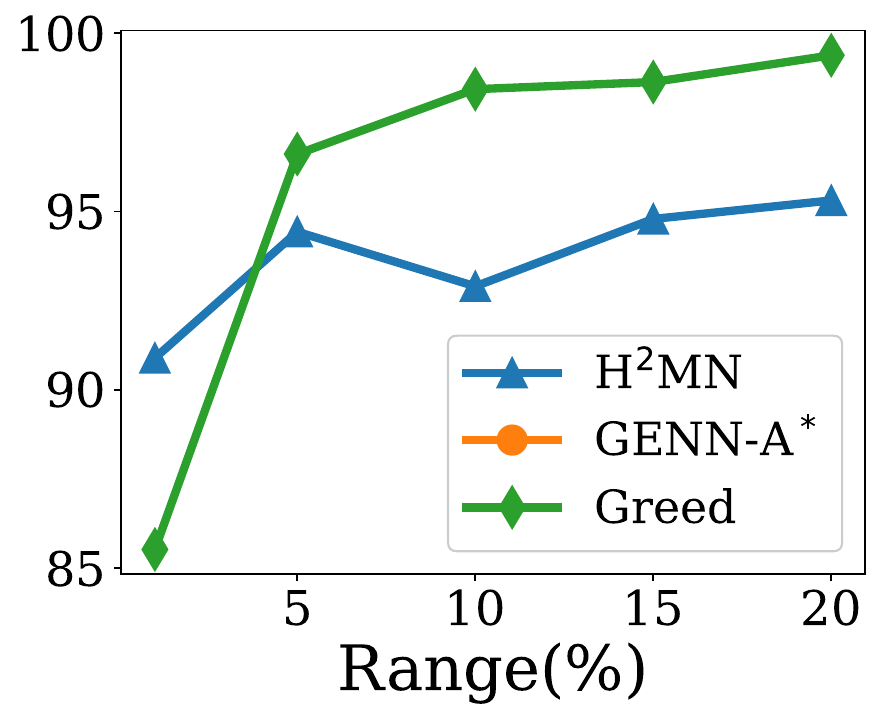}}
     \subfloat[\dblp]{\label{fig:ablation_dblp}\includegraphics[height=0.13\textwidth]{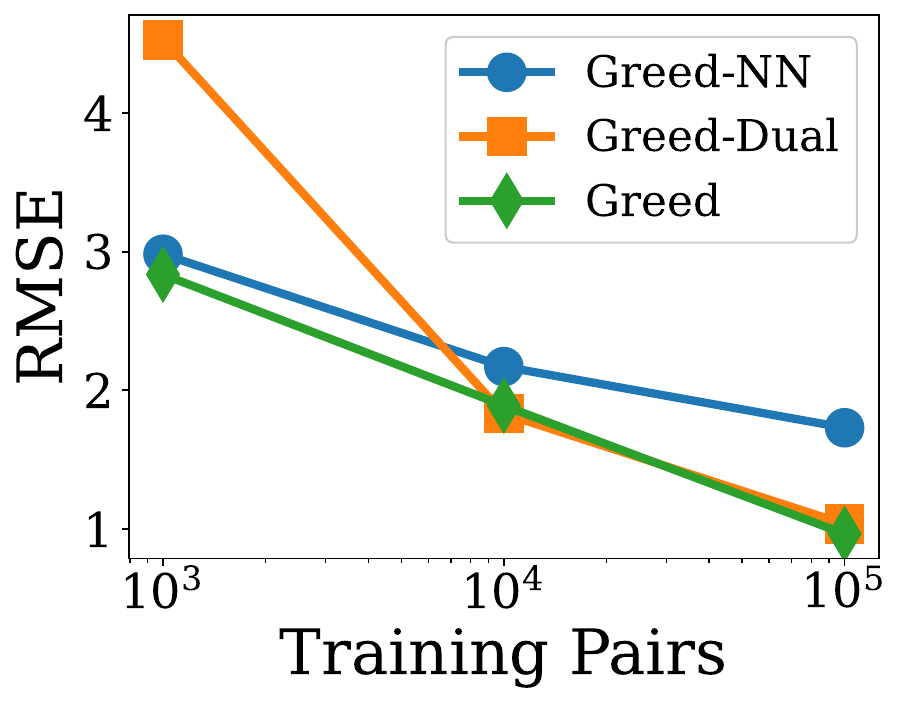}}
    \subfloat[\pubmed]{\label{fig:ablation_pubmed}\includegraphics[height=0.13\textwidth]{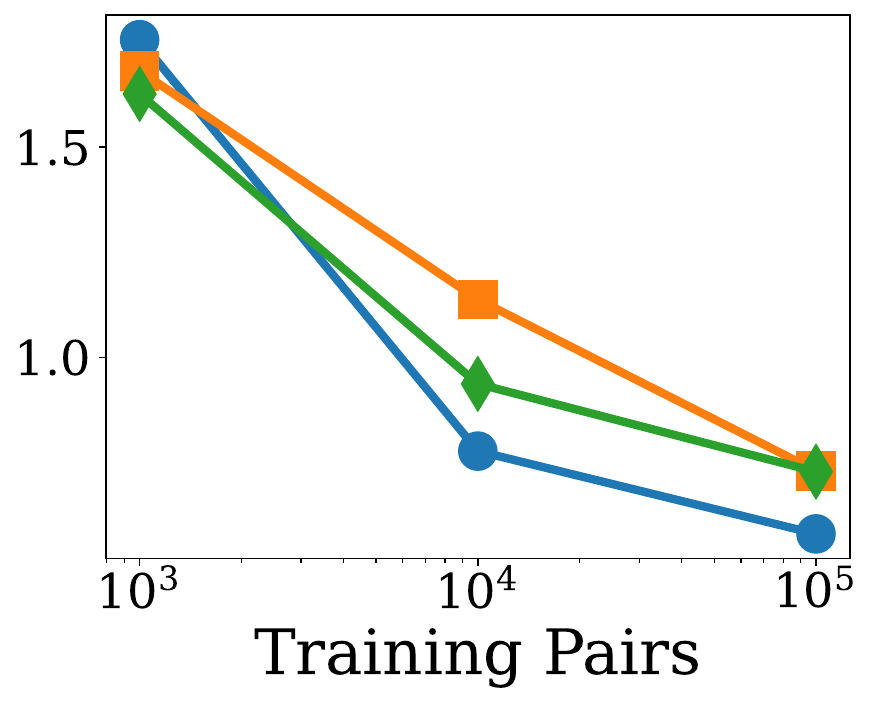}}
    \subfloat[\seer]{\label{fig:ablation_cite}\includegraphics[height=0.13\textwidth]{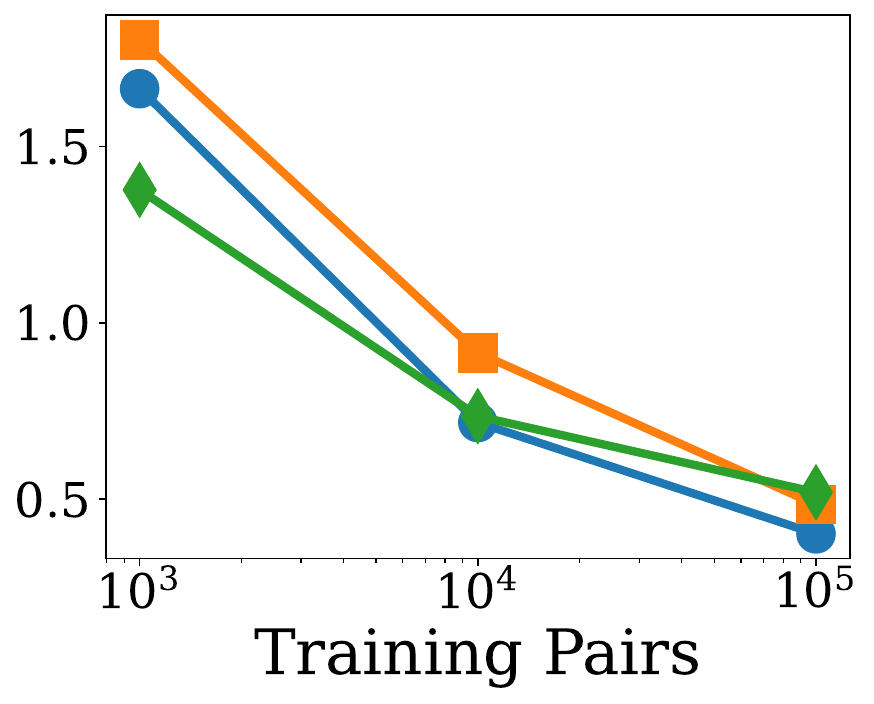}}\hfill\\
    \caption{\textbf{F1-score in range queries on \sed (a-e) and \ged (f-h). The range threshold is set as a percentage of the max distance observed in the test set. The legend for Figs.~(a)-(e) is provided in (a) and for (f-h) is provided in (h). (i-k) Ablation study to analyze the impact of siamese architecture and function $\mathcal{F}$. The legend for Figs.~(i)-(k) is provided in (i).}}
    \label{fig:query_vis_tight}
\end{figure}

In Fig.~\ref{fig:heatmap}, we plot the \textit{heat map} of RMSE against query graph size. In this plot, each dot corresponds to a query graph $\CG_{\CQ}$. The co-ordinate of a query is $(\ged(\CG_{\CQ}, \CG_{\CT}),|\CV_{\CQ}|)$ (analogously defined for \sed). The color of a dot represents the RMSE; the darker the color, the higher is the RMSE. When we compare the heat maps of \name with \hmnr, we observe that \hmnr is noticeably darker. Furthermore, the concentration of dark colors is noticeably higher on the upper-right corner indicating deterioration with larger query sizes and higher distance values. This indicates that \name scales better with query sizes and distances.\\
\textbf{Visualization: } A case study to visually illustrate the efficacy of \name is provided in App.~\ref{app:visualization}.\\
 \begin{table}[t]
\centering
\begin{subtable}{1.8in}
\scalebox{0.75}{
\begin{tabular}{cccc}
    \toprule
        \textbf{Methods} & \aids & \linux & \imdb \\ 
        \midrule
        \textbf{$\name$} &\textbf{0.80} &0.89 &\textbf{0.87}\\
        \textbf{\hmnr}  & 0.74 &0.88 &0.80\\
        \textbf{\genn}  & 0.75 &\textbf{0.90} &NA\\
        \textbf{\simgnn}  &0.72 &0.86 &0.67\\
        \bottomrule
\end{tabular}}
\caption{Ranking in \ged.}
\label{tab:ranking_ged_KT}
\end{subtable}
\begin{subtable}{2.8in}
\scalebox{0.57}{
\label{tab:ranking_SED_KT}
   \begin{tabular}{rrrrrr}
    \toprule
        \textbf{Methods} &\pubmed &\seer &\cora &\protein &\aidsg \\
        \midrule
\textbf{\name} &\textbf{0.90} &\textbf{0.90} &\textbf{0.91} &\textbf{0.75} &\textbf{0.80} \\
\textbf{\hmnr} &0.87 &0.88 &0.88 &0.70 &0.72 \\
\textbf{\nsc} & 0.89& 0.88&0.88 &0.74 & 0.78\\
\textbf{\simgnn} &0.85 &0.87 &0.86 &0.63 &0.73 \\
\textbf{\neuromatch} &0.70 &0.75 &0.73 &0.57 &0.59 \\
        \bottomrule
\end{tabular}}%
\caption{Ranking in \sed.}
\label{tab:ranking_sed_KT}  
    \end{subtable}
    \caption{\textbf{Kendall's tau scores (higher is better).}}
\end{table}
\textbf{Range and $\mathbf{k}$-NN queries:} 
To quantify performance in range and $k$-NN queries, we measure \textit{F1-score (Range query)} and \textit{Kendalls's tau ($k$-NN)}~\citep{kendall} of the predicted answer set, when compared against the ground truth. In Figs.~\ref{fig:sed_range_pubmed}-\ref{fig:ged_range_imdb}, we measure the performance in range queries. In \sed, \name consistently outperforms all baselines in F1-score. In \ged, the trend remains similar. Although, \genn outperforms \name for a brief region in \linux, overall, \name has the highest F1-score. We also note the \genn is not included in Fig.~\ref{fig:ged_range_imdb} since it fails to scale on \imdb. In $k$-NN queries (Tables~\ref{tab:ranking_ged_KT} and~\ref{tab:ranking_sed_KT}), \name outperforms all algorithms in \sed. In \ged, similar to the trend in range queries, \name is the dominant method and \genn marginally outperforms \name in \linux. 

\subsection{Efficiency}
\label{sec:efficiency}
Tables~\ref{tab:prediction_ged_RT}-\ref{tab:prediction_sed_RT} present the inference times per $10K$ query-target pairs. In this experiment, we do not index embeddings by \name so that the comparison unearths the raw difference in computation efficiency  of solely the neural architectures. As visible, \name is up to $1800$ times faster than the non-neural baselines and up to $10$ to $20$ times faster than \hmnr, the current state of the art in \ged prediction. Also note that \genn is exorbitantly slow (Table~\ref{tab:prediction_ged_RT}). \genn is slower since it not only predicts the \ged but also the alignment via an A$^*$ search. While the alignment information is indeed useful, computing this information across all graphs in the database may generate redundant information since an user is typically interested only on a small minority of graphs that are in the answer set. In App.~\ref{app:alignment}, we discuss this issue in detail.

\begin{table*}[h!]
\centering
\begin{subtable}{1.9in}
\scalebox{0.75}{
\begin{tabular}{rrrr}
    \toprule
        \textbf{Methods} & \aids & \linux & \imdb  \\ 
        \midrule
        \textbf{$\name$} & \textbf{0.49}& \textbf{0.70}& \textbf{0.63} \\
        \textbf{\hmnr} & 9.50& 8.74& 8.83 \\
        \textbf{\genn} & 12190& 1340& NA \\
        \textbf{\branch} & 10.70& 8.24& 127.90 \\
        \textbf{\f} & 593.34& 191.88& 1173.548 \\
        \bottomrule
\end{tabular}}
\caption{\ged}
\label{tab:prediction_ged_RT}
\end{subtable}
\begin{subtable}{3.5in}
\scalebox{0.65}{
   \begin{tabular}{rrrrrrrr}
    \toprule
        \textbf{Methods} &\dblp &\amz &\pubmed &\seer &\cora &\protein &\aidsg \\
\midrule
\textbf{\name} &\textbf{6.84} &\textbf{1.46} &\textbf{1.30} &\textbf{1.28} &\textbf{1.25} &\textbf{0.86} &\textbf{0.84} \\
\textbf{\hmnr} &44.68 &23.2 &25.79 &27.54 &29.04 &19.33 &9.63 \\
\textbf{\nsc} & NA &21 & 35.05 &24.46 &70.59 &21 &4 \\
\textbf{\simgnn} &109.56 &47.68 &39.80 &39.40 &40.73 &39.02 &43.83 \\
\textbf{\branch} &626.489 &79.25 &99.11 &155.09 &132.98 &52.26 &12.93 \\
\textbf{\f} &1979.185 &861.95 &606.01 &827.65 &790.01 &881.77 &360.12 \\
        \bottomrule
\end{tabular}}%
\caption{\sed}
\label{tab:prediction_sed_RT}
\end{subtable}

\begin{subtable}{2.5in}
\scalebox{0.7}{
\begin{tabular}{ccccccccc}
    \toprule
        \textbf{Datasets} & \multicolumn{4}{c}{\textbf{Range ($\theta=2$)}} & \multicolumn{4}{c}{\textbf{$10$-NN}}\\
                \cmidrule(lr){2-5}\cmidrule(lr){6-9}

        & \multicolumn{2}{c}{\textbf{CPU}} & \multicolumn{2}{c}{\textbf{GPU}}& \multicolumn{2}{c}{\textbf{CPU}} & \multicolumn{2}{c}{\textbf{GPU}}  \\ 
        \cmidrule(lr){2-3}\cmidrule(lr){4-5}\cmidrule(lr){6-7}\cmidrule(lr){8-9}
        &\textbf{L-Scan}&\textbf{Indexed}&\textbf{L-Scan}&\textbf{H2MN}&\textbf{L-Scan}&\textbf{Indexed}&\textbf{L-Scan}&\textbf{H2MN}\\
        \midrule
        \pubmed&0.693&0.56&0.004&26.6&1.01&0.49&0.004&27.5\\
        \amz&9.09&5.07&0.025&371&11.3&4.75&0.027&372\\
        \dblp&48&20.9&0.070&696&50.4&18.6&0.126&698\\
        \bottomrule
\end{tabular}}
\caption{Scalability}
    \label{tab:scalability}
    \end{subtable}\hspace{10em}
    \caption{\textbf{(a-b) Running times of all methods in seconds per 10k query-target pair. (c) Querying time (s) for \sed in the three largest datasets. L-Scan indicates time taken by linear scan in \name (times differ based on whether executed on CPU or GPU). 
    }}
    \label{tab:time}
\end{table*}
\subsection{ Pair-independence and Indexability} 
Here, we showcase how pair-independent embeddings, and ensuring triangle inequality leads to further boost in scalability. For this experiment, we use the three largest datasets of \pubmed, \amz and \dblp. For each dataset, we pre-compute \name embeddings of all database graphs by exploiting pair-independent embeddings. Such pre-computation is not possible in the neural or non-neural baselines. Furthermore, since the predictions of \name satisfy triangle inequality, we index the pre-computed embeddings of the database graphs as discussed in \S~\ref{sec:indexing}. Consequently, for \name, we only need to embed the query graph and evaluate $\CF$ to make predictions at query time. Table~\ref{tab:scalability} presents the results on range and $10$-NN queries. When computations are done on a GPU, \name is more than 1000 times faster than \hmnr. In the absence of a GPU, \hmnr is practically infeasible since expensive pair-dependent computations are done at query time. In contrast, even on a CPU, through indexing, \name is $\approx 50$ times faster than GPU-based \hmnr. Note that indexing enables up to $3$-times speed-up on \name over linear scan, which demonstrates the gain from ensuring triangle inequality. These results establish that \name breaks new ground in scalability of neural graph distance computations; not only is it faster, it overcomes the barrier of GPU-dependence and hence better suited for low-resource environments.
\subsection{Ablation Study}
\label{sec:ablation}
In this study, we explore the impact of our inductive biases in learning from low-volume data. We create two variants of \name: \textbf{(1)} $\name$-Dual trains the two parallel \gnn models separately without weight-sharing, and \textbf{(2)}
\name-NN uses an MLP instead of $\mathcal{F}$.
Both have strictly better representational capacity than \name, so are expected to match the performance with infinite data. Figs.~\ref{fig:ablation_dblp}-\ref{fig:ablation_cite} present the results on \sed. \rev{The results of the same experiment on \ged is provided in Figs.~\ref{fig:ablation_ged} in the appendix.} The RMSE of \name is \rev{generally} better than \name-Dual, with the difference being more significant at low volumes. This indicates that siamese structure helps. Compared to \name, \name-NN achieves marginally better performance at larger train sizes in \pubmed and \seer. However, in \dblp, \name is consistently better. The number of subgraphs in a dataset grows exponentially with the node set size. Hence, an MLP needs growing training data to accurately model the intricacies of this search space. In \dblp, even 100k pairs is not enough to improve upon $\mathcal{F}$. Furthermore, since computing \ged and \sed is NP-hard, generating large volumes of training data is not desirable. Overall, these trends indicate that $\mathcal{F}$ enables better generalization and scalability with respect to accuracy. Furthermore, given that its performance is close to an MLP even on high-volume training data, and it enables indexing, the benefits outweigh the marginal reduction in accuracy. 
\looseness=-1

\rev{We also observe that generally, \name-NN performs better than \name-dual. \name-NN retains the inductive bias imparted by the Siamese architecture, but ablates the inductive bias of the custom prediction function. The opposite happens in the case of \name-Dual. The observed phenomenon of \name-NN generally out-performing \name-Dual can be interpreted as evidence for the Siamese architecture providing a stronger inductive bias than the custom prediction function.}

More ablations studies justifying our choice of \gin and the sum-pool layer are provided in App.~\ref{app:ablation}.
\looseness=-1
\begin{table}[b]
\centering
\hspace{-1.5em}
\begin{subtable}{1.5in}
    \scalebox{0.75}{
    \begin{tabular}{c c c c}
    \toprule
         Sampler & \pubmed & \seer & \amz \\
         \midrule
         BFS & $0.728$ & $0.519$ &	$0.495$  \\
        \RW & $0.508$ & $0.770$ &	$0.490$\\ 
        \rwr & $0.545$ & $0.754$ & $0.299$\\ 
        \shadow & $0.966$ &	$0.753$ &	$0.830$ \\ 
        \bottomrule
    \end{tabular}}
    \caption{Query distributions}
        \label{tab:querydist}
    \end{subtable}
    \hspace{3.5em}
    \begin{subtable}{3.3in}
       \scalebox{0.55}{
    \begin{tabular}{crrrrrr}
    \toprule
        \textbf{Method} & \multicolumn{2}{c}{\textbf{\pubmed}} & \multicolumn{2}{c}{\textbf{\seer}} & \multicolumn{2}{c}{\textbf{\amz}}  \\
        \cmidrule(lr){2-3}\cmidrule(lr){4-5}\cmidrule(lr){6-7}
        & $\CV_Q\in[0,50]$ & $\CV_Q\in[25,50]$& $\CV_Q\in[0,50]$ & $\CV_Q\in[25,50]$& $\CV_Q\in[0,50]$ & $\CV_Q\in[25,50]$\\
        \midrule
       \textbf{ \name-50} & 1.294& 1.917& 0.728& 0.948 & 0.638& 0.782\\
        \textbf{\name-25}& 2.824& 4.999& 4.740& 9.052& 1.152& 1.724\\
        \textbf{ \hmnr-50} & 3.133& 5.112& 4.9380& 8.583 & 6.014& 9.550\\
        \textbf{\hmnr-25}& 7.417& 13.366& 10.459& 19.787& 5.720& 9.462\\
        \textbf{MIP-F2} & 3.507 & 6.278 & 4.831 & 8.505& 6.454 & 10.293\\
         \bottomrule
    \end{tabular}}
    \caption{Query size}
    \label{tab:querysize}
    \end{subtable}
    \caption{\bf (a) RMSE on unseen query distributions. BFS (seen) is the baseline to compare against. (b) RMSE against query sizes. \name-50 indicates \name trained on a dataset containing queries of size up to $50$. \name-25 is defined analogously.}
\end{table}
\subsection{Generalization to Unseen Query Distributions in \sed}
\label{sec:querydist}
We train the model by sampling queries from the graph database through BFS enumerations. \textit{How does \name generalize to unseen distributions?} Towards that end, we generate queries from the three unseen distributions of \textbf{(1)} Random Walks (\RW), \textbf{(2)} Random Walks with Restarts (\rwr), and \textbf{(3)} \shadow~\citep{shadow} (See App.~\ref{app:querydist} for details on the sampling strategies). We first note that in \aidsg, we use real queries of functional groups, and thus the good performance in \aidsg indicates good generalizability. In Table~\ref{tab:querydist}, we more exhaustively analyze this aspect. As visible, the errors remain low. Even more surprisingly, the errors on \RW and \rwr are better than the train distribution of BFS itself. This indicates good generalization to unseen distributions.
\subsection{Generalizability to Unseen, Larger Query Sizes:} 
Generating training data for learning \ged and \sed is expensive since optimal distance computations are NP-hard. Hence, a desirable property would be to learn from small graphs and then generalize to larger unseen graphs. We evaluate this ability for \name and \hmnr. Table~\ref{tab:querysize} provides the numbers. We notice that although there is some deterioration in the quality for query sizes in the range $[25,50]$ when compared to the entire set, it is not severe (\name-50 in Table~\ref{tab:querysize}). However, if the train set only contains queries till size $25$ and we deploy the learned model to infer on queries of larger unseen sizes, the drop in quality is significant (\name-25 in Table~\ref{tab:querysize}). This drop is even more dramatic in \hmnr. On the positive side, \name remains superior to the optimal non-neural approach (MIP-F2) when run with a generous time limit of 60 seconds per query. Overall, this experiment highlights one direction that needs further study and improvement.

%% file: 7_conclusion.tex
\vspace{-0.10in}
\section{Conclusions, Limitation, and Future Directions}
\label{sec:conclusion}
\vspace{-0.10in}
The problem of learning graph distances from their embeddings has seen much interest over the last few years. This thread of research is important since it allows us to overcome the bottleneck of exponential graph alignment space. Our experiments clearly establish \name as the state of the art for both \ged and \sed \rev{(See App.~\ref{app:mcs} for preliminary results on\textit{ maximum common subgraph similarity})}. In addition, it is significantly faster and provides better theoretical correspondence between properties of the original space and predicted space. One clear direction of future work that emerges from our experiments is that \name, and existing methods of graph distance learning, do not generalize well to unseen larger query sizes. We hope to address this limitation next. 

%% file: 8_appendix.tex
\section*{Appendix}
\label{sec:paper_appendix}

\appendix
\renewcommand{\thesubsection}{\Alph{subsection}}
\renewcommand{\thefigure}{\Alph{figure}}
\renewcommand{\thetable}{\Alph{table}}

\subsection{\ged and \sed}
\label{app:gedsed}
The computation of \ged relies on a \textit{graph mapping}. 

 \vspace{-0.05in}
\begin{definition}[Graph Mapping]
\label{def:mapping}
Given two graphs $\CG_1$ and $\CG_2$, let $\Tilde{\CG_1}=(\Tilde{\CV_1},\Tilde{\CE_1},\Tilde{\CL_1})$ and $\Tilde{\CG_2}=(\Tilde{\CV_2},\Tilde{\CE_2},\Tilde{\CL_2})$ be obtained by adding dummy nodes and edges (labeled with $\epsilon$) to $\CG_1$ and $\CG_2$ respectively, such that $|\CV_1| = |\CV_2|$ and $|\CE_1| = |\CE_2|$. A \textit{node mapping} between $\CG_1$ and $\CG_2$ is a bijection $\pi: \Tilde{\CG_1} \rightarrow \Tilde{\CG_2}$ where \textbf{(i)} $\forall v \in \Tilde{\CV_1}, \pi(v)\in \Tilde{\CV_2}$ and at least one of $v$ and $\pi(v)$ is not a dummy; \textbf{(ii)} $\forall e=(v_1,v_2) \in \Tilde{\CE_1}, \pi(e)=(\pi(v_1),(\pi(v_2)))\in \Tilde{\CE_2}$ and at least one of $e$ and $\pi(e)$ is not a dummy.
\end{definition}

\vspace{-0.05in}
\begin{example} 
Fig.~\ref{fig:gedsed} shows a graph mapping. Edge mappings can be trivially inferred.
\end{example}
\vspace{-0.05in}
\begin{definition}[Graph Edit Distance (\ged) under mapping $\pi$]
\label{def:gedmap}
\ged between $\CG_1$ and $\CG_2$ under $\pi$ is 
\vspace{-0.05in}
\begin{equation}
\ged_{\pi}(\CG_1,\CG_2) = \sum_{v \in \Tilde{V_1}}d(\CL(v),\CL(\pi(v))) + \sum_{e \in \Tilde{E_1}}d(\CL(e),\CL(\pi(e)))
\end{equation}
where $d: \Sigma \times \Sigma \rightarrow \mathbb{R}^+_0$ is a distance function over the label set. $d(\ell_1,\ell_2)$ models an \textit{insertion} if $\ell_1=\epsilon$, \textit{deletion} if $\ell_2=\epsilon$ and \textit{replacement} if $\ell_1\neq \ell_2$ and neither $\ell_1$ nor $\ell_2$ is a dummy.
\end{definition}
\noindent
\vspace{-0.10in}
We assume $d$ to be a binary function, where $d(\ell_1,\ell_2)=1$ if $\ell_1\neq\ell_2$, otherwise, $0$. 
\begin{definition}[Graph Edit Distance (\ged)]
\label{def:ged}
\ged is the minimum distance under all mappings.
\vspace{-0.05in}
\begin{equation}
    \ged(\CG_1,\CG_2) = \min_{\forall \pi \in \Phi(\CG_1,\CG_2)}\ged_{\pi}(\CG_1,\CG_2)
\end{equation} 
$\Phi(\CG_1, \CG_2)$ denotes the set of all possible node maps from $\CG_1$ to $\CG_2$.
\end{definition}
\vspace{-0.05in}
\begin{definition}[Subgraph Edit Distance (\sed)]
\sed is the minimum \ged over all subgraphs of $\CG_2$. 
 \vspace{-0.05in}
\begin{equation}
\label{eq:sed}
    \sed(\CG_1, \CG_2) = \min_{\CS \subseteq \CG_2}\ged(\CG_1, \CS)
\end{equation}
\end{definition}

\vspace{-0.05in}
\begin{obs}
\label{obs:sed_ged_non_neg}
\textbf{(i)} $\ged(\CG_1, \CG_2) \ge 0$, \textbf{(ii)} $\sed(\CG_1, \CG_2) \ge 0$.
\end{obs}
\begin{obs}
\label{obs:sed_isomor}
\textbf{(i)} $\ged(\CG_1, \CG_2) = 0$ iff $\CG_1$ is isomorphic to $\CG_2$,
\textbf{(ii)} $\sed(\CG_1, \CG_2) = 0$ iff $\CG_1$ is subgraph isomorphic to $\CG_2$.
\end{obs}
\subsection{Additional Proofs}
\label{app:proofs}
\subsubsection{Proof of Theorem.~\ref{thm:sed_as_ged}}
\label{app:sedti}

Our proof relies on two lemmas.

\textsc{Proof} of Theorem~\ref{thm:sed_as_ged}. It suffices to prove \textbf{(i)} $\sed(\CG_1, \CG_2) \ge \widehat{\ged}(\CG_1, \CG_2)$ and \textbf{(ii)} $\widehat{\ged}(\CG_1, \CG_2) \ge \sed(\CG_1, \CG_2)$.

\textbf{(i)} Let $\CS=(\CV_{\CS},\CE_{\CS},\CL_{\CS})\subseteq\CG_2$ be the subgraph minimizing $\sed(\CG_1, \CS)$ (Recall Eq.~\ref{eq:sed}). Consider the mapping $\pi$ from $\CG_1$ to $\CS$ corresponding to  $\sed(\CG_1, \CS)$ (and hence $\ged(\CG_1, \CS)$ as well). We extend $\pi$ to define a mapping $\widehat{\pi}$ from $\CG_1$ to $\CG_2$ by mapping of all nodes in set $\CV_2\setminus\CV_{\CS}$ to dummy nodes in $\CG_1$; the edge mappings are defined analogously.

Under this construction, $\sed(\CG_1,\CS)=\ged(\CG_1,\CG_2)=$ \\ $\widehat{\ged}_{\widehat{\pi}}(\CG_1,\CG_2)\geq \widehat{\ged}(\CG_1,\CG_2)$. This follows from the property that under $\widehat{d}$, insertion costs are zero, that is $\widehat{d}(\epsilon,\ell)=0$. Thus, the additional mappings introduced in $\widehat{\pi}$ do not incur additional costs under $\widehat{d}$.  



\textbf{(ii)} Consider $\CS\subseteq \CG_2$ and a mapping $\pi$ from $\CG_1$ to $\CS$ such that $\ged_{\pi}(\CG_1,\CS) = \widehat{\ged}(\CG_1, \CG_2)$. The existence of such a subgraph is guaranteed (See Lemma~\ref{lem:ged_c_dash}). From the definition of \ged, $\ged_{\pi}(\CG_1,\CS)\geq \ged(\CG_1,\CS)$. Furthermore, since $\CS\subseteq\CG_2$, $\ged(\CG_1,S)\geq\sed(\CG_1,\CG_2)$. Combining all these results, we have $\widehat{\ged}(\CG_1, \CG_2) \geq \ged(\CG_1, S) \ge \sed(\CG_1, \CG_2)$.\\ Hence, the claim is proved. $\hfill\square$

\subsubsection{Proof of Thm~\ref{thm:sed_actual_triangle}.}
\label{app:sedti1}
\textsc{Proof.} From Thm.~\ref{thm:sed_as_ged}, we know $\sed(\CG_1,\CG_2)=\widehat{\ged}(\CG_1,\CG_2)$. Combining Obs.~\ref{obs:metric} with Theorem~\ref{thm:sed_as_ged}, if we show  $\widehat{d}(\ell_1, \ell_3) \le \widehat{d}(\ell_1, \ell_2) + \widehat{d}(\ell_2, \ell_3)$, then the triangle inequality of \sed is established. We divide the proof into four cases: \\\textbf{(i)} None of $\ell_1$, $\ell_2$, $\ell_3$ is $\epsilon$. Hence, $\widehat{d}(\ell_1, \ell_3)=d(\ell_1, \ell_3)$ and \revision{the} triangle inequality is satisfied.\\
\textbf{(ii)} $\ell_1=\epsilon$. The \textsc{Lhs} is $0$ and hence \revision{the} triangle inequality is satisfied.\\
(\textbf{iii)} $\ell_1\neq\epsilon$ and $\ell_2=\epsilon$. \textsc{Lhs}$\leq 1$ and \textsc{Rhs}$=1$. Hence, satisfied.\\
\textbf{(iv)} Only $\ell_3=\epsilon$. Here, \textsc{Lhs}$=1$ and \textsc{Rhs}$\geq 1$.\\
These four cases cover all possible situations and hence,  \revision{the} triangle inequality is established.$\hfill\square$

\subsubsection{Proof of Lemma \ref{lem:ged_c_dash}}

\begin{lem}
\label{lem:ged_c_dash}
There exists a subgraph $\CS$ of $\CG_2$ and a node map $\pi$ from $\CG_1$ to $\CS$ such that $\ged_{\pi}(\CG_1,\CS) = \widehat{\ged}(\CG_1, \CG_2)$.
\end{lem}
\textsc{Proof.}
Let $\pi'$ be a node map from $\CG_1$ to $\CG_2$ corresponding to $\widehat{\ged}(\CG_1, \CG_2)$. Let $h_1,\cdots,h_l$ be the nodes of $\CG_2$ which are inserted in $\pi'$. Construct subgraph $\CS$ of $\CG_2$ by removing nodes $h_1,\cdots,h_l$ and their incident edges from $\CG_2$. Let $\pi$ be the node map from $\CG_1$ to $\CS$ which is obtained by removing $h_1',\cdots,h_l'$ from the domain and $h_1,\cdots,h_l$ from the co-domain of $\pi$. Since insertion costs are $0$ in $\widehat{d}$ and $\pi$ contains only non-insert operations, then $ \ged_{\pi}(\CG_1,\CS)= \widehat{\ged}(\CG_1, \CG_2)$. Hence, the claim is proved. \hfill $\square$

\subsubsection{Proof of Lemma~\ref{lem:ti_for_f}.}
\label{app:sed_embed_ti}
\textsc{Proof.}
Let $\cx, \cy \in \mathbb{R}^n$ be vectors of dimension $n$. We use the notation $\cx[i]$ to denote the $i^{th}$ coordinate of $\cx$. We observe that :
\vspace{-0.05in}
\begin{alignat}{2}
\nonumber
(\relu(\cx) + \relu(\cy))[i] &= \relu(\cx)[i] + \relu(\cy)[i]= \relu(\cx[i]) + \relu(\cy[i])\\
\nonumber
&\ge \relu(\cx[i] + \cy[i])= (\relu(\cx+\cy))[i]
\end{alignat}
Since $\|.\|$ is monotonic, this implies $\|\relu(\cx) + \relu(\cy)\| \ge \|\relu(\cx + \cy)\|$. Using the triangle inequality for $\|.\|$, we get:
\begin{alignat}{1}
\|\relu(\cx)\| + \|\relu(\cy)\|
&\ge \|\relu(\cx) + \relu(\cy)\| \ge \|\relu(\cx + \cy)\|
\end{alignat}
\vspace{-0.05in}
Substituting $\cx = \CZ_{\CG_{\CQ}}-\CZ_{\CG_{\CT'}}, \cy = \CZ_{\CG_{\CT'}}-\CZ_{\CG_{\CT}}$, we get,
\vspace{-0.05in}

{\small
\begin{alignat}{1}
\nonumber
\|\relu(\CZ_{\CG_{\CQ}} - \CZ_{\CG_{\CT'}})\| + \|\relu(\CZ_{\CG_{\CT'}} - \CZ_{\CG_{\CT}})\| \ge \|\relu(\CZ_{\CG_{\CQ}} - \CZ_{\CG_{\CT}})\|.
\end{alignat}}
\vspace{-0.05in}
This implies $\CF(\CZ_{\CG_{\CQ}}, \CZ_{\CG_{\CT'}}) + \CF(\CZ_{\CG_{\CT'}}, \CZ_{\CG_{\CT}}) \ge \CF_s\left(\CZ_{\CG_{\CQ}}, \CZ_{\CG_{\CT}}\right)$. $\hfill \square$

\subsection{Complexity Analysis}
\label{app:complexity}
For this analysis, we make the simplifying assumption that the hidden dimension in the Pre-\mlp, \gin and Post-\mlp are all $d$. The average density of the graph is $g$. The number of hidden layers in Pre-\mlp, and Post-\mlp are $L$, and $k$ in \gin.

The computation cost per node for each of these components are as follows.
\begin{itemize}
\item \textbf{Pre-\mlp:} The operations in the \mlp involve linear transformation over the input vector $\cx_v$ of dimension $|\Sigma|$, followed by non-linearity. This results in $O(|\CV|(\lvert\Sigma\rvert\cdot d+d^2L))$ cost.
\item \textbf{\gin:} \gin aggregates information from each of the neighbors, which consumes $O(d\cdot g)$ time. The linear transformation consumes an additional $O(d^2)$ time. Applying non-linearity takes $O(d)$ time since it is a linear pass over the hidden dimensions. Finally these operations are repeated over each of the $k$ hidden layers, results in a total $O(k(d^2+dg))$  computation time per node. Across, all nodes, the total cost is $O(|\CV|kd^2+|\CE|kd)$ time. The degree $g$ terms gets absorbed since each edge passes message twice across all nodes.
\item \textbf{Concatenation:} This step consumes $O(kd)$ time per node.
\item \textbf{Pool:} Pool iterates over the \gin representation of each node requiring $O(|\CV|dk)$ time.
\item \textbf{Post-\mlp:} The final \mlp takes $dk$ dimensional vector as input and maps it to a $d$ dimensional vector over $L$ layers. This consumes $O(kd^2+d^2L)$ time.
\end{itemize}

Combining all these factors, the total inference complexity for a graph is $O(|\CV|(\lvert\Sigma\rvert\cdot d+d^2L+kd^2)+|\CE|kd)$. This operation is repeated on both the query and target graphs to compute their embeddings, on which distance function  $\mathcal{F}$ is operated. Thus, the final cost is  $O(n(\lvert\Sigma\rvert\cdot d+d^2L+kd^2)+mkd)$, where $n=|\CV_{\CQ}|+|\CV_{\CT}|$ and $m=|\CE_{\CQ}|+|\CE_{\CT}|$.

\input{4_1_querying}

\subsection{Datasets}
\label{app:datasets}

\textbf{\dblp:} \dblp is a co-authorship network where each node is an author and two authors are connected by an edge if they have co-authored a paper. The label of a node is the venue where the authors has published most frequently. 
The dataset has been obtained from \url{https://www.aminer.org/citation}.\\
\textbf{\amz~\cite{snap}:} Each node in \amz represents a product and two nodes are connected by an edge if they are frequently co-purchased. The graph is unlabeled and hence equivalent to a graph containing a single label on all nodes.\\
\textbf{\pubmed:} \pubmed dataset is a citation network which consists of scientific publications from \pubmed database pertaining to diabetes classified into one of three classes.\\
\textbf{\protein:} \protein dataset consists of protein graphs. Each node is labeled with a one of three functional roles of the protein.\\
\textbf{\aidsg:} \aidsg dataset consists of graphs constructed from the AIDS antiviral screen database. These graphs representing molecular compounds with Hydrogen atoms omitted. Atoms are represented as nodes and chemical bonds as edges.\\
\textbf{\aids:} \aids dataset is another collection of graphs constructed from the AIDS antiviral screen database. The graphs and their properties differ from those in \aidsg. These graphs also represent chemical compound structures. \\
\textbf{\seer:} \seer is a citation network which consists of scientific publications classified into one of six classes. Generally a smaller version is used for this dataset, but we use the larger version from ~\citep{bojchevski2017deep}.\\
\textbf{\cora:} Cora dataset is a citation dataset consisting of many scientific publications classified into one of seven classes based on paper topic. \cora is a smaller datset extracted from Cora~\citep{bojchevski2017deep}.\\
\textbf{\linux:} \linux dataset is a collection of program dependence graphs, where each graph is a function and nodes represent statements while edges represent dependency between statements.\\
\textbf{\imdb:} \imdb dataset is a collection of ego-networks of actors/actresses that have appeared together in
any movie.

\subsection{Baselines}
\label{app:baseline}
GMN explicitly assumes the distance function to be symmetric, which violates \sed. \textsc{GraphSim} has an  assumption that a large difference in size of the query and target graphs leads to a large distance, which is not true in \sed. 
 \rev{In \gotsim, there is an explicit assumption of modeling a symmetric distance function since they use cosine similarity to compare node neighborhoods The normalization factor in Eq. 6 of ~\cite{graphotsim} is also based on whole graph matching.} Finally, \genn is not included for \sed since it does not scale on graphs beyond $10$ nodes (See \S~\ref{sec:efficiency}).

\subsection{Ablation Study}
\label{app:ablation}

\begin{table}[t]
\centering
\end{table}

\begin{figure*}[tbh]
    \centering
    \captionsetup[subfigure]{labelfont={normalsize,bf},textfont={normalsize}}
    \subfloat[Query]{\includegraphics[width=0.12\textwidth]{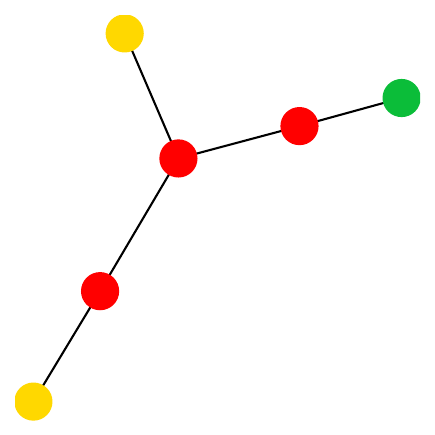}}\hfill
    \subfloat[Rank 1]{\includegraphics[width=0.12\textwidth]{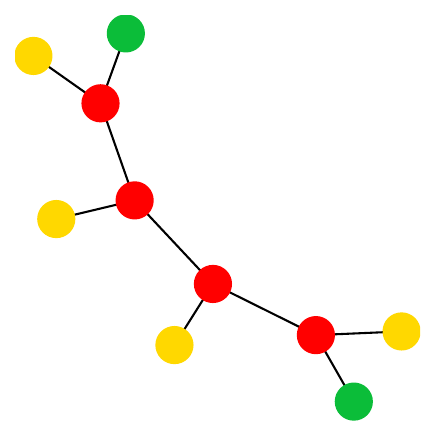}}\hfill
     \subfloat[Rank 2]{\includegraphics[width=0.12\textwidth]{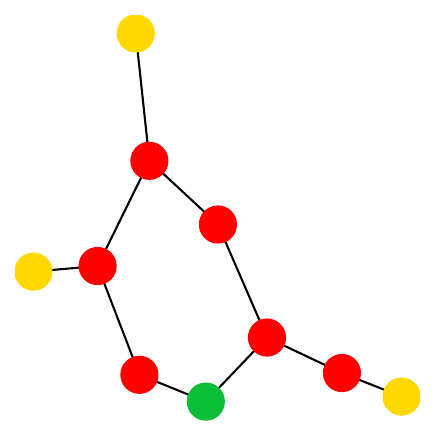}}\hfill
    \subfloat[Rank 3]{\includegraphics[width=0.12\textwidth]{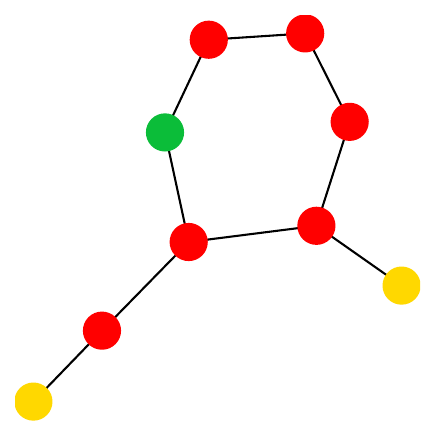}}\hfill
    \subfloat[Rank 4]{\includegraphics[width=0.12\textwidth]{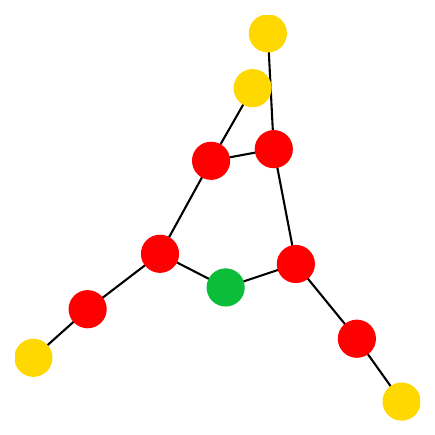}}\hfill
    \subfloat[Rank 5]{\includegraphics[width=0.12\textwidth]{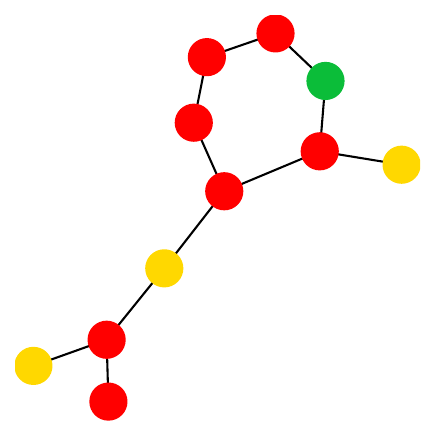}}\hfill
    \caption{\textbf{Visualizations of query and resulting matches produced by \name. Red, Green and Yellow colors indicate Carbon, Nitrogen and Oxygen atoms respectively. The actual and predicted \sed for the target graphs are (b) $0,0.4$, (c) $0,0.5$, (d) $1,0.6$, (e) $0,0.6$ and (f) $1,0.6$.}}
    \label{fig:query_vis}
\end{figure*}

\textbf{Impact of \gin:} To highlight the importance of \gin, we conduct ablation studies by replacing the GIN convolution layers in the model with several other convolution layers. As visible in the Table \ref{tab:gin_ablation}, \gin consistently achieves the best accuracy. This is not surprising since \gin is provably the most expressive among \gnns in distinguishing graph structures (essential to \sed or \ged computation) and is as powerful as the Weisfeiler-Lehman Graph Isomorphism test \citep{gin}.

\begin{table}[h]
\centering
\scriptsize
\vspace{-0.10in}
   \begin{tabular}{rrrrr}
    \toprule
        \textbf{Methods} & \seer (\sed) & \pubmed (\sed) & \amz (\sed) & \imdb (\ged) \\ 
        \midrule
        \textbf{\name (\gin)} & \textbf{0.519} & \textbf{0.728} & \textbf{0.495} & \textbf{6.734}\\
        \textbf{\name-GCN} & 0.556& 0.756& 0.532& 12.151\\
        \textbf{\name-\gsage} & 1.364& 1.156& 1.841& 91.312\\
        \textbf{\name-GAT} & 1.294 & 1.259 & 1.843 & 89.034\\
        \bottomrule
\end{tabular}%
    \caption{\textbf{Ablation studies: GIN vs others. RMSE produced by different methods are shown and \name with \gin produces the best results.}}
    \label{tab:gin_ablation}
\end{table}

\begin{table}[h!]
\centering
\scriptsize
    \begin{tabular}{rrrrr}
    \toprule
        \textbf{Pool functions} & \seer (\sed) & \pubmed (\sed) &  \amz (\sed) & \imdb (\ged)\\ 
        \midrule
        \textbf{$\name$ (Sum)} & \textbf{0.519}& 0.728& \textbf{0.495} & \textbf{6.734} \\
        \textbf{$\name$-Max} & 0.795& \textbf{0.709}& 0.603 & 52.519 \\
        \textbf{$\name$-Mean} & 0.922& 0.732& 0.846 & 52.483 \\
        \textbf{$\name$-Attention} & 0.914& 0.797& 0.868 & 130.47 \\
        \bottomrule
\end{tabular}
    \caption{\textbf{Ablation studies: sum-pool vs others. The sum-pool is the best choice among the considered alternatives.}}
    \label{tab:pool_ablation}
\end{table}
\textbf{Impact of sum-pool:} To substantiate our choice of the pooling layer, we have performed ablation studies with various pooling functions as replacements for sum-pool. It is clear from Table \ref{tab:pool_ablation} that sum-pool is the best choice among the considered alternatives.

\rev{Sum-pool can better distinguish graph sizes better than other aggregation functions such as mean-pool or max-pool. To elaborate, let us consider a graph $G_1$ that is significantly larger than another graph $G_2$. In this scenario, the individual coordinates of $G_1$’s embedding can potentially be significantly larger than those of $G_2$ since in $G_1$ the summation is being done over a larger set of embeddings. Both mean-pool and max-pool fail to capture the size information as effectively, since the max and the mean operations do not scale with the number of inputs.}

\rev{\textbf{Impact of Pre-mlp layer:} Table~\ref{tab:premlp} presents the results. As visible, we do not see any significant difference in performance on average.}
\begin{table}[h]
    \centering
    \scalebox{0.8}{
    \begin{tabular}{ccc}
    \toprule
\textbf{RMSE}&	\textbf{With Pre-MLP}&	\textbf{Without Pre-MLP}\\
\midrule
\aids (SED)&	0.51&	0.51\\
\amz	&0.5&	\textbf{0.39}\\
\seer	&0.52&	\textbf{0.51}\\
\cora &	\textbf{0.64}	&0.68\\
\aidsg (\ged)&	\textbf{0.8}	&0.85\\
\imdb (\ged)&	\textbf{6.73}&	7.68\\
\linux (\ged)&	0.42&	\textbf{0.41}\\
Protein	&0.52&	0.52\\
\pubmed	&0.73&	0.73\\
\bottomrule
    \end{tabular}}
    \caption{\rev{RMSE of \name with and without the Pre-mlp layer.}}
    \label{tab:premlp}
\end{table}

\begin{figure}[h]
    \centering
    \subfloat[\aidsg]{\includegraphics[width=1.6in]{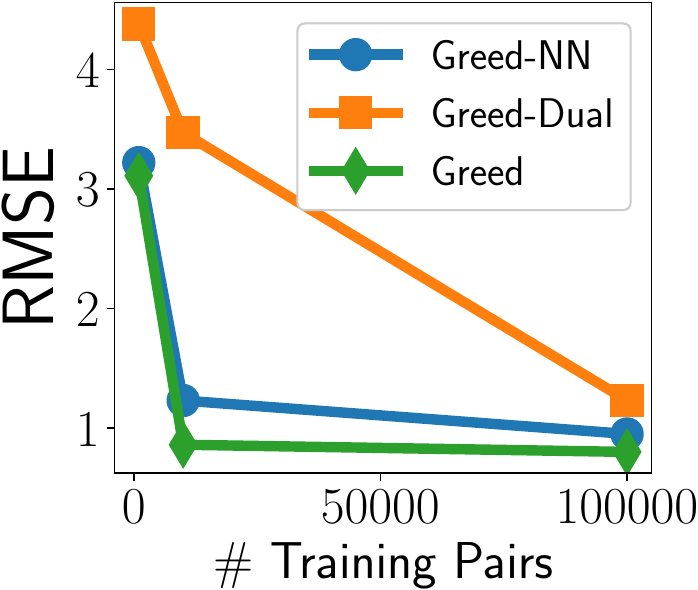}}\hfill
     \subfloat[\linux]{\includegraphics[width=1.6in]{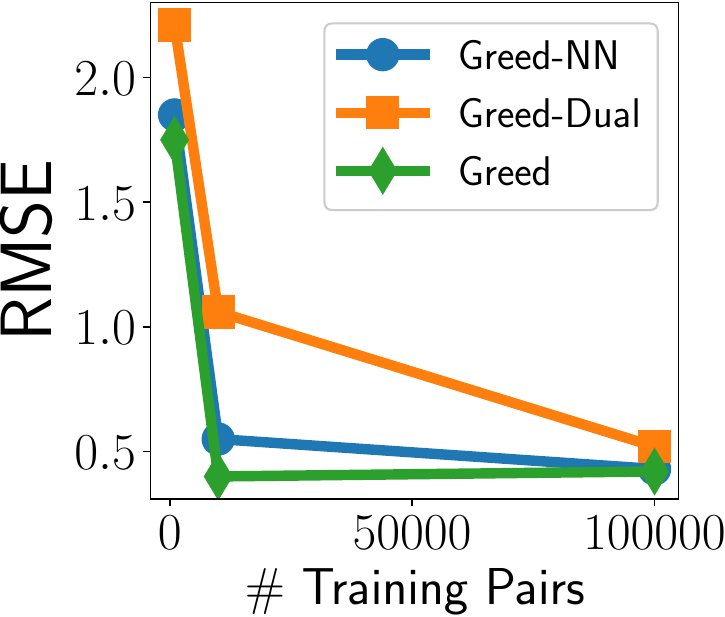}}\hfill
      \subfloat[\imdb]{\includegraphics[width=1.6in]{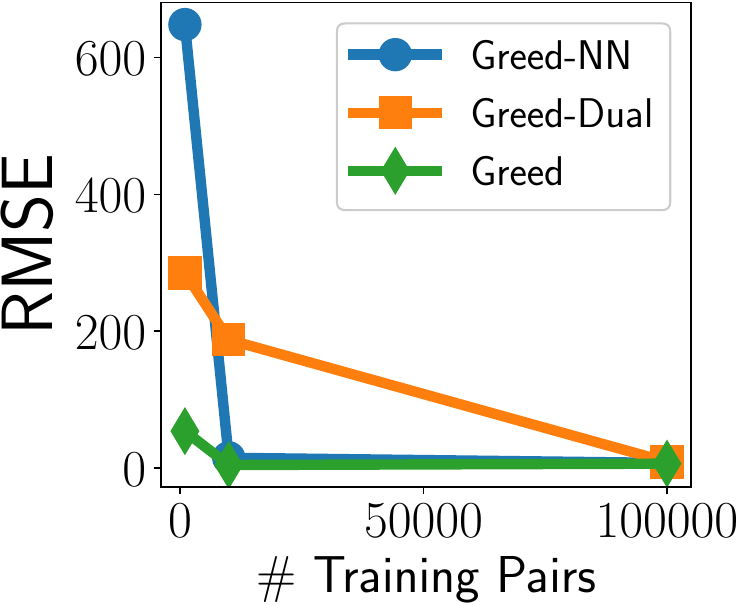}}\hfill
      \caption{\rev{Performance of \name-NN and \name-Dual on \ged. Refer to \S~\ref{sec:ablation} for details.}}
    \label{fig:ablation_ged}
\end{figure}

\rev{\textbf{Performance of \name-NN and \name-Dual on \ged}: Fig.~\ref{fig:ablation_ged} presents the results. The trends are similar to what we observed for \sed in Sec.~\ref{sec:ablation}. \name-NN closes the performance gap with \name as more training data is provided. Furthermore, \name-dual is consistently worse that \name.}

\subsection{Alignment}
\label{app:alignment}
In real-world applications of subgraph similarity search, alignments are of interest only for a small number of similar subgraphs. Our framework is intended to serve as a filter to retrieve this small set of similar subgraphs from a large number of candidates.
To elaborate, a graph database may contain thousands or millions of graphs (or alternatively, thousands or millions of neighborhoods of a large graph) which need to be inspected for similar subgraphs. A user is typically interested in only a handful of these subgraphs that are highly similar to the query. Since the filtered set is significantly smaller, a non-neural exact algorithm suffices to construct the alignments (Lemma 1 allows us to adapt general cost \ged alignment techniques for \sed alignment). Computing alignments across the entire database is unnecessary and slows down the query response time. 

To substantiate our claim, we show the average running time for answering $10$-NN queries. We break up the running time into two components: (i) $10$-NN retrieval time by \name, (ii) exact alignment time using MIP-F2 for the $10$-NN neighborhoods retrieved by \name. We observe that exact alignment by existing methods on the $10$-NN neighborhoods completes in reasonable time. In contrast, \genn does not scale on either \pubmed or \amz since it computes alignments across \textit{all} (sub)graphs.

\begin{table}[t!]
\centering
\small
    \begin{tabular}{rrr}
    \toprule
        \textbf{Methods} & \pubmed &  \amz \\ 
        \midrule
        \textbf{$\name$ Retrieval} & 0.373& 6.471 \\
        \textbf{MIP-F2 Alignment} & 52.8& 68.4 \\
        \bottomrule
\end{tabular}
    \caption{\textbf{The average running times in seconds per top-10 query. Our technique is much faster than MIP-F2 alignment.}}
    \label{tab:time_alignment}
\end{table}

\subsection{Visualization}
\label{app:visualization}
Searching for molecular fragment containment is a routine task in drug discovery~\citep{ctree}. Motivated by this, we show the top-$5$ matches to an \sed query on the \aidsg dataset produced by \name in Fig.~\ref{fig:query_vis}. The query is a functional group (Hydrogen atoms are not represented). \name is able to extract chemical compounds that contain this molecular fragment (except for ranks $3$ and $5$, which contain this group with $1$ edit) from around $2000$ chemical compounds with varying sizes and structures. This validates the efficacy of \name at a semantic level.

\subsection{Subgraph sampling strategies for \sed generalizability}
\label{app:querydist}
In \rwr, we perform fixed length random walks, where the length of a walk is the average diameter size of the queries generated through BFS during training. Next, we merge the walks to form a graph.

In \RW, we perform random walks, till the diameter of the resultant graph is the same as the average diameter of the BFS sampled train graphs.

The details of the \shadow sampler is explained in \cite{shadow}. 

\subsection{Extension to maximum common subgraph similarity (MCSS)}
\label{app:mcs}
\rev{Besides \ged, MCSS is also a popular similarity measure for whole-graph comparison. Except the inductive bias injected through $\mathcal{F}$, all components of \name is generic for any graph distance function. In this section, we replace $\mathcal{F}$ with an \mlp and model MCSS. Table~\ref{tab:mcss} presents the results. As visible, the trends hold even in this distance function, where \name outperforms the closest baseline of \hmnr.}

\begin{table}[h!]
\centering
\small
    \begin{tabular}{rrrr}
    \toprule
        \textbf{Methods} & \aidsg &  \linux & \imdb\\ 
        \midrule
        \textbf{$\name$} & \textbf{0.514}& \textbf{0.085} & \textbf{0.293} \\
        \textbf{\hmnr} &0.652& 0.152& 0.475 \\
        \bottomrule
\end{tabular}
    \caption{\textbf{\rev{RMSE on MCSS.}}}
    \label{tab:mcss}
\end{table}

\subsection{Heat Maps for Prediction Error}
\label{app:heatmaps}

\begin{figure}[ht]
    \centering
    \captionsetup[subfigure]{labelfont={normalsize,bf},textfont={normalsize}}
    \subfloat[\name]{\includegraphics[height=0.18\textwidth]{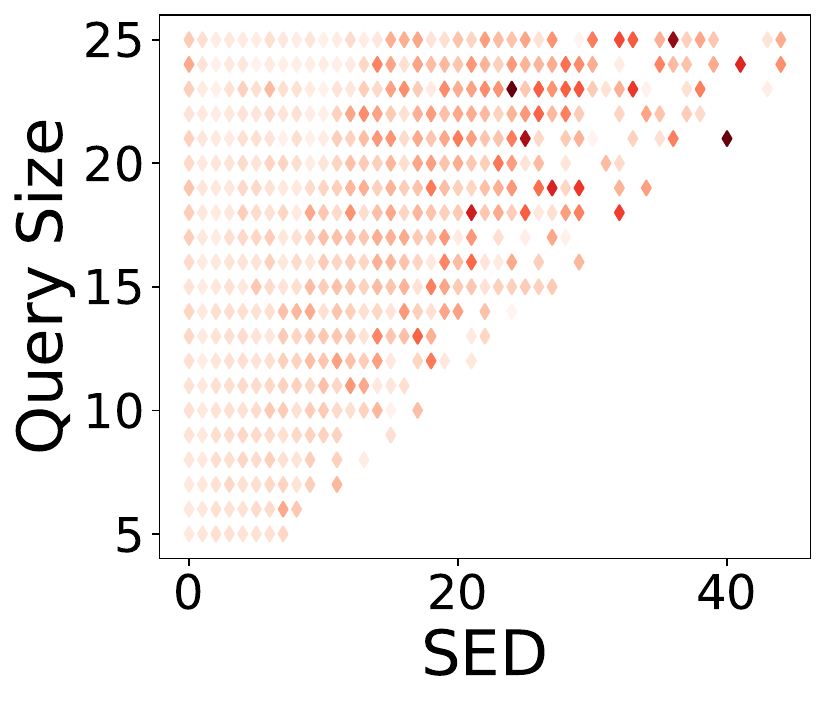}}
    \subfloat[\hmnr]{\includegraphics[height=0.18\textwidth]{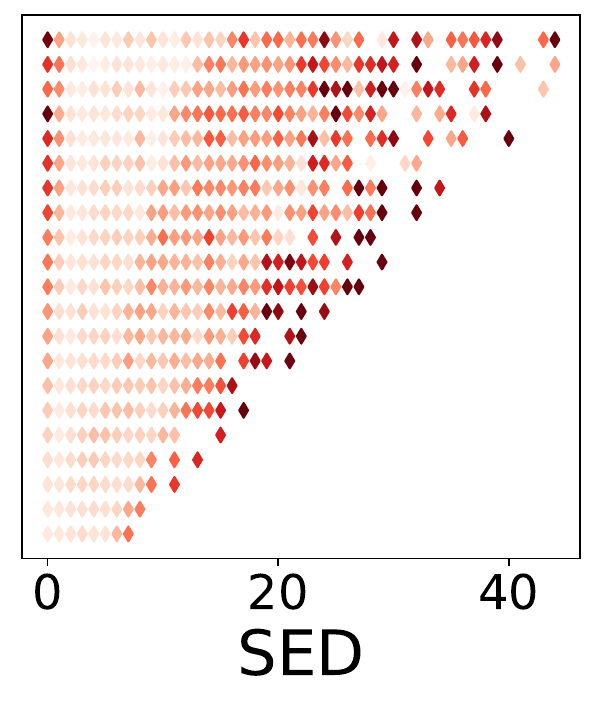}}
    \subfloat[\hmnn]{\includegraphics[height=0.18\textwidth]{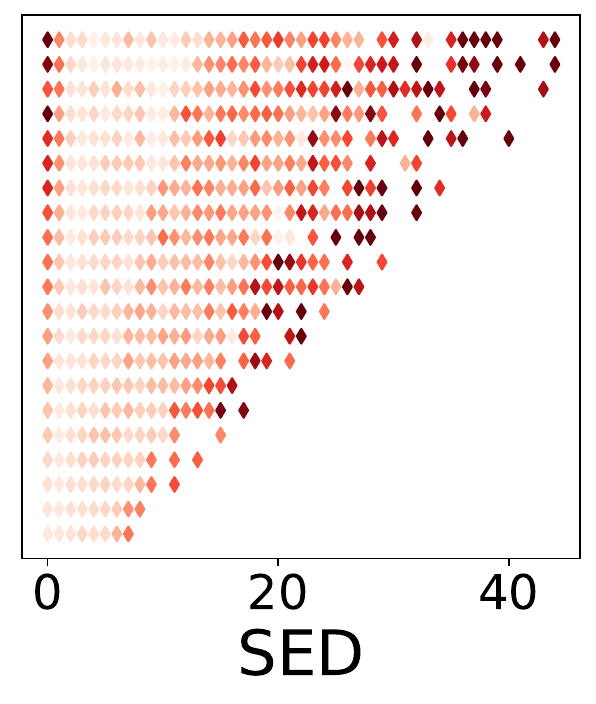}}
    \subfloat[\simgnn]{\includegraphics[height=0.18\textwidth]{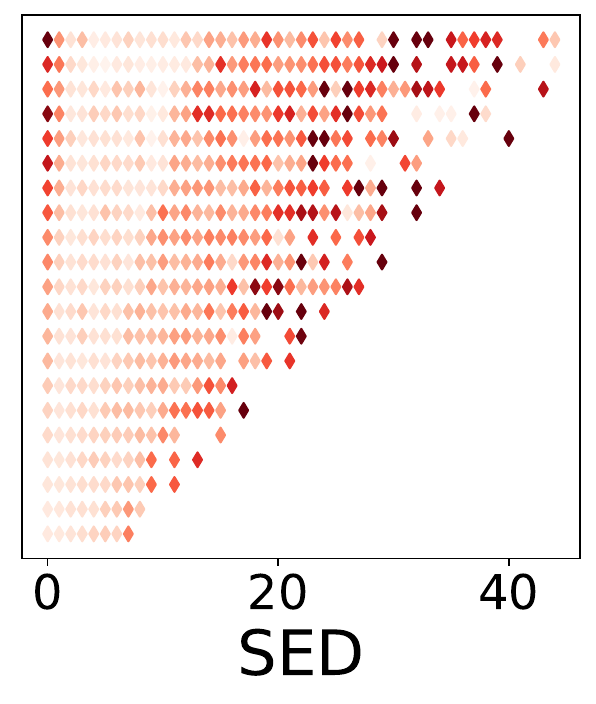}}
    \subfloat[\branch]{\includegraphics[height=0.18\textwidth]{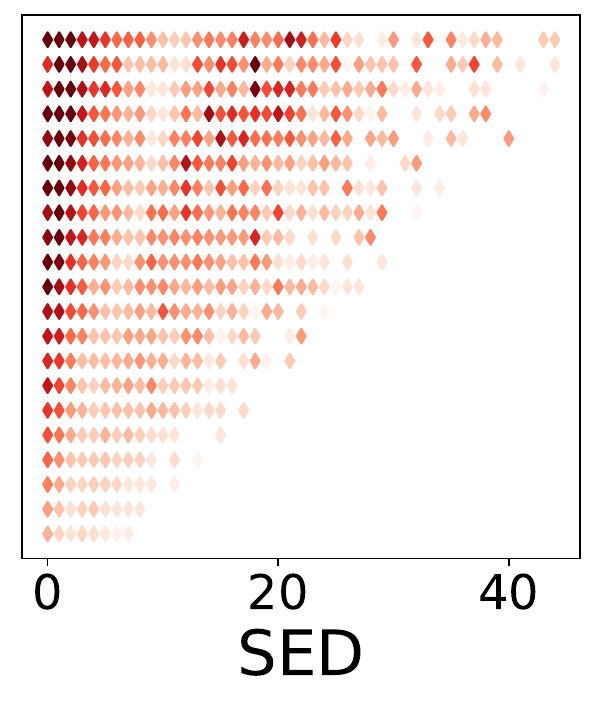}}
    \subfloat[\f]{\includegraphics[height=0.1845\textwidth]{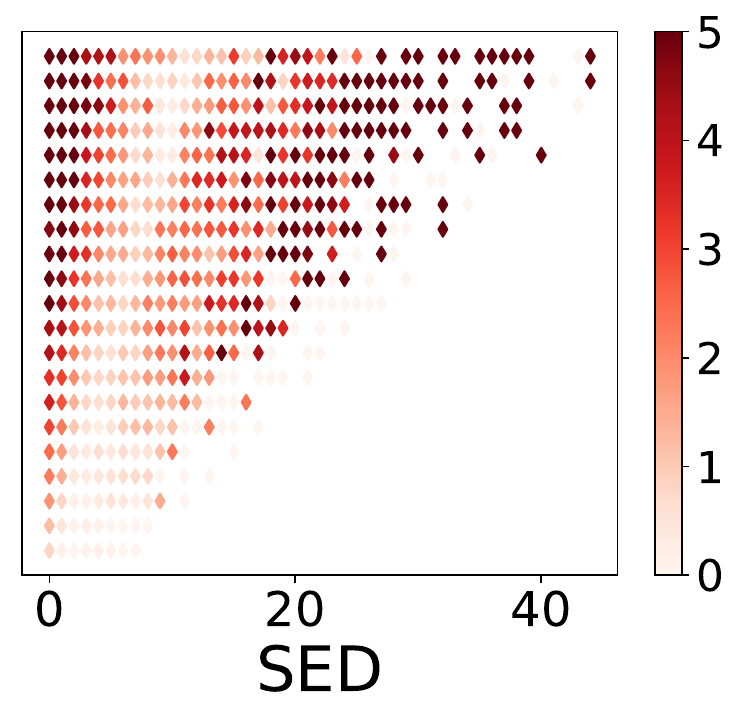}}\hfill
    \caption{\textbf{Heat Maps of \sed error against query size and \sed values for \dblp. Darker means higher error.}}
    \label{fig:heat_dblp}
    \vspace{-0.02in}
\end{figure}

\begin{figure}[ht]
    \centering
    \vspace{-0.06in}
    \captionsetup[subfigure]{labelfont={normalsize,bf},textfont={normalsize}}
    \subfloat[\name]{\includegraphics[height=0.18\textwidth]{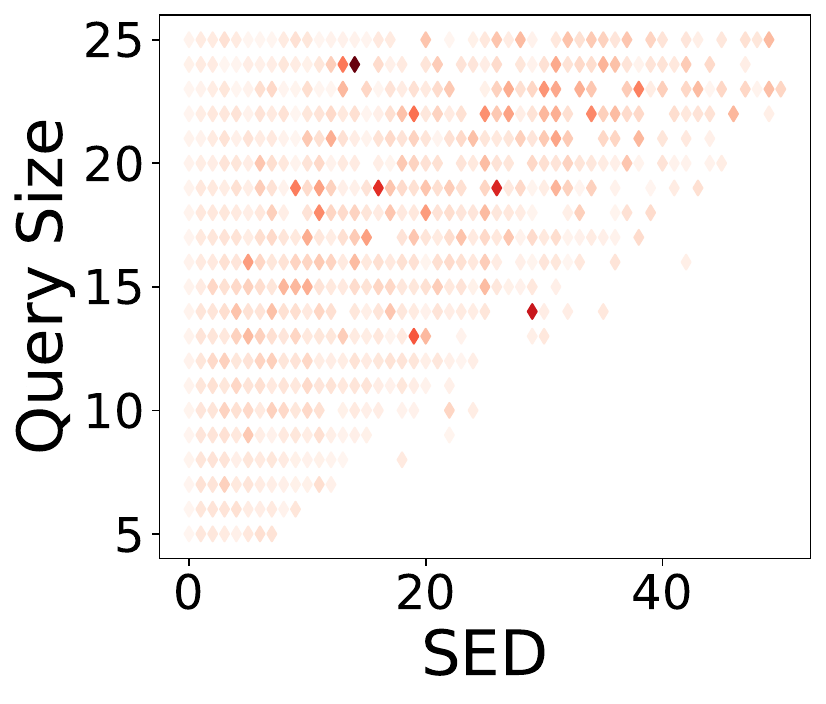}}
    \subfloat[\hmnr]{\includegraphics[height=0.18\textwidth]{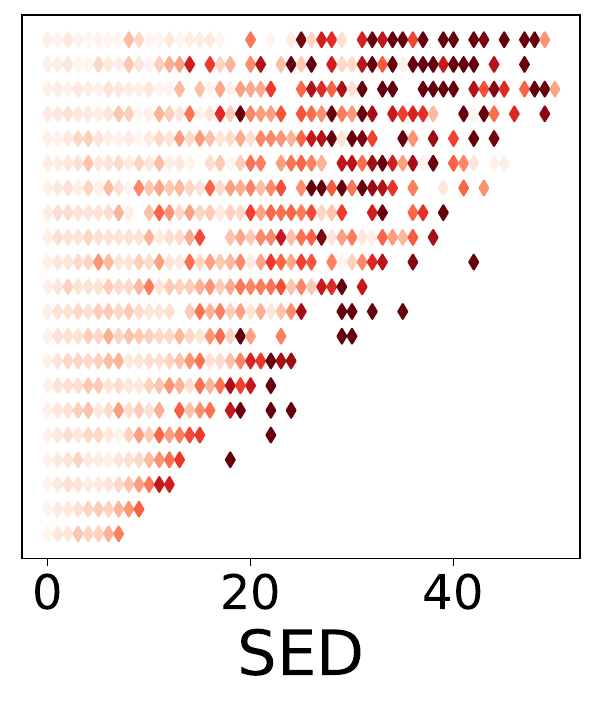}}
    \subfloat[\hmnn]{\includegraphics[height=0.18\textwidth]{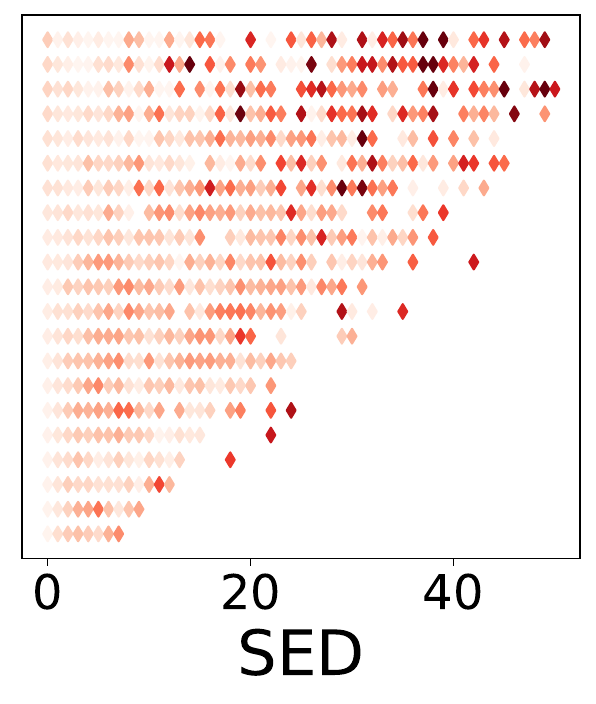}}
    \subfloat[\simgnn]{\includegraphics[height=0.18\textwidth]{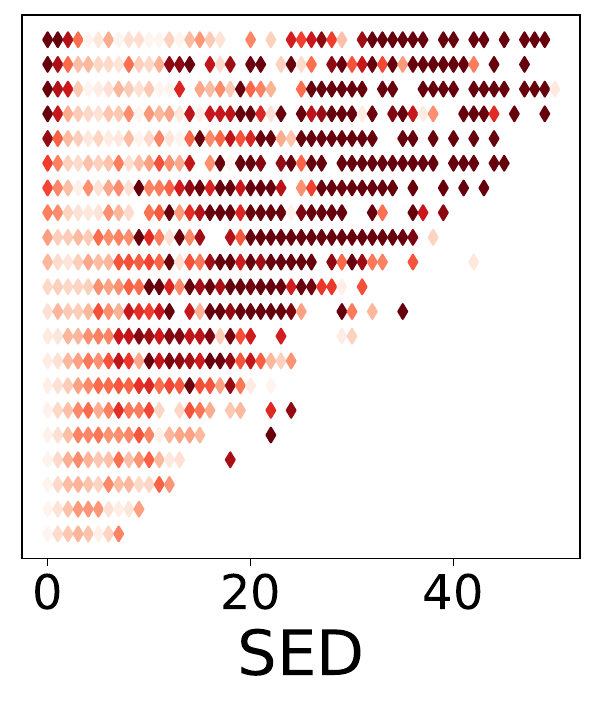}}
    \subfloat[\branch]{\includegraphics[height=0.18\textwidth]{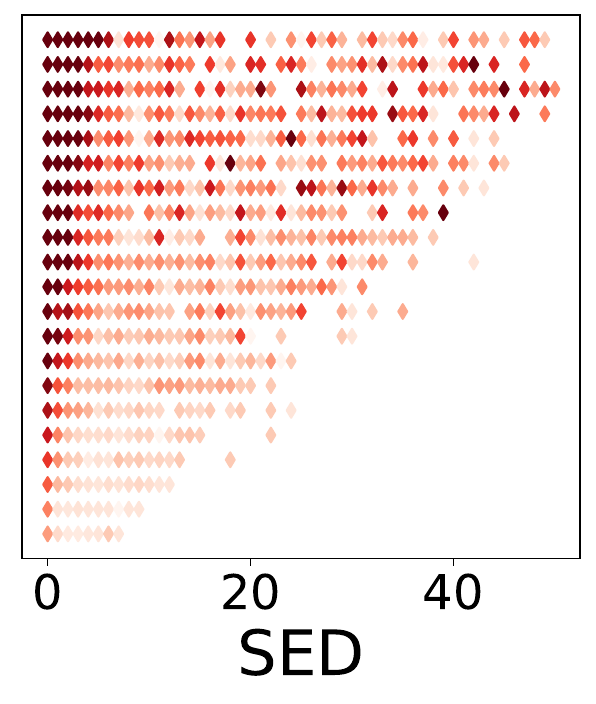}}
    \subfloat[\f]{\includegraphics[height=0.1845\textwidth]{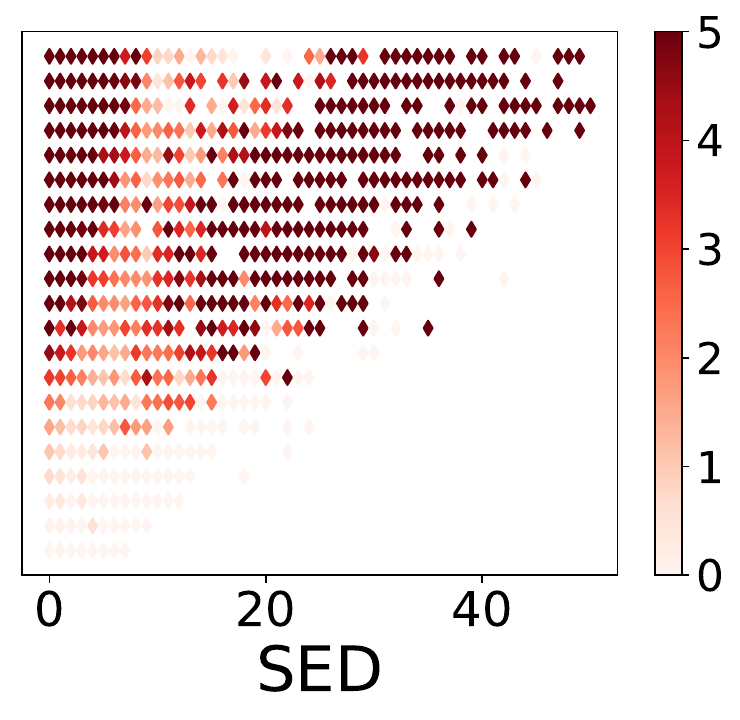}}\hfill
    \caption{\textbf{Heat Maps of \sed error against query size and \sed values for \amz. Darker means higher error.}}
    \label{fig:heat_amazon}
    \vspace{-0.08in}
\end{figure}

\begin{figure}[ht]
    \centering
    \vspace{-0.02in}
    \captionsetup[subfigure]{labelfont={normalsize,bf},textfont={normalsize}}
    \subfloat[\name]{\includegraphics[height=0.18\textwidth]{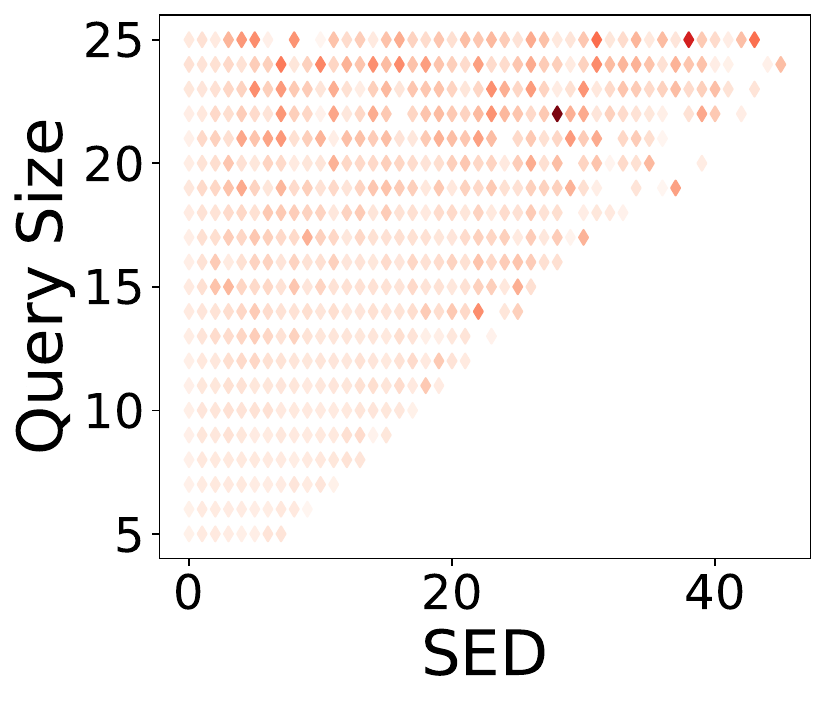}}
    \subfloat[\hmnr]{\includegraphics[height=0.18\textwidth]{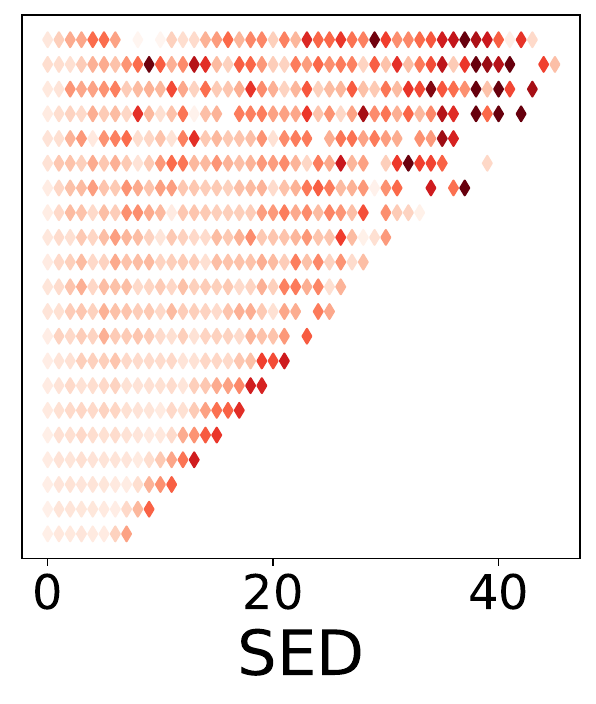}}
    \subfloat[\hmnn]{\includegraphics[height=0.18\textwidth]{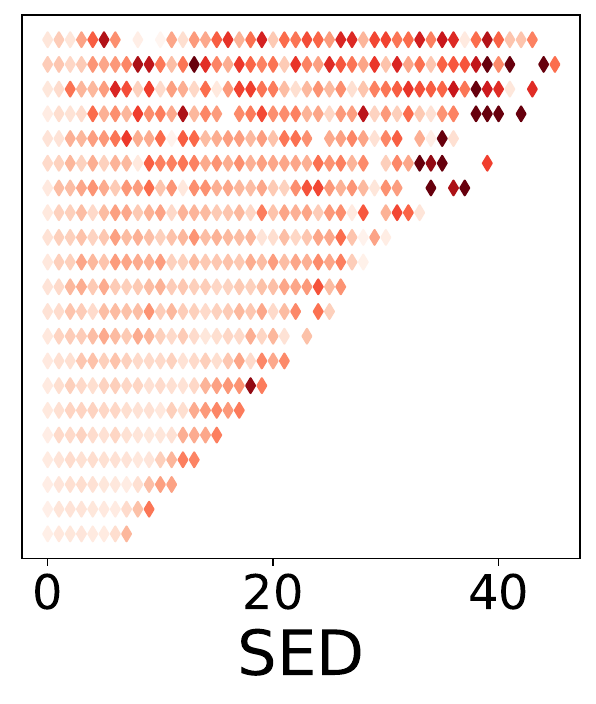}}
    \subfloat[\simgnn]{\includegraphics[height=0.18\textwidth]{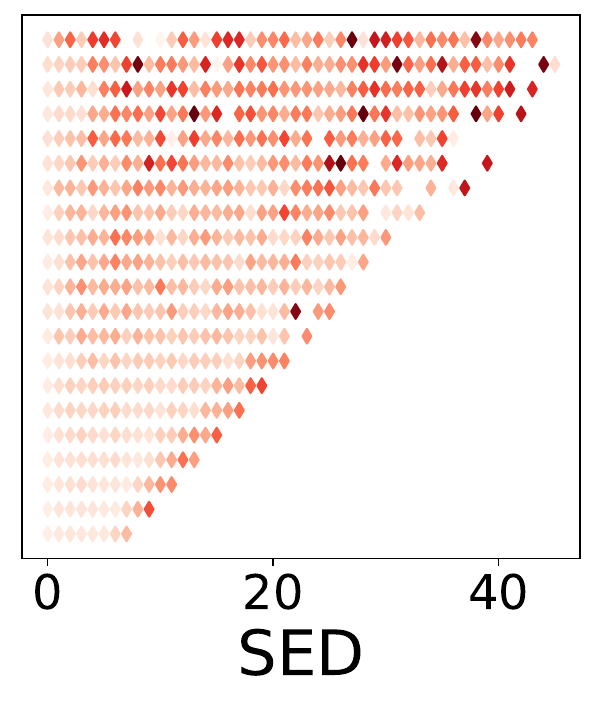}}
    \subfloat[\branch]{\includegraphics[height=0.18\textwidth]{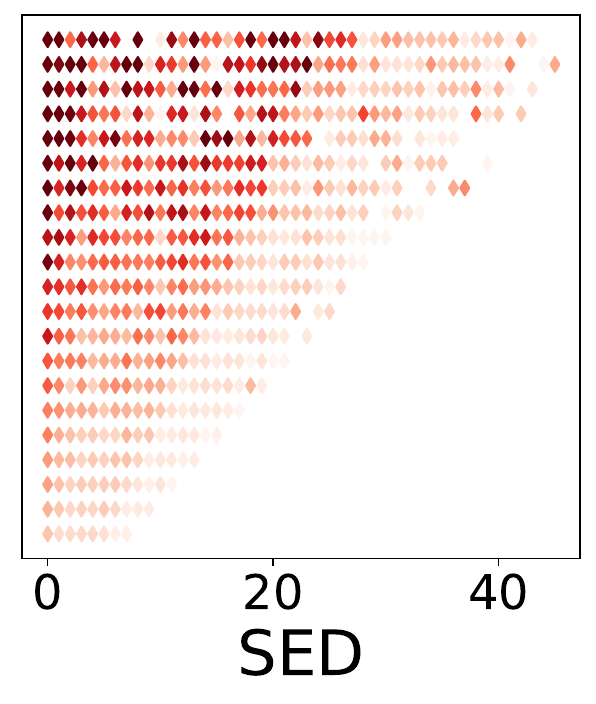}}
    \subfloat[\f]{\includegraphics[height=0.1845\textwidth]{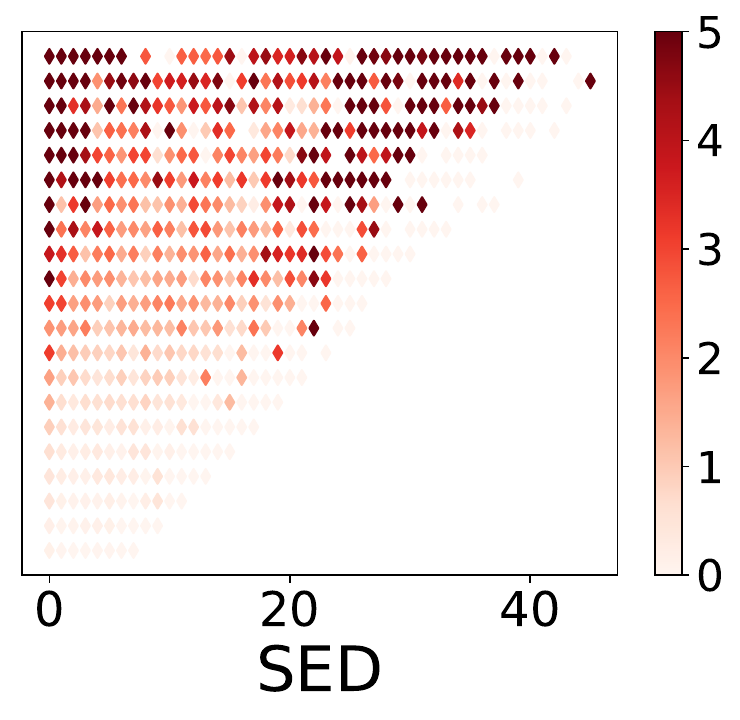}}\hfill
    \caption{\textbf{Heat Maps of \sed error against query size and \sed values for \pubmed. Darker means higher error.}}
    \label{fig:heat_pubmed}
    \vspace{-0.06in}
\end{figure}

\begin{figure}[ht]
    \centering
    \vspace{-0.02in}
    \captionsetup[subfigure]{labelfont={normalsize,bf},textfont={normalsize}}
    \subfloat[\name]{\includegraphics[height=0.18\textwidth]{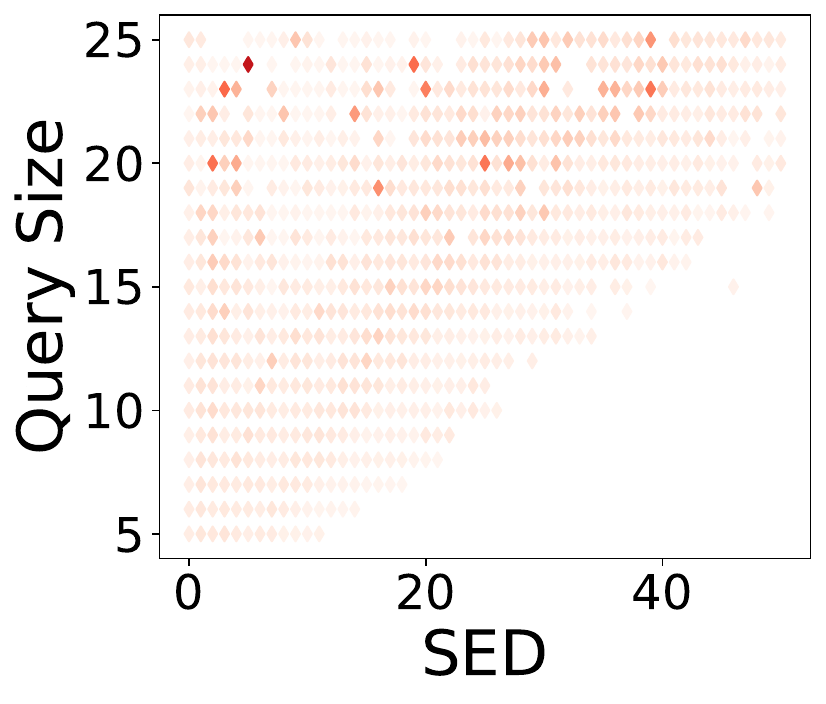}}
    \subfloat[\hmnr]{\includegraphics[height=0.18\textwidth]{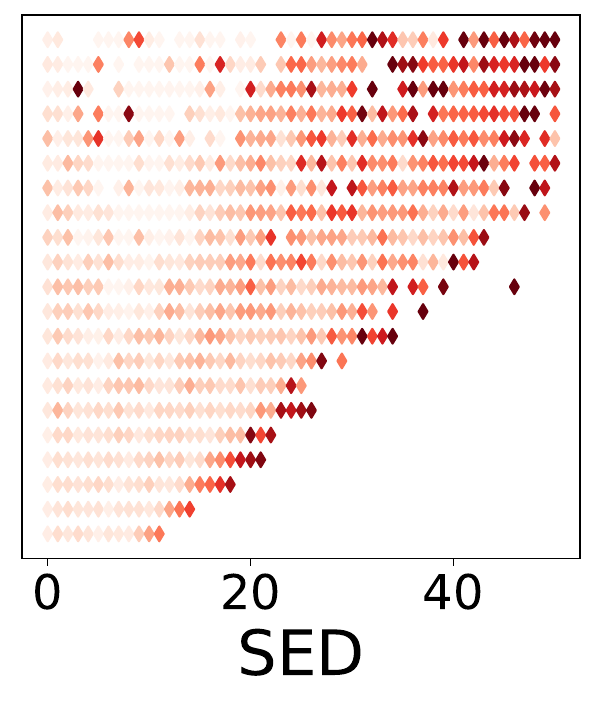}}
    \subfloat[\hmnn]{\includegraphics[height=0.18\textwidth]{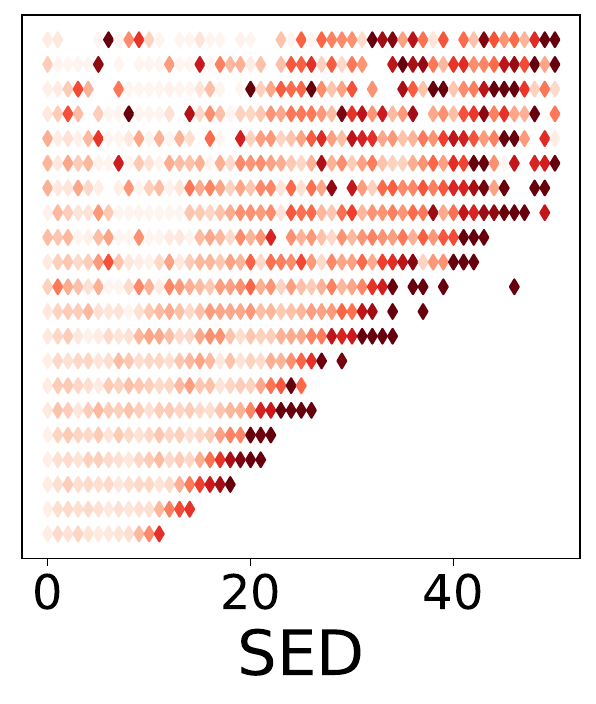}}
    \subfloat[\simgnn]{\includegraphics[height=0.18\textwidth]{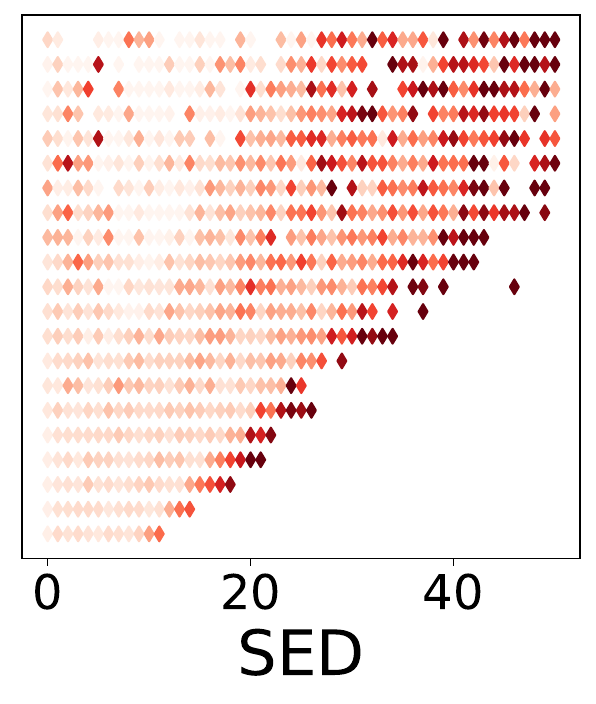}}
    \subfloat[\branch]{\includegraphics[height=0.18\textwidth]{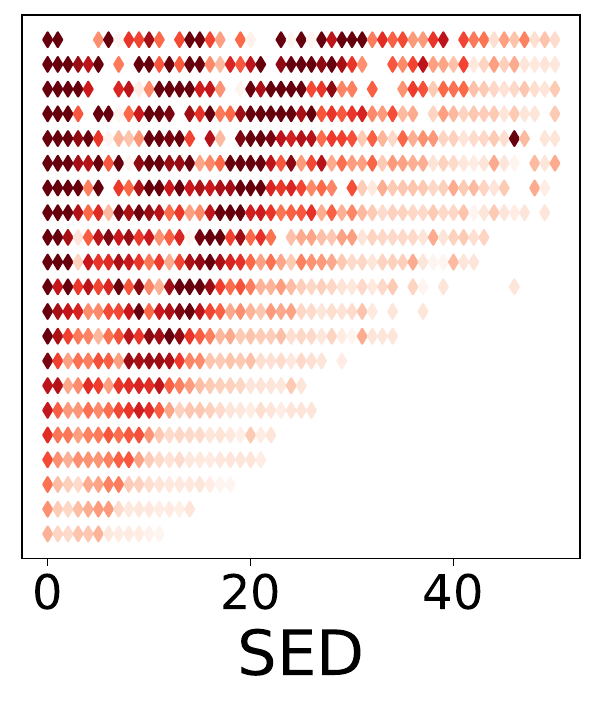}}
    \subfloat[\f]{\includegraphics[height=0.1845\textwidth]{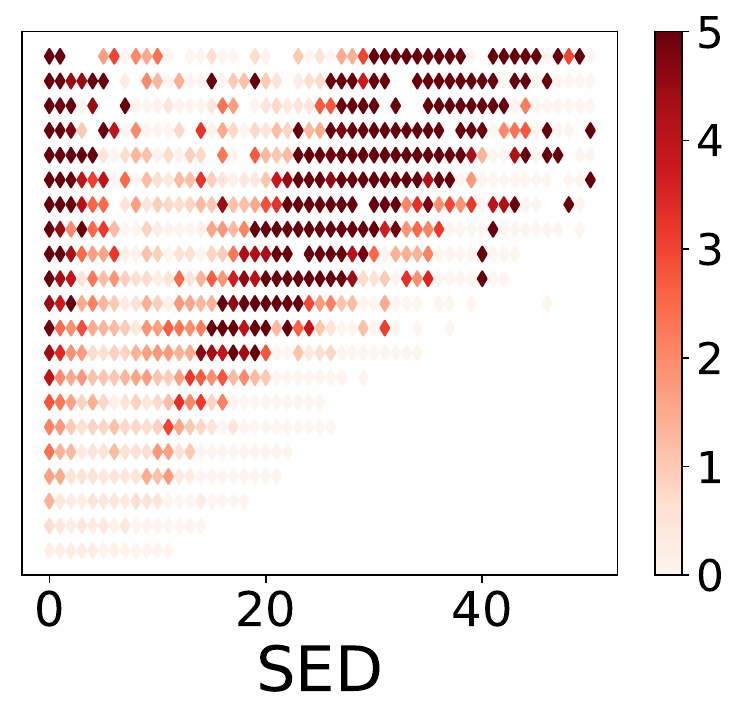}}\hfill
    \caption{\textbf{Heat Maps of \sed error against query size and \sed values for \seer. Darker means higher error.}}
    \label{fig:heat_citeseer}
    \vspace{-0.06in}
\end{figure}

\begin{figure}[ht]
    \centering
    \vspace{-0.02in}
    \captionsetup[subfigure]{labelfont={normalsize,bf},textfont={normalsize}}
    \subfloat[\name]{\includegraphics[height=0.18\textwidth]{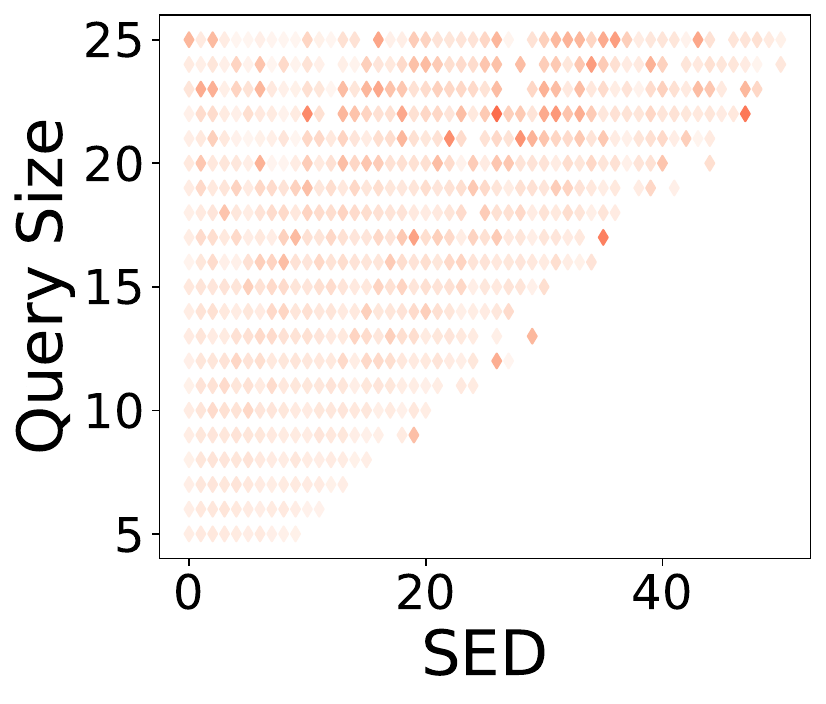}}
    \subfloat[\hmnr]{\includegraphics[height=0.18\textwidth]{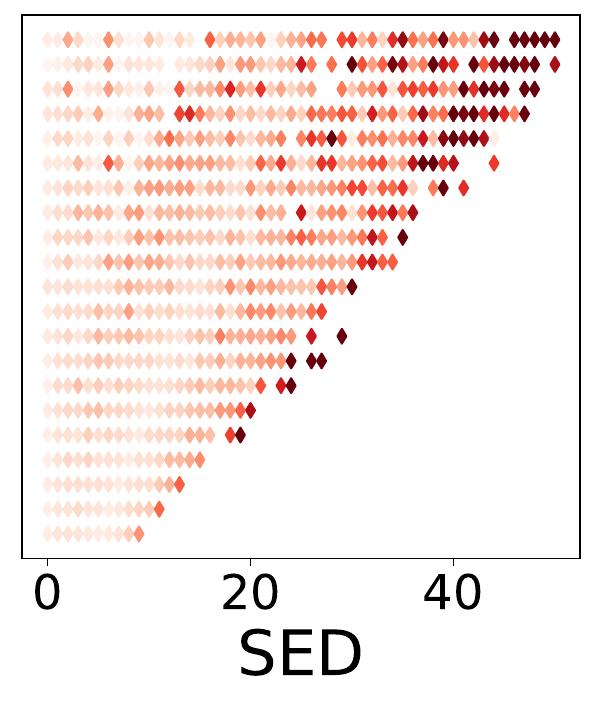}}
    \subfloat[\hmnn]{\includegraphics[height=0.18\textwidth]{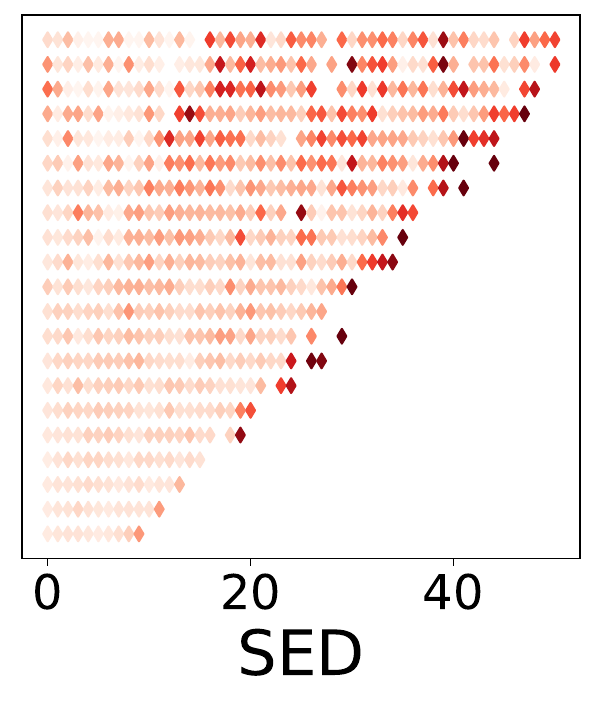}}
    \subfloat[\simgnn]{\includegraphics[height=0.18\textwidth]{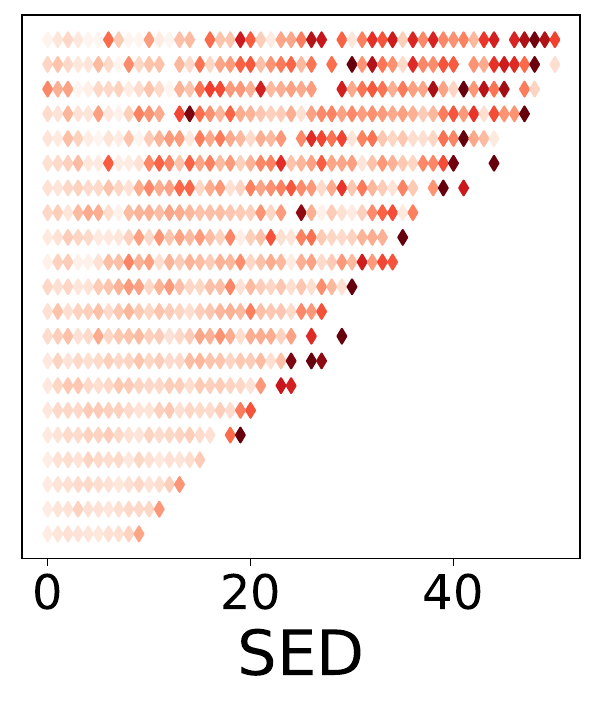}}
    \subfloat[\branch]{\includegraphics[height=0.18\textwidth]{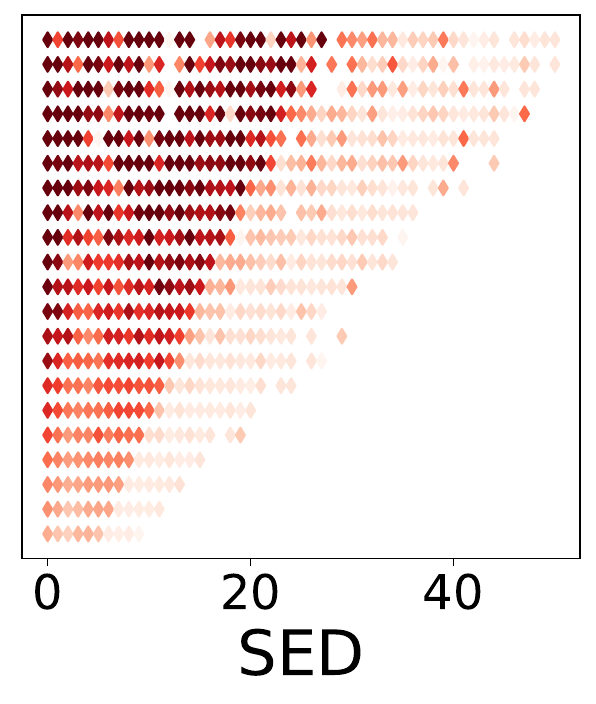}}
    \subfloat[\f]{\includegraphics[height=0.1845\textwidth]{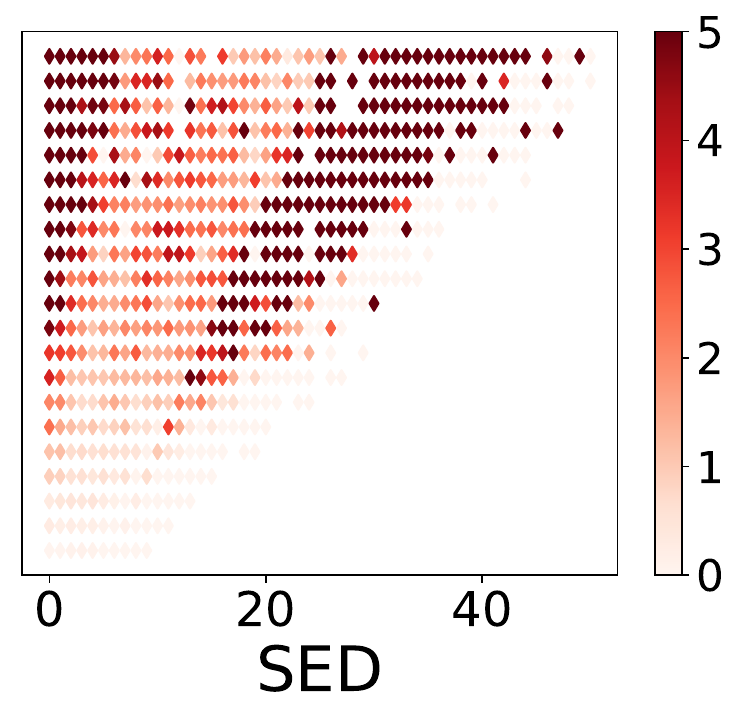}}\hfill
    \caption{\textbf{Heat Maps of \sed error against query size and \sed values for \cora. Darker means higher error.}}
    \label{fig:heat_cora}
    \vspace{-0.06in}
\end{figure}

\begin{figure}[ht]
    \centering
    \captionsetup[subfigure]{labelfont={normalsize,bf},textfont={normalsize}}
    \subfloat[\name]{\includegraphics[height=0.18\textwidth]{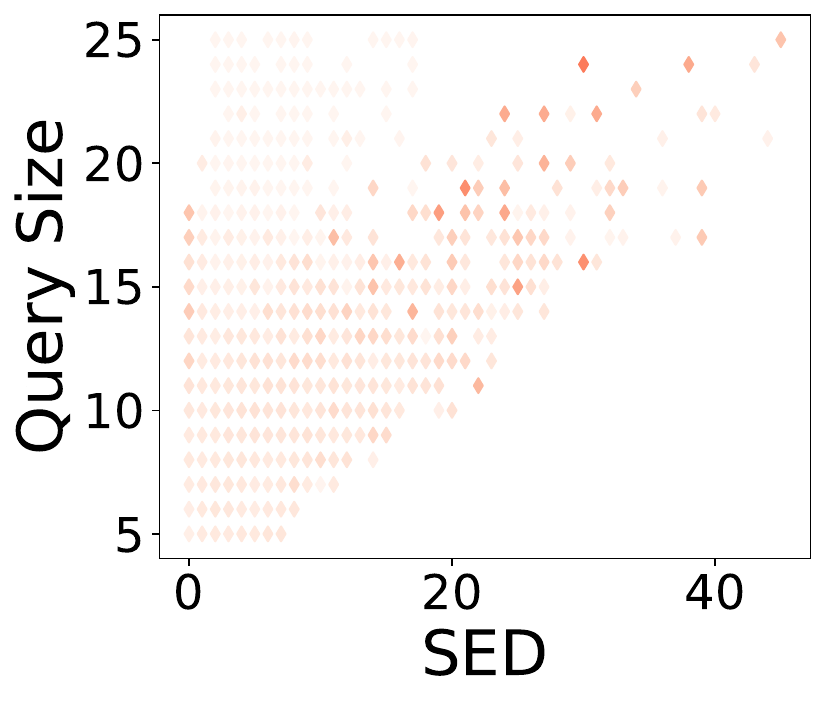}}
    \subfloat[\hmnr]{\includegraphics[height=0.18\textwidth]{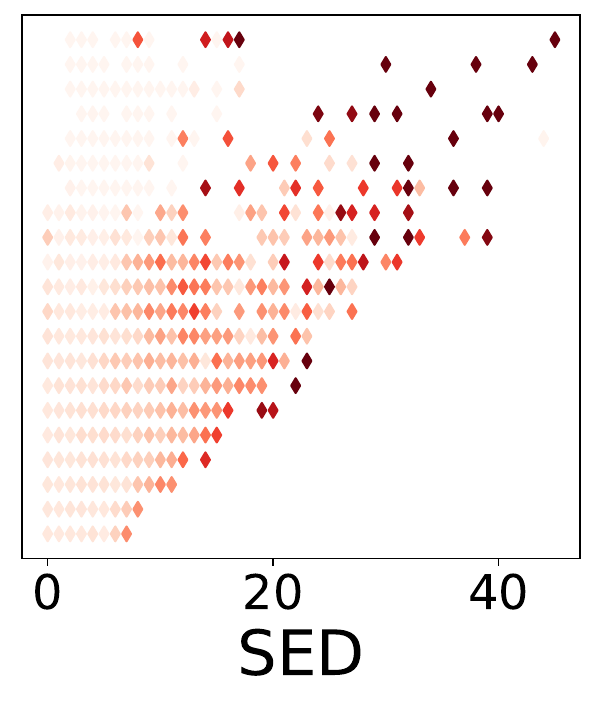}}
    \subfloat[\hmnn]{\includegraphics[height=0.18\textwidth]{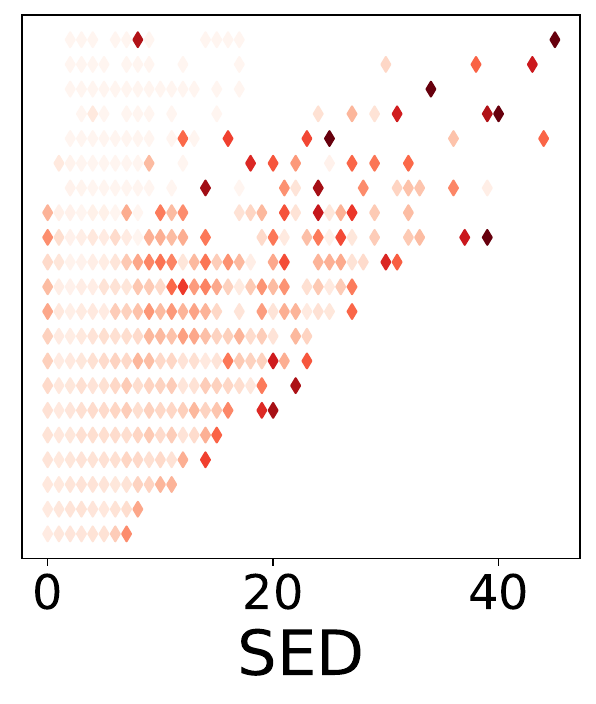}}
    \subfloat[\simgnn]{\includegraphics[height=0.18\textwidth]{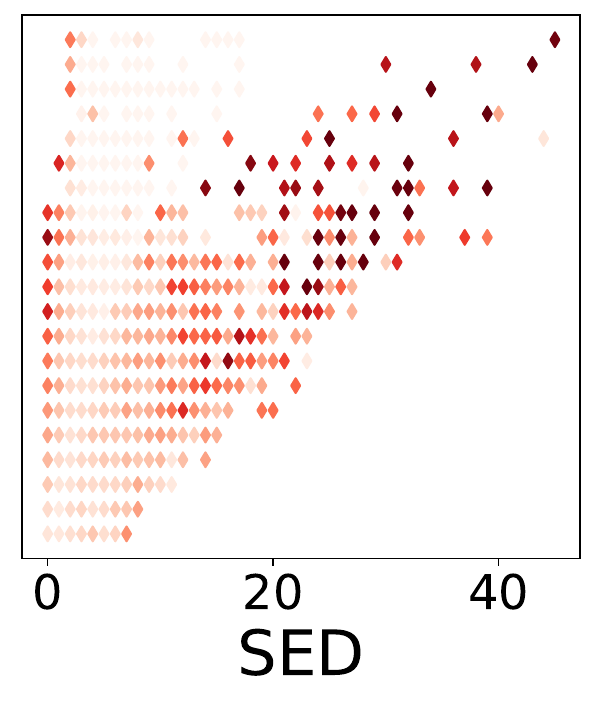}}
    \subfloat[\branch]{\includegraphics[height=0.18\textwidth]{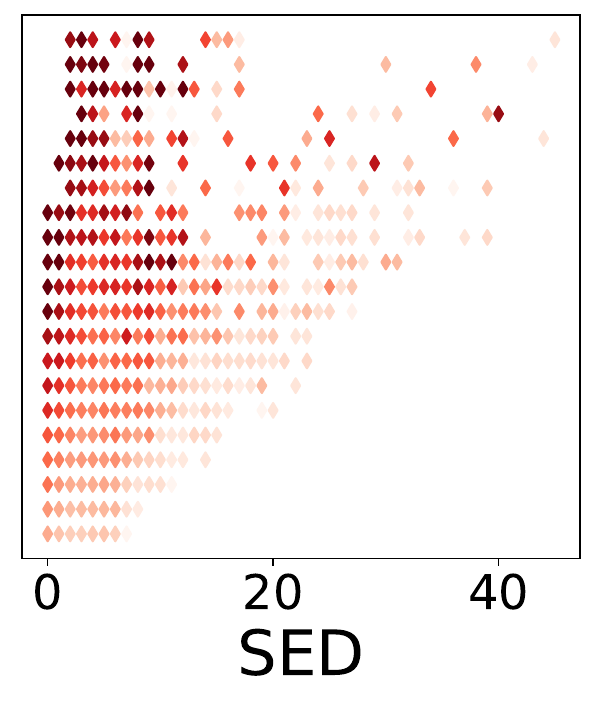}}
    \subfloat[\f]{\includegraphics[height=0.1845\textwidth]{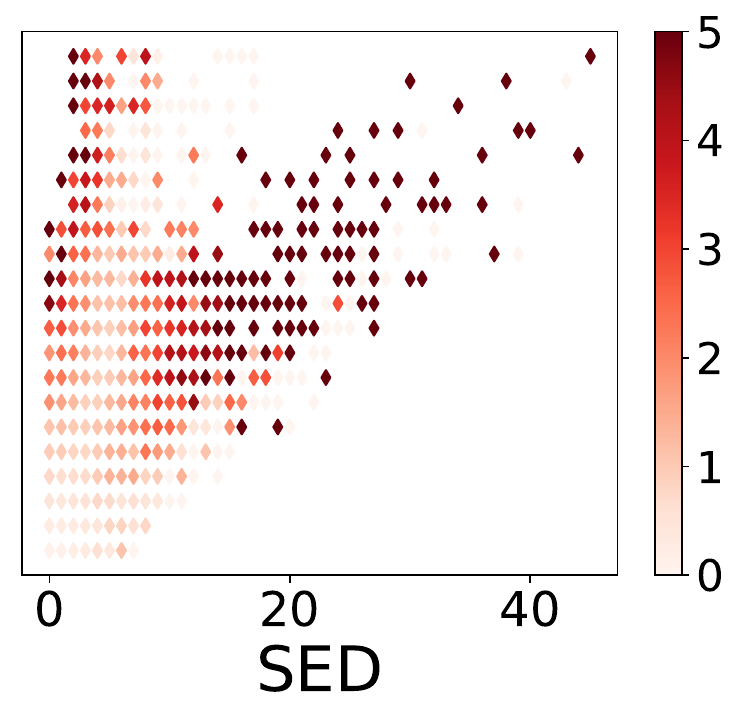}}\hfill
    \caption{\textbf{Heat Maps of \sed error against query size and \sed values for \protein. Darker means higher error.}}
    \label{fig:heat_protein}
    \vspace{-0.02in}
\end{figure}

\begin{figure}[ht]
    \centering
    \captionsetup[subfigure]{labelfont={normalsize,bf},textfont={normalsize}}
    \subfloat[\name]{\includegraphics[height=0.18\textwidth]{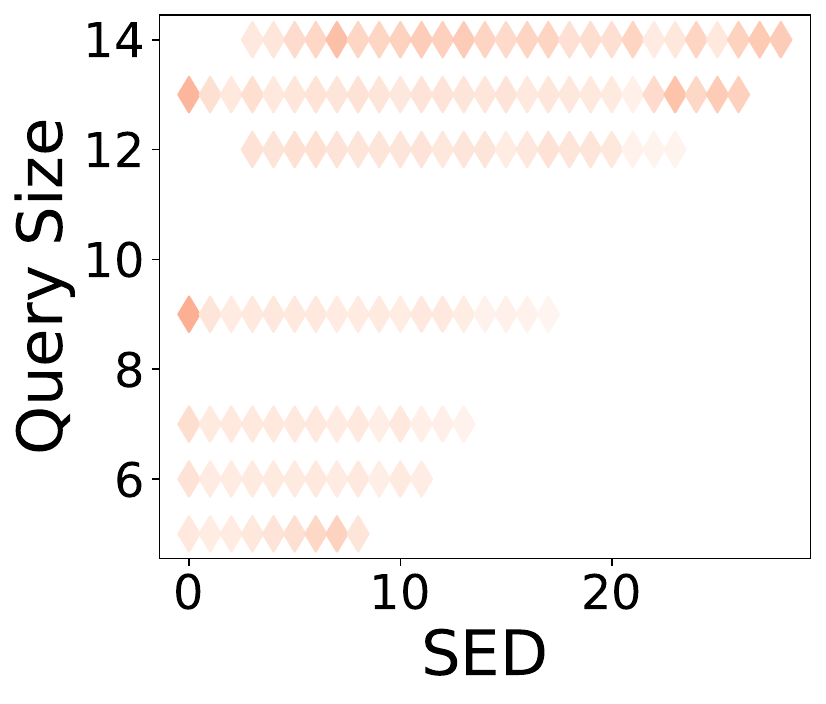}}
    \subfloat[\hmnr]{\includegraphics[height=0.18\textwidth]{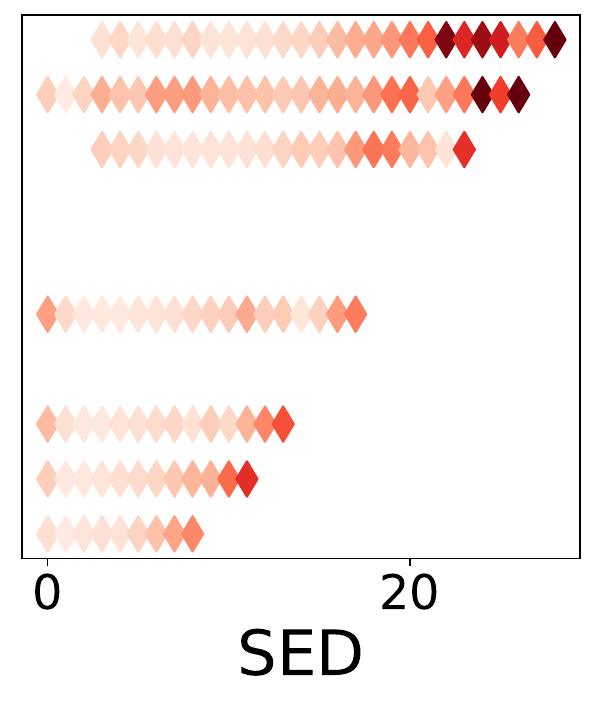}}
    \subfloat[\hmnn]{\includegraphics[height=0.18\textwidth]{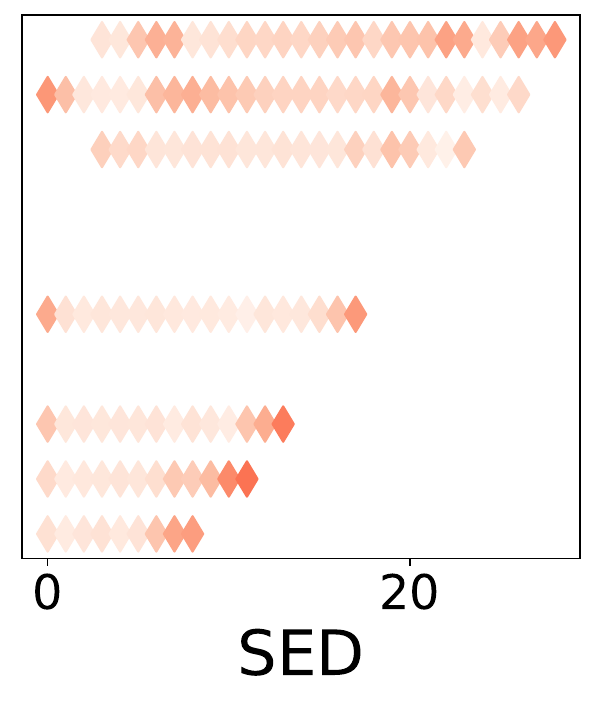}}
    \subfloat[\simgnn]{\includegraphics[height=0.18\textwidth]{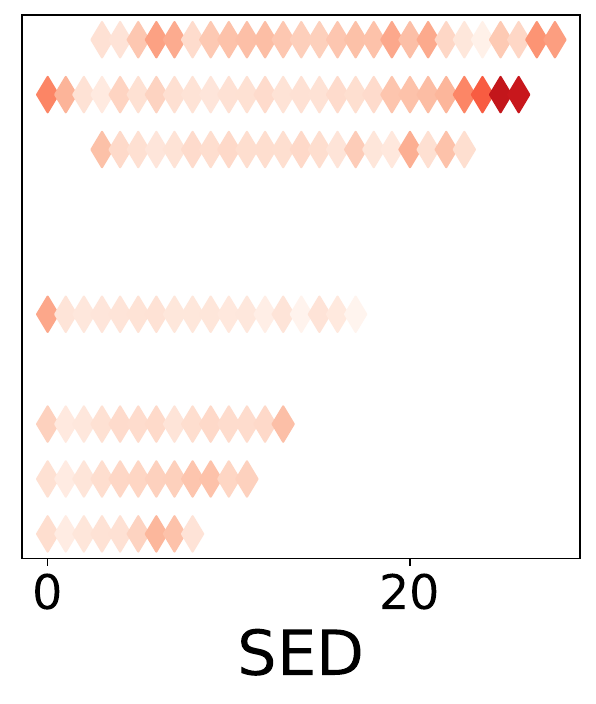}}
    \subfloat[\branch]{\includegraphics[height=0.18\textwidth]{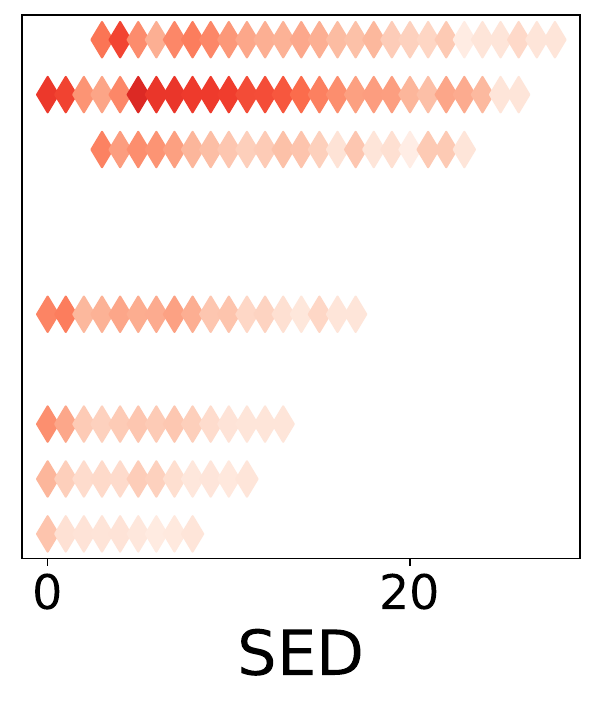}}
    \subfloat[\f]{\includegraphics[height=0.1845\textwidth]{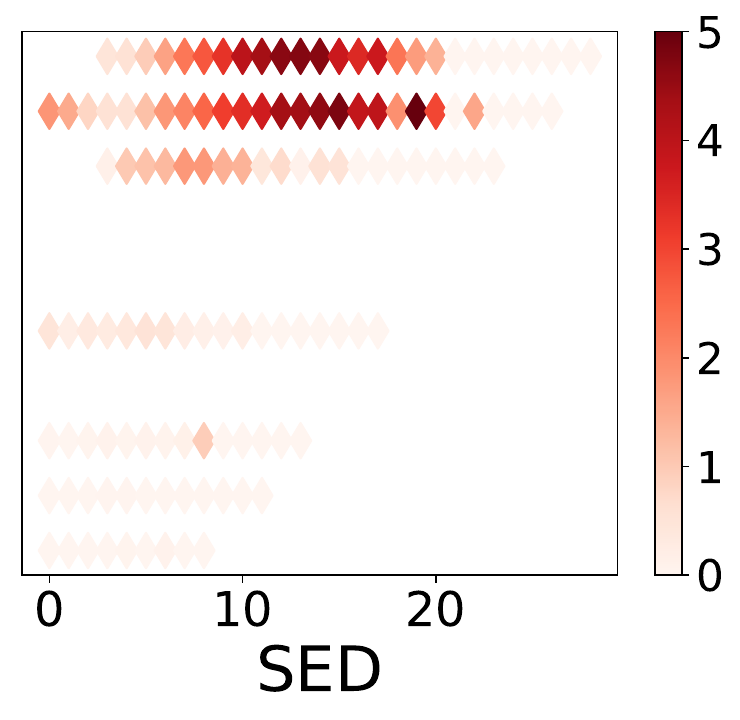}}\hfill
    \caption{\textbf{Heat Maps of \sed error against query size and \sed values for \aidsg. Darker means higher error.}}
    \label{fig:heat_aidsg}
    \vspace{-0.02in}
\end{figure}

\begin{figure}[ht]
    \centering
    \captionsetup[subfigure]{labelfont={normalsize,bf},textfont={normalsize}}
    \subfloat[\name]{\includegraphics[height=0.18\textwidth]{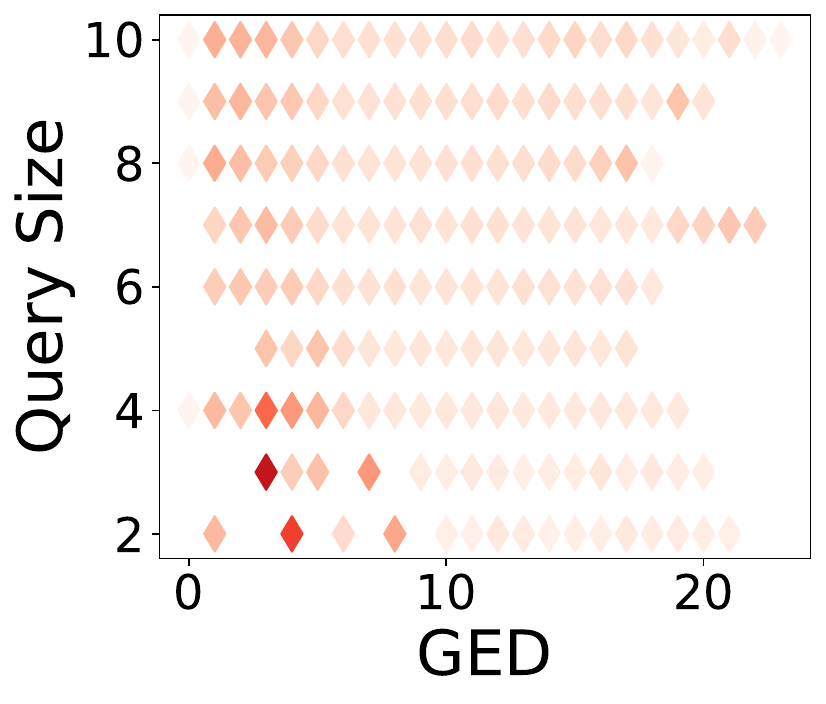}}
    \subfloat[\hmnr]{\includegraphics[height=0.18\textwidth]{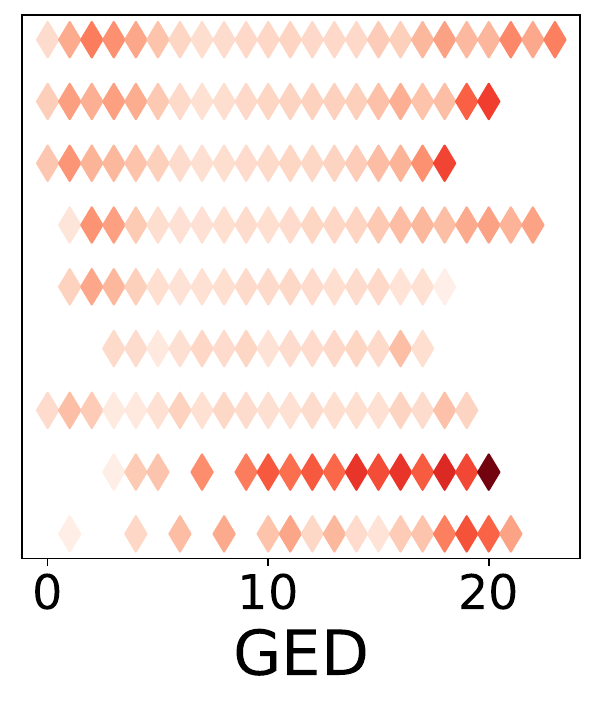}}
    \subfloat[\hmnn]{\includegraphics[height=0.18\textwidth]{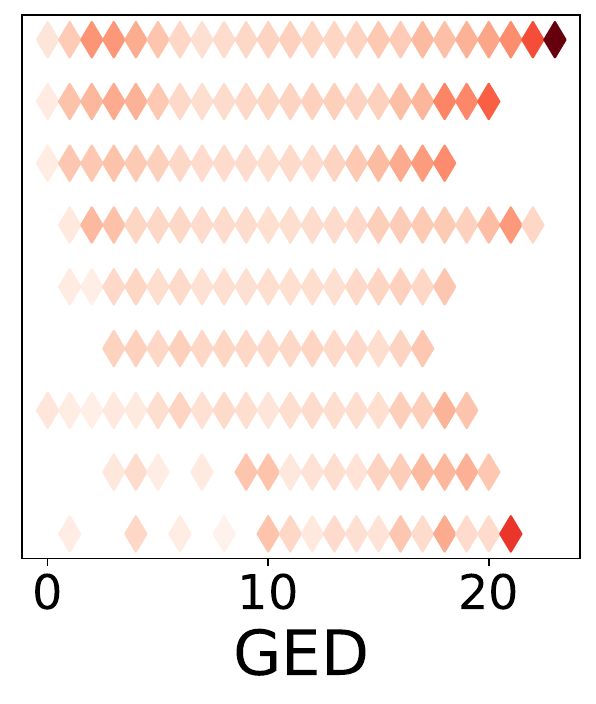}}
    \subfloat[\simgnn]{\includegraphics[height=0.18\textwidth]{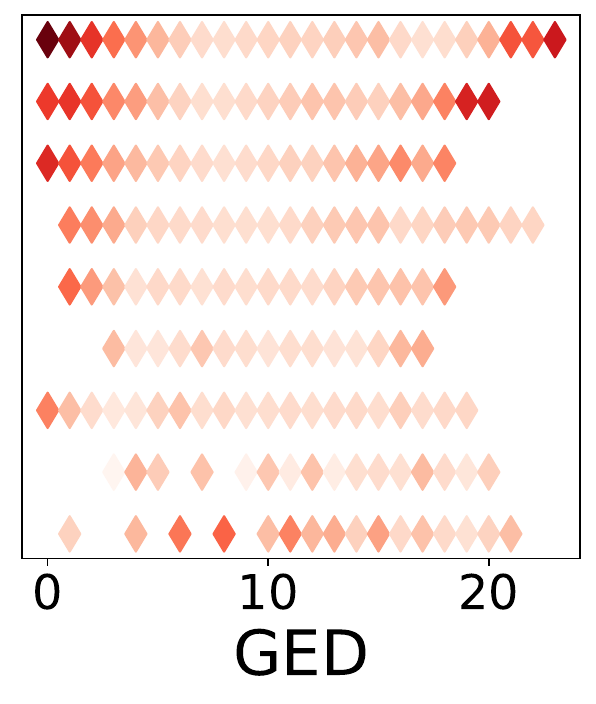}}
    \subfloat[\branch]{\includegraphics[height=0.18\textwidth]{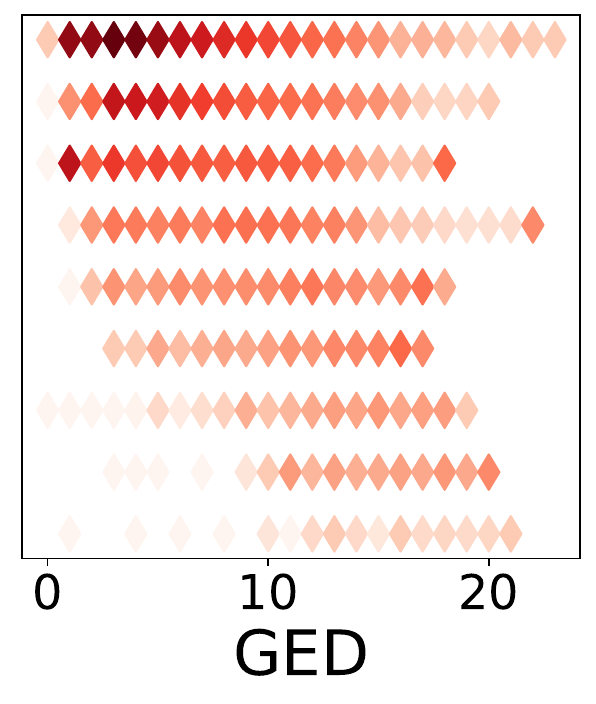}}
    \subfloat[\f]{\includegraphics[height=0.1845\textwidth]{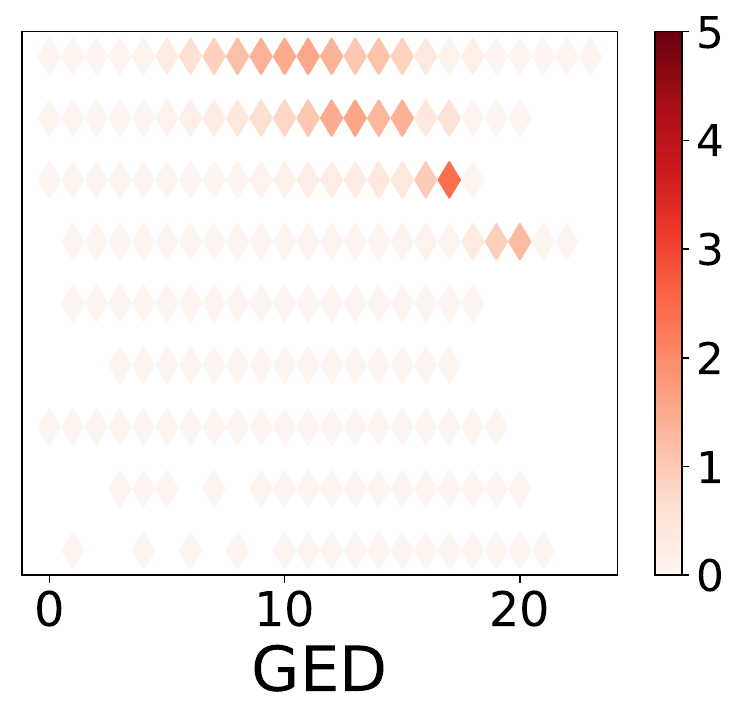}}\hfill
    \caption{\textbf{Heat Maps of \ged error against query size and \ged values for \aids. Darker means higher error.}}
    \label{fig:heat_aids_ged}
    \vspace{-0.06in}
\end{figure}

\begin{figure}[ht]
    \centering
    \vspace{-0.02in}
    \captionsetup[subfigure]{labelfont={normalsize,bf},textfont={normalsize}}
    \subfloat[\name]{\includegraphics[height=0.18\textwidth]{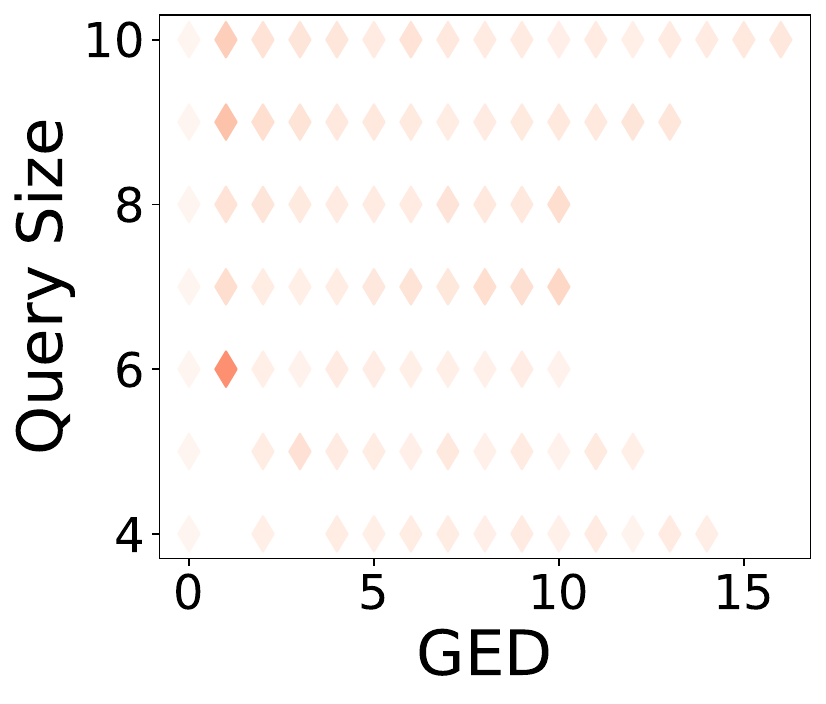}}
    \subfloat[\hmnr]{\includegraphics[height=0.18\textwidth]{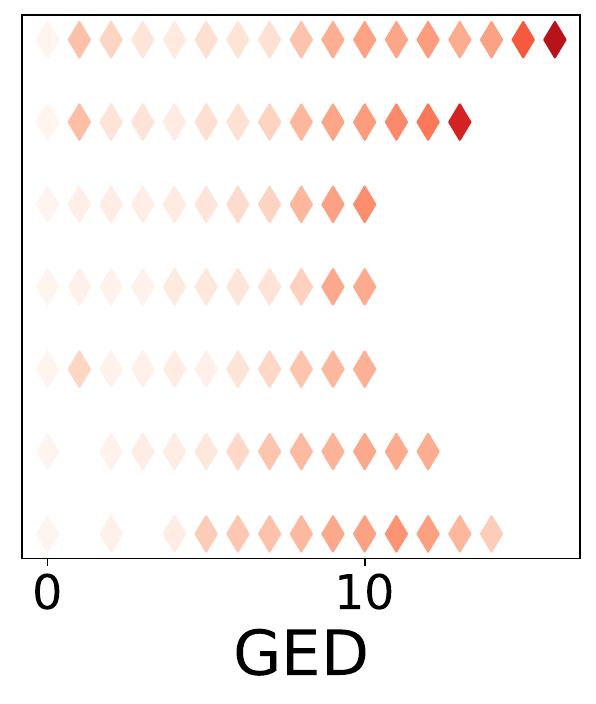}}
    \subfloat[\hmnn]{\includegraphics[height=0.18\textwidth]{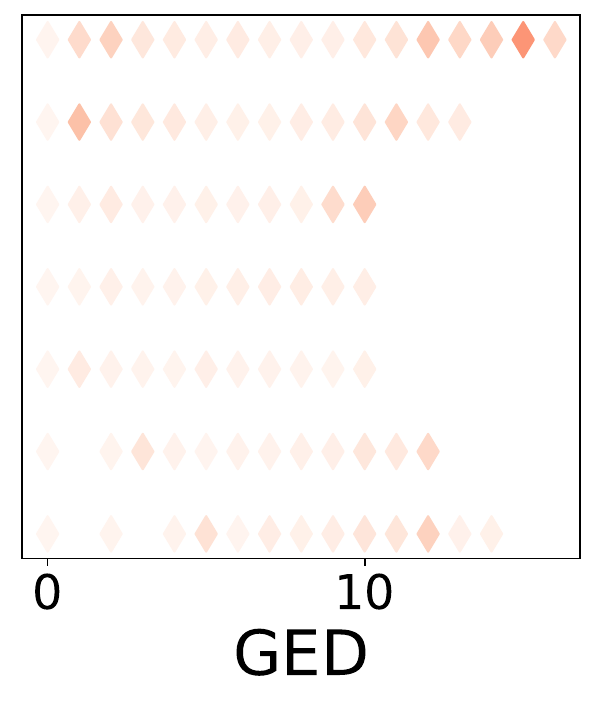}}
    \subfloat[\simgnn]{\includegraphics[height=0.18\textwidth]{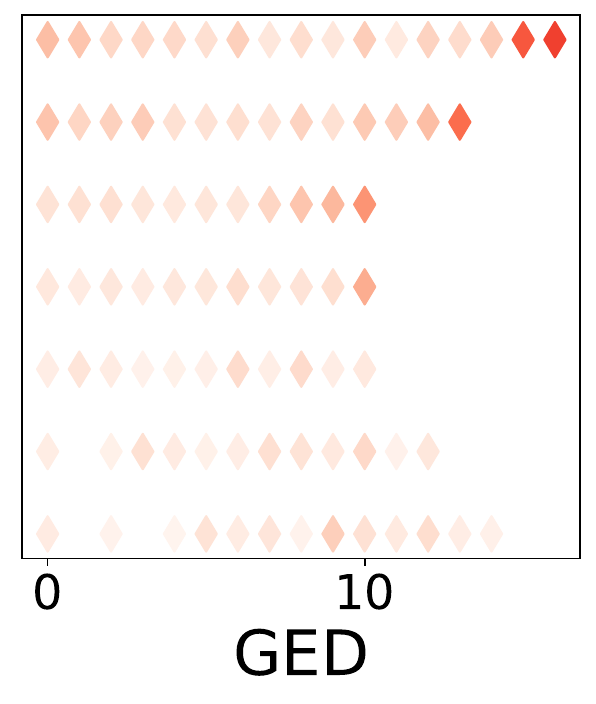}}
    \subfloat[\branch]{\includegraphics[height=0.18\textwidth]{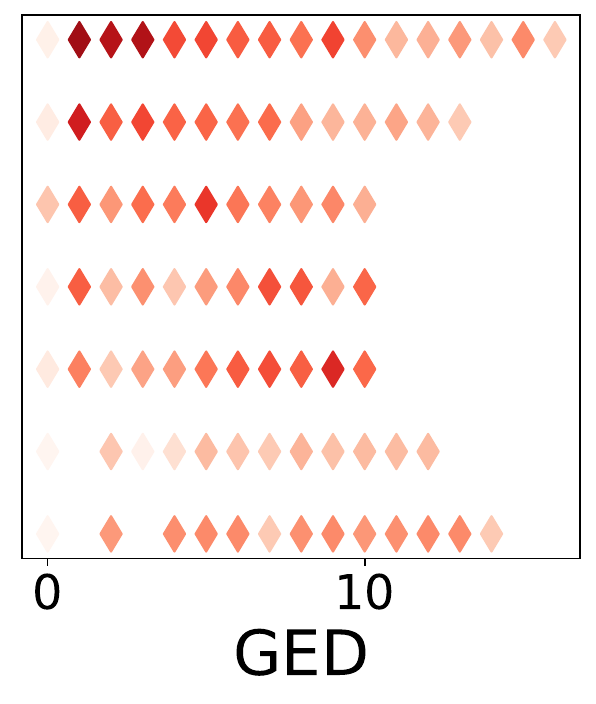}}
    \subfloat[\f]{\includegraphics[height=0.1845\textwidth]{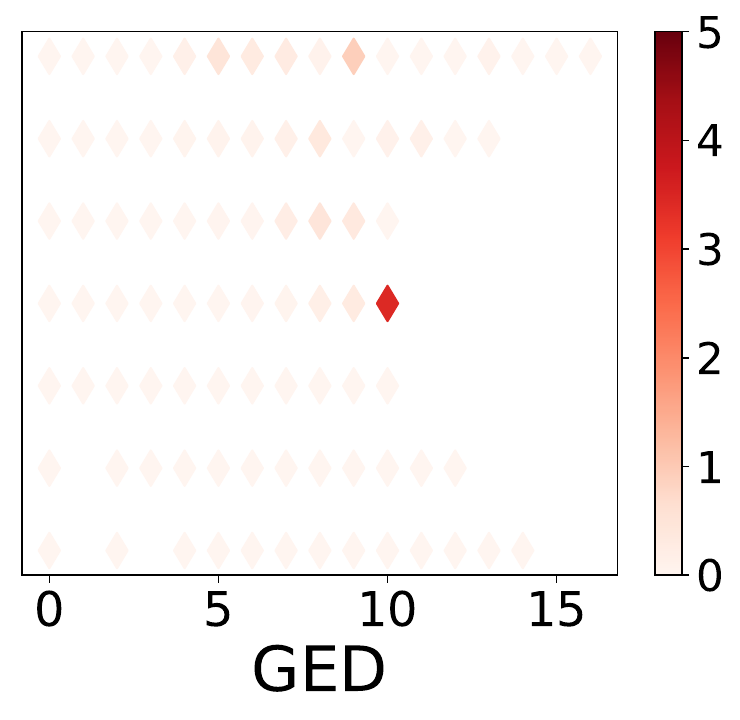}}\hfill
    \caption{\textbf{Heat Maps of \ged error against query size and \ged values for \linux. Darker means higher error.}}
    \label{fig:qheat_linux_ged}
    \vspace{-0.02in}
\end{figure}

\begin{figure}[ht]
    \centering
     \vspace{-0.06in}
    \captionsetup[subfigure]{labelfont={normalsize,bf},textfont={normalsize}}
    \subfloat[\name]{\includegraphics[height=0.18\textwidth]{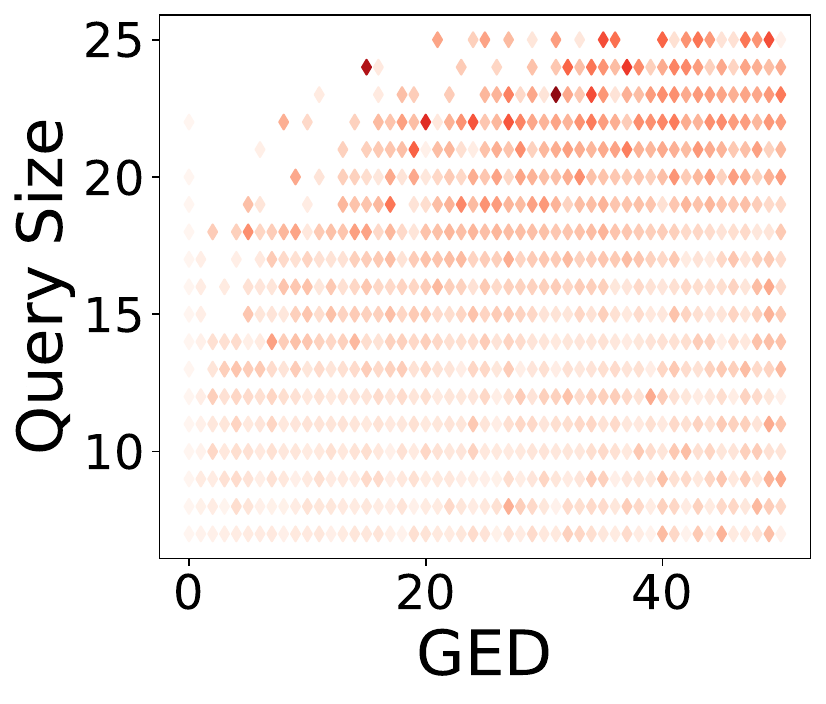}}
    \subfloat[\hmnr]{\includegraphics[height=0.18\textwidth]{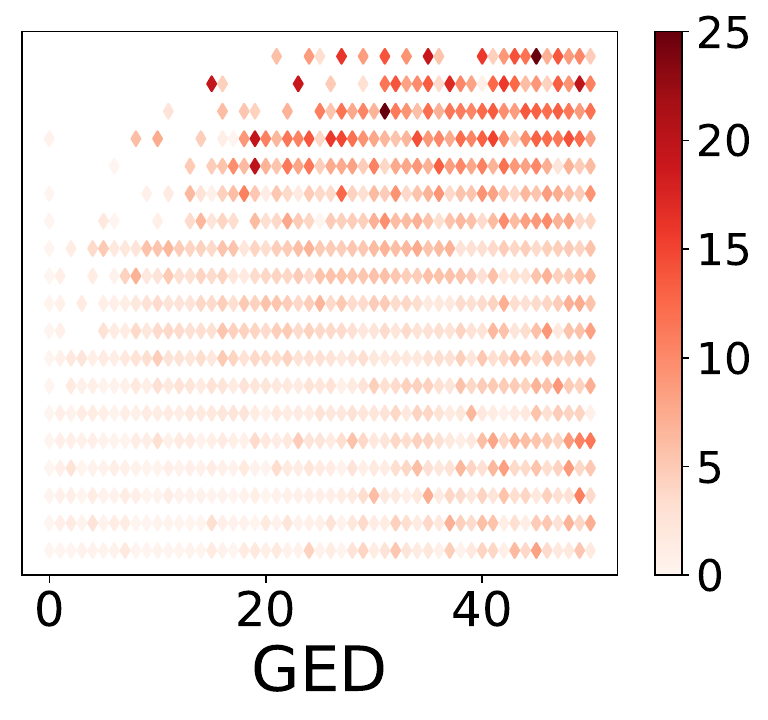}}
    \subfloat[\hmnn]{\includegraphics[height=0.18\textwidth]{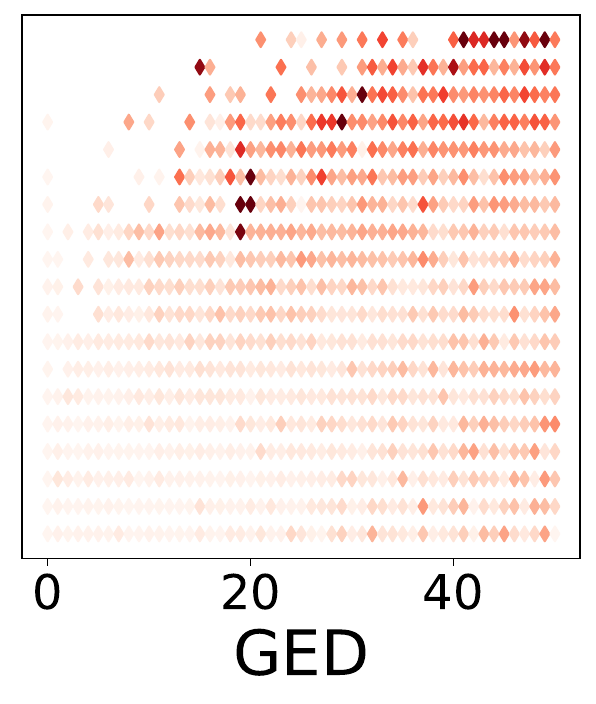}}
    \subfloat[\simgnn]{\includegraphics[height=0.18\textwidth]{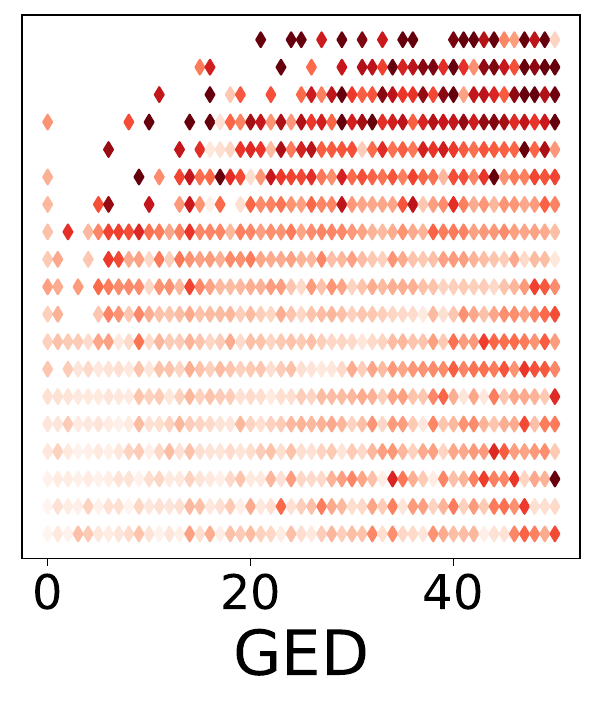}}
    \subfloat[\branch]{\includegraphics[height=0.18\textwidth]{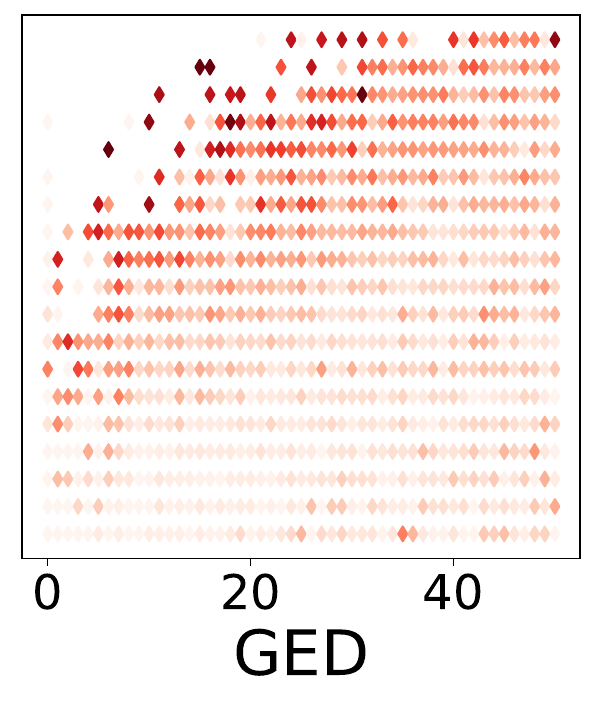}}
    \subfloat[\f]{\includegraphics[height=0.1845\textwidth]{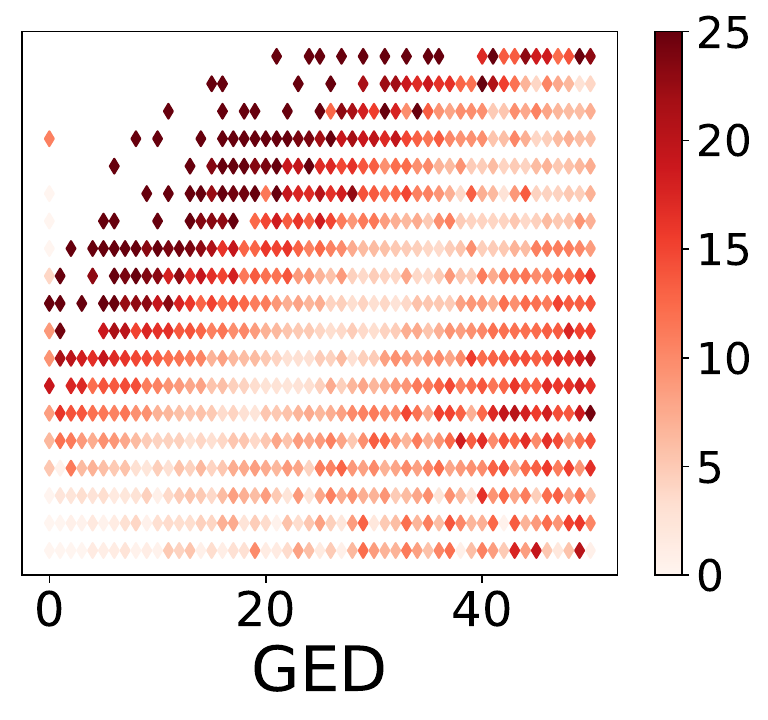}}\hfill
    \caption{\textbf{Heat Maps of \ged error against query size and \ged values for \imdb. Darker means higher error.}}
    \label{fig:heat_imdb_ged}
    \vspace{2in}
\end{figure}

In Figures \ref{fig:heat_dblp} to \ref{fig:heat_imdb_ged}, we show the variation of the errors on \sed and \ged prediction with query sizes and ground truth values for \name and the baselines on all the corresponding datasets. This experiment is an extension of the heat-map results in Fig.~\ref{fig:heatmap} in the main paper. These datasets show variations in the distributions of \sed and \ged values. It is interesting to observe that among the baselines, different methods perform well on different regions (i.e., combinations of high/low \sed and high/low query sizes). The baselines do not show good performance on all regions. However, for \name, we see a much better coverage for all types of regions in the domain. Furthermore, for every region, our models outperform (or are at least competitive with) the best performing baseline for that region.

%% file: 4_1_querying.tex
\renewcommand{\algorithmicrequire}{\textbf{Input:}}
\renewcommand{\algorithmicensure}{\textbf{Output:}}
\begin{algorithm}[t]
\caption{\textsc{BuildIndex}}
\label{alg:indexing}
{\scriptsize
\begin{algorithmic}[1]
\Require Embeddings $\mathbb{D}$ of graphs
\Ensure Root node of the constructed tree
\If{$\mathbb{D} = \emptyset$} \Return \textsc{null} \EndIf
\State $\CZ_{\CP}\leftarrow$ arbitrary embedding in  $\mathbb{D}$ as pivot
\State $m_1 \gets \textsc{Median}(\{\CF_s(\CZ_{\CP},\CZ_{\CG}) : \CZ_{\CG} \in \mathbb{D}\setminus\{\CZ_{\CP}\}\})$
\State $m_2 \gets \textsc{Median}(\{\CF_s(\CZ_{\CG},\CZ_{\CP}) : \CZ_{\CG} \in \mathbb{D}\setminus\{\CZ_{\CP}\}\})$
\State $\mathbb{D}_1 \gets \{\CZ_{\CG} : \CF_s(\CZ_{\CP},\CZ_{\CG}) \le m_1, \CF_s(\CZ_{\CG},\CZ_{\CP}) \le m_2\}$
\State $\mathbb{D}_2 \gets \{\CZ_{\CG} : \CF_s(\CZ_{\CP},\CZ_{\CG}) \le m_1, \CF_s(\CZ_{\CG},\CZ_{\CP}) > m_2\}$
\State $\mathbb{D}_3 \gets \{\CZ_{\CG} : \CF_s(\CZ_{\CP},\CZ_{\CG}) > m_1, \CF_s(\CZ_{\CG},\CZ_{\CP}) \le m_2\}$
\State $\mathbb{D}_4 \gets \{\CZ_{\CG} : \CF_s(\CZ_{\CP},\CZ_{\CG}) > m_1, \CF_s(\CZ_{\CG},\CZ_{\CP}) > m_2\}$
\For{$i = 1$ to $4$}
\State $t_i \gets \textsc{BuildIndex}(\mathbb{D}_i)$
\EndFor
\State \textbf{Return} $\textsc{node}(\CZ_{\CP}, m_1, m_2, t_1, t_2, t_3, t_4)$
\end{algorithmic}}
\end{algorithm}

\begin{algorithm}[b]
\caption{\textsc{RangeQuery}}
\label{alg:rangequery}
\scriptsize{
\begin{algorithmic}[1]
\Require Query embedding $\CZ_{\CG_{\CQ}}$, threshold $\theta$, root node $t=\textsc{node}(\CZ_{\CP}, m_1, m_2, t_1, t_2, t_3, t_4)$
\Ensure $\mathbb{A}\leftarrow\{\CZ_{\CG} \mid \CF_s(\CZ_{\CG_{\CQ}},\CZ_{\CG}) \le \theta\}$
\If{$t = \textsc{null}$} \Return $\emptyset$\EndIf
\If{$\CF_s(\CZ_{\CP},\CZ_{\CG_{\CQ}}) \le m_1-\theta$} \State mark $t_3$ and $t_4$ for pruning \EndIf
\If{$\CF_s(\CZ_{\CG_{\CQ}},\CZ_{\CP}) > m_2+\theta$} \State mark $t_1$ and $t_3$ for pruning \EndIf
\If{$\CF_s(\CZ_{\CG_{\CQ}},\CZ_{\CP}) \le \theta-m_1$}
\State $\mathbb{A} \gets \mathbb{A}\cup \mathbb{D}_1\cup \mathbb{D}_2$ 
\State mark $t_1$ and $t_2$ for pruning
\Else
\State $\mathbb{A} \gets \emptyset$
\EndIf
\For{$i = 1$ to $4$}
\If{$t_i$ is not marked for pruning}
\State $\mathbb{A} \gets \mathbb{A} \cup \textsc{RangeQuery}\left(\CZ_{\CG_{\CQ}}, \theta, t_i\right)$
\EndIf
\EndFor
\State \textbf{Return} $\mathbb{A}$
\end{algorithmic}}
\end{algorithm}
\vspace{-0.05in}

\subsection{Querying in the Embedding Space}
\label{app:index}
We assume the standard querying setup where the database graphs are known apriori, while the query graph is unseen and provided at query time. Since \name generates pair-independent embeddings, representations of the database graphs can be generated apriori and stored. Furthermore, due to both the predicted \ged and \sed satisfying the triangle inequality, the embeddings can be indexed. Thus, at query time, we need to perform only two operations: \textbf{(1)} embed the query graph $\CG_{\CQ}$, and \textbf{(2)} scan the database embeddings against the query embedding to compute the answer set. We focus the discussion on the two most common database queries of \textit{range} and \textit{$k$-NN} queries.
\begin{definition}[Range Query]
Given a database $\mathbb{D}=\{\CZ_{\CG_1},\cdots, \CZ_{\CG_n}\}$ of graph embeddings, a query graph $\CG_{\CQ}$ and a threshold $\theta$, find the answer set $\mathbb{A}=\{\CG_i\mid \CF_g(\CZ_{\CG_i},\CZ_{\CG_{\CQ}})\leq \theta\}$.
\end{definition}
\begin{definition}[$k$-NN Query]
Given a database $\mathbb{D}=\{\CZ_{\CG_1},\cdots, \CZ_{\CG_n}\}$ of graph embeddings, a query graph $\CG_{\CQ}$ and $k$, find the database graphs with the $k$ smallest distance to $\CG_{\CQ}$ as per $\CF_g(\CZ_{\CG_i},\CZ_{\CG_{\CQ}})$.
\end{definition}
The above definitions can be adopted for \sed by using $\CF_s$. For the rest of the discussion, we assume $\CF_s$ since due to asymmetry, \sed requires some additional considerations. 

\vspace{-0.05in}
\subsubsection{Indexing}
\label{sec:indexing}
We exploit the triangle inequality of \ged and \sed to index the database embeddings. Alg.~\ref{alg:indexing} presents the pseudocode. We choose a random embedding $\CZ_{\CP} \in \mathbb{D}$ as the \textit{pivot} (line 2), based on which we split the remaining embeddings into four groups (lines 3-8). 
This process continues recursively on each group (lines 9-10) till a partition gets empty (line 1). Note that in \ged the distances are symmetric and hence $m_1$ and $m_2$ will converge. Consequently, we will have two partitions at each node instead of four.
\vspace{-0.05in}
\subsubsection{Range Query}
\label{sec:rangequery}
 From the triangle inequality, we can infer the lower bounds listed in lines 2 and 4 of Alg.~\ref{alg:rangequery}. Hence, if these bounds are larger than $\theta$, the corresponding sub-trees are pruned. Similarly, if the upper bound is smaller than $\theta$ (line 6), the entire sub-tree is added to the answer set (line 7). Otherwise, we recurse (lines 11-13).
\begin{algorithm}[b]
\caption{$k$-NN}\label{alg:knn}
{\scriptsize
\begin{algorithmic}[1]
\Require Query embedding $\CZ_{\CG_{\CQ}}$, $k$, root node $t=\textsc{node}(\CZ_{\CP}, m_1, m_2, t_1, t_2, t_3, t_4)$
\Ensure $\mathbb{A}\leftarrow$ top-$k$ closest embeddings in $\mathbb{D}$
\State $\mathbb{A}\leftarrow$ A priority queue of size up to $k$. Stores entries in descending order of distance.
\State $Cands\leftarrow$ A priority queue. Stores entries in ascending order of distance lower bound.
\State $Cands$.insert$\left(\left\langle t, \CF_s\left(\CZ_{\CG_{\CQ}},\CZ_{\CP}\right) \right\rangle\right)$
\While {$Cands.size()>0$ \textbf{And} $Cands.top().LB<\mathbb{A}.top().distance$ }
\State $\CP\leftarrow Cands.pop()$
\If {$|\mathbb{A}|<k$ \textbf{Or} $\CF_s\left(\CZ_{\CG_{\CQ}},\CZ_{\CP}\right)<\mathbb{A}.top().distance$}
\State $\mathbb{A}.insert\left(\left\langle \CP,\CF_s\left(\CZ_{\CG_{\CQ}},\CZ_{\CP}\right)\right\rangle\right)$
\EndIf
\State $LB_1\leftarrow |\CF_s\left(\CZ_{\CP},\CZ_{\CG_{\CQ}}\right) -m_1|$
\State $LB_2\leftarrow |\CF_s\left(\CZ_{\CG_{\CQ}},\CZ_{\CP}\right) -m_2|$
\State $LB=\max\{LB_1,LB_2\}$
\If {$ LB \leq \theta$}
\State $Cands.insert\left(\left\langle t_1,LB \right\rangle\right)$
\State $Cands.insert\left(\left\langle t_2,LB \right\rangle\right)$
\State $Cands.insert\left(\left\langle t_3,LB \right\rangle\right)$
\State $Cands.insert\left(\left\langle t_ 4,LB \right\rangle\right)$
\EndIf
\EndWhile
\State \textbf{Return} $\mathbb{A}$
\end{algorithmic}}
\end{algorithm}
\vspace{-0.05in}
\subsubsection{$k$-NN query}
\label{sec:knn}
$k$-NN utilizes the same bounds from Alg.~\ref{alg:rangequery} to prune and prioritize the search space. However, exploration proceeds in a \textit{best-first search} manner. Alg.~\ref{alg:knn} presents the pseudocode. Alg.~\ref{alg:knn} maintains two priority-queues; one to keep track of the $k$-NN till the current stage of the search process ($\mathbb{A}$ in line 1), and the second to store index nodes in ascending order of their lower bound distance ($Cands$ in line 2). We pop the \emph{best} node $\CP$ from $Cands$ and include it to the answer set if the distance is within top-$k$ (lines 6-7). Further, the sub-tree at $\CP$ is processed if it satisfies the lower bound criteria (lines 8-15). The search ends when either $Cands$ is either empty or the lower bound of the top-most node is larger than the $k$-th distance in $\mathbb{A}$ (line 4). In case of \ged, $LB_1$ (line 8) and $LB_2$ (line 9) will converge to the same value due to symmetry
\setcounter{table}{0}.
\begin{table}[t]
\centering
\captionsetup[subfigure]{labelfont={normalsize,bf},textfont={normalsize},position=b}
\scalebox{0.65}{
\begin{tabular}{rrrrrrr}
    \toprule
       \textbf{Name}  & $Avg. |\CV|$&$ Avg. |\CE|$& $|\Sigma|$& \#Graphs & $Avg. |\CV_{\CQ}|$&$ Avg. |\CE_{\CQ}|$\\
       \midrule
       \textbf{\dblp} & $1.66M$& $7.2M$ & $8$ & $1$&$15$&$14$ \\
       \textbf{\amz} & $334k$& $925k$ & $1$ & $1$&$12$&$16$ \\
        \textbf{\pubmed} & $19.7k$& $44.3k$ & $3$ & $1$&$12$&$11$ \\
        \textbf{\seer} & $4.2k$& $5.3k$ & $6$ & $1$&$12$&$12$ \\
        \textbf{\cora}& $3k$& $8.2k$ & $7$ & $1$&$11$&$11$ \\
        \textbf{\protein}& $38$ & $70$ & $3$ & $1,071$&$9$&$11$\\ 
        \textbf{\aidsg}& $14$& $15$ & $38$ & $1,811$& $7$&$7$\\
        \textbf{\aids}& $9$ & $9$ & $29$ & $700$& $9$ & $9$\\
        \textbf{\linux}& $8$& $7$ & $1$ & $1,000$& $8$ & $7$ \\
        \textbf{\imdb}& $13$& $65$ & $1$ & $1,500$& $13$ & $65$ \\
        \bottomrule
    \end{tabular}}
    \caption{Datasets}
        \label{tab:datasets}
\end{table}

\vspace{-0.05in}
\subsubsection{Scaling \sed to million-scale graphs}
\label{sec:scaling}
\sed is typically encountered in situations where the query is a small graph ($<50$ nodes)~\cite{querysize1,querysize2} and the target graph is a single large graph, potentially containing millions of nodes and edges. To scale to million-sized target graphs, we perform \textit{neighborhood decomposition}. Specifically, we extract the $k$-hop neighborhood $\CG_v$ around each node $v\in \CV_{\CT}$ in the target graph $\CG_{\CT}$, embed them into feature space using \name, and then indexed as outlined \S~\ref{sec:indexing}. The distance between query $\CG_{\CQ}$ and $\CG_{\CT}$ is computed as:
\vspace{-0.05in}
\begin{equation}
    \CF_s\left(\CG_{\CQ},\CG_{\CT}\right)=\min_{\forall \CG_v\in\CG_{\CT}} \left\{\CF_s\left(\CG_{\CQ},\CG_{v}\right)\right\}
\end{equation}
We next show that as long as the $k$-hop neighborhoods are \textit{sufficiently large}, the proposed neighborhood decomposition strategy is optimal. 


\begin{thm}
\label{thm:sub_in_nbr}
A nearest subgraph $\CS\subseteq\CG_{\CT}$ to $\CG_{\CQ}$ can be found in an $l/2$-hop neighborhood $\CG_v\subseteq\CG_{\CT}$, centered at some $v \in \CV_{\CG_{\CT}}$, where $l$ is the length of the longest path in $\CG_{\CQ}$.
\end{thm}
\textsc{Proof.} From Thm.~\ref{thm:sed_as_ged}, $\sed(\CG_{\CQ}, \CG_{\CT}) = \widehat{\ged}(\CG_{\CQ}, \CG_{\CT})$. Let $\pi$ be the node map produced by $\widehat{\ged}(\CG_{\CQ}, \CG_{\CT})$. Let us now consider the subgraph $\CS\subseteq \CG_{\CT}$ induced by all node maps of $\pi$ that are not inserts. It follows that $\sed(\CG_{\CQ}, \CG_{\CT}) = \widehat{\ged}(\CG_{\CQ}, \CG_{\CT})=\sed(\CG_{\CQ}, \CS)$. Since there are no inserts, the topology of $\CS$ is a subgraph of $\CG_1$ (i.e., subgraph isomorphic if we ignore label information). Hence,  the diameter of $\CS$ is $\le$ the length $l$ of the longest path in $\CG_{\CQ}$. Since $\CS \subseteq \CG_{\CT}$, $\CS$ is contained in some $l/2$ neighborhood $\CG_v$ of $\CG_{\CT}$. $\hfill\square$

Finding the length of the longest path is NP-hard. However, we do not need to find the exact length of the longest path: any upper bound suffices. Moreover we do not even need the length of the longest path for $l$. The nearest subgraph having a diameter equal to $l$ is rare. In practice, the diameter of the nearest subgraph is unlikely to be much higher than the diameter of the query itself.